\documentclass{article} % For LaTeX2e
\usepackage{iclr2020_conference,times}

\usepackage{hyperref}
\usepackage{url}

\usepackage[utf8]{inputenc} % allow utf-8 input
\usepackage[T1]{fontenc}    % use 8-bit T1 fonts
\usepackage{booktabs}       % professional-quality tables
\usepackage{amsfonts}       % blackboard math symbols
\usepackage{nicefrac}       % compact symbols for 1/2, etc.
\usepackage{microtype}      % microtypography
\usepackage[noend]{algcompatible}
\usepackage{algorithm}
\usepackage{amsthm}
\usepackage{amssymb}

\usepackage{graphicx}
\usepackage{subfigure}
\usepackage{mathtools}
\usepackage[pdf]{pstricks}
\usepackage{epsfig}
\usepackage{pst-grad} % For gradients
\usepackage{pst-plot} % For axes
\usepackage{comment}
\usepackage{wrapfig}

\newtheorem{innercustomgeneric}{\customgenericname}
\providecommand{\customgenericname}{}
\newcommand{\newcustomtheorem}[2]{%
  \newenvironment{#1}[1]
  {%
   \renewcommand\customgenericname{#2}%
   \renewcommand\theinnercustomgeneric{##1}%
   \innercustomgeneric
  }
  {\endinnercustomgeneric}
}

\newcustomtheorem{customlemma}{Lemma}
\newcustomtheorem{customthm}{Theorem}
\newcustomtheorem{customcorollary}{Corollary}

\title{Robust Reinforcement Learning for Continuous Control with Model Misspecification}

\author{
  Daniel J. Mankowitz\thanks{Equal contribution} , Nir Levine$^*$, Rae Jeong, Abbas Abdolmaleki, Jost Tobias Springenberg, \\\textbf{Yuanyuan Shi\thanks{Work done during an internship at Deepmind}, Jackie Kay, Timothy Mann, Todd Hester, Martin Riedmiller}\\
  DeepMind\\
  \texttt{\{dmankowitz, nirlevine, raejeong, aabdolmaleki, springenberg}\\
  \texttt{yyshi, kayj, timothymann, toddhester, riedmiller\}@google.com}
}

\iclrfinalcopy % Uncomment for camera-ready version, but NOT for submission.
\begin{document}

\maketitle

\begin{abstract}
We provide a framework for incorporating robustness -- to perturbations in the transition dynamics which we refer to as model misspecification -- into continuous control Reinforcement Learning (RL) algorithms. We specifically focus on incorporating robustness into a state-of-the-art continuous control RL algorithm called Maximum a-posteriori Policy Optimization (MPO). We achieve this by learning a policy that optimizes for a worst case expected return objective and derive a corresponding robust entropy-regularized Bellman contraction operator. In addition, we introduce a less conservative, soft-robust, entropy-regularized objective with a corresponding Bellman operator. We show that both, robust and soft-robust policies, outperform their non-robust counterparts in nine Mujoco domains with environment perturbations. In addition, we show improved robust performance on a high-dimensional, simulated, dexterous robotic hand. Finally, we present multiple investigative experiments that provide a deeper insight into the robustness framework. This includes an adaptation to another continuous control RL algorithm as well as learning the uncertainty set from offline data. Performance videos can be found online at \url{https://sites.google.com/view/robust-rl}.
\end{abstract}

\section{Introduction}
%What is the problem
Reinforcement Learning (RL) algorithms typically learn a \textit{policy} that optimizes for the expected return \citep{sutton98reinforcement}. That is, the policy aims to maximize the sum of future expected rewards that an agent accumulates in a particular task. This approach has yielded impressive results in recent years, including playing computer games with super human performance \citep{mnih2015human,Tessler2018}, multi-task RL \citep{Rusu2016PolicyD, Devin2017LearningMN, Teh2017DistralRM, Mankowitz2019, Riedmiller2018LearningBP} as well as solving complex continuous control robotic tasks \citep{Duan2016BenchmarkingDR, abdolmaleki2018maximum, Kalashnikov2018ScalableDR, Haarnoja2018SoftAA}.

The current crop of RL agents are typically trained in a single environment (usually a simulator). As a consequence, an issue that is faced by many of these agents is the sensitivity of the agent's policy to environment perturbations. Perturbing the dynamics of the environment during test time, which may include executing the policy in a real-world setting, can have a significant \textit{negative} impact on the performance of the agent \citep{andrychowicz2018learning,peng2018sim,derman2018soft,Dicastro2012,mankowitz2018learning}. 
This is because the training environment is not necessarily a very good model of the perturbations that an agent may \textit{actually} face, leading to potentially unwanted, sub-optimal behaviour. 
There are many types of environment perturbations. These include changing lighting/weather conditions, sensor noise, actuator noise, action delays etc \citep{Dulac2019}. 

%\todd{you could add that this is in contrast to supervised learning where there's separate train and test sets, here we train and test on the same env usually}
% RE-MPO
% E-MPO
% R-MPO

% Robust E MPO

% Find a domain where RE-MPO is significantly better than R-MPO. 
% State on average it is never worse and in some cases significanlty better - Add figure and reference figure in appendix

% We provide a generalized framework for incororating robustness in to continuous control algorithms for both the regular expetced return and entropy regularized objectives. Specifically, 

%Why is it important
It is desirable to train agents that are agnostic to environment perturbations. This is especially crucial in the Sim2Real setting \citep{andrychowicz2018learning,peng2018sim,wulfmeier2017mutual,rastogi2018sample,Christiano2016} where a policy is trained in a simulator and then executed on a real-world domain. As an example, consider a robotic arm that executes a control policy to perform a specific task in a factory. If, for some reason, the arm needs to be replaced and the specifications do not exactly match, then the control policy still needs to be able to perform the task with the `perturbed' robotic arm dynamics. In addition, sensor noise due to malfunctioning sensors, as well as actuator noise, may benefit from a robust policy to deal with these noise-induced perturbations.   

\textbf{Model misspecification}: For the purpose of this paper, we refer to an agent that is trained in one environment and performs in a different, perturbed version of the environment (as in the above examples) as \textit{model misspecification}. By incorporating robustness into our agents, we correct for this misspecification yielding improved performance in the perturbed environment(s).

% How do I propose to solve it?
In this paper, we propose a framework for incorporating robustness into continuous control RL algorithms. We specifically focus on robustness to model misspecification in the transition dynamics. Our main contributions are as follows: 

%For the remainder of the paper, when we mention robustness, we refer to this particular form of robustness. 

%We provide a generalized framework for incorporating robustness to model misspecification into continuous control RL algorithms. Specifically, algorithms that learn a value function (e.g., a critic) or perform policy evaluation. As a proof-of-concept, w
\textbf{(1)} We incorporate  robustness into a state-of-the-art continuous control RL algorithm called Maximum a-posteriori Policy Optimization (MPO) \citep{abdolmaleki2018maximum} to yield Robust MPO (R-MPO). We also carry out an additional experiment, where we incorporate  robustness into an additional continuous RL algorithm called Stochastic Value Gradients (SVG) \citep{heess2015learning}.

\textbf{(2)} Entropy regularization encourages exploration and helps prevent early convergence to sub-optimal policies \citep{nachum2017bridging}. To incorporate these advantages, we: (i) Extend the Robust Bellman operator \citep{iyengar2005robust} to  robust and soft-robust entropy-regularized versions, and show that these operators are contraction mappings. In addition, we (ii) extend MPO to \textit{Robust} Entropy-regularized MPO (RE-MPO) and  Soft RE-MPO (SRE-MPO) and show that they perform at least as well as R-MPO and in some cases significantly better. All the derivations have been deferred to Appendices \ref{app:rempo-bo}, \ref{app:srempo-bo} and \ref{app:rempo-pe}. 

We want to emphasize that, while the theoretical contributions are novel, our most significant contribution is that of the extensive experimental analysis we have performed to analyze the robustness performance of our agent. Specifically:

\textbf{(3)} We present experimental results in nine Mujoco domains showing that RE-MPO, SRE-MPO  and R-MPO, SR-MPO outperform both E-MPO and MPO respectively.

\textbf{(4)} To ensure that our method scales, we show robust performance on a high-dimensional, simulated, dexterous robotic hand called Shadow hand which outperforms the non-robust MPO baseline.

\textbf{(5)} Multiple investigative experiments to better understand the robustness framework. These include (i) an analysis of modifying the uncertainty set; (ii) comparing our technique to data augmentation; (iii) a comparison to domain randomization; (iv) comparing with and without entropy regularization; (v) We also train the transition models from offline data and use them as the uncertainty set to run R-MPO. We show that R-MPO with \textit{learned} transition models as the uncertainty set can lead to improved performance over R-MPO. 

\vspace{-0.3cm}
\section{Background}
\label{sec:background}
% Define the MDP

\textbf{A Markov Decision Process (MDP)} is defined as the tuple $\langle S, A, r, \gamma, P \rangle$ where $S$ is the state space, $A$ the action space, $r:S\times A \rightarrow \mathbb{R}$ is a bounded reward function; $\gamma \in [0,1]$ is the discount factor and $P:S\times A \rightarrow \Delta^S$ maps state-action pairs to a probability distribution over next states. We use $\Delta^S$ to denote the $|S|-1$ simplex. The goal of a Reinforcement Learning agent for the purpose of control is to learn a policy $\pi:S \rightarrow \Delta^A$ which maps a state and action to a probability of executing the action from the given state so as to maximize the expected return $J(\pi)=\mathbb{E}^{\pi} [\sum_{t=0}^\infty \gamma^t r_t]$ where $r_t$ is a random variable representing the reward received at time $t$ \citep{sutton2018reinforcement}.  The value function is defined as $V^{\pi}(s) = \mathbb{E}^{\pi}[\sum_{t=0}^{\infty} \gamma^t r_t | s_{0}=s]$ and the action value function as $Q^{\pi}(s,a) = r(s,a) + \gamma \mathbb{E}_{s' \sim P(\cdot | s,a)}[V^{\pi}(s')]$.

%\nir{why does the policy depend on the actions, what is $R$? why isn't it $\pi:S \rightarrow \Delta^A$?}

% Robust MDPs

\textbf{A Robust MDP (R-MDP)} is defined as a tuple $\langle S, A, r, \gamma, \mathcal{P} \rangle$ where $S,A,r$ and $\gamma$ are defined as above; $\mathcal{P}(s,a) \subseteq \mathcal{M}(S)$ is an uncertainty set where $\mathcal{M}(S)$ is the set of probability measures over next states $s' \in S$. This is interpreted as an agent selecting a state and action pair, and the next state $s'$ is determined by a conditional measure $p(s' \vert s, a) \in \mathcal{P}(s,a)$ \citep{iyengar2005robust}. A robust policy optimizes for the worst-case expected return objective: $J_{\text{R}}(\pi)=\inf_{p \in \mathcal{P}}\mathbb{E}^{p,\pi} [\sum_{t=0}^\infty \gamma^t r_t ]$.
%\todd{I was a bit confused by this notation. M is defined on $s_{t+1}$? Doesn't it need to be a function of $s,a$?}

%\nir{$\mathcal{P}$ is not time-dependent} \nir{how about simply $\mathcal{P}(s,a) \subseteq \mathcal{M}(S)$ is an uncertainty set where $\mathcal{M}(S)$ is the set of probability measures over states.} - P is time dependent in the original robustness formulation. We make an assumption in our experiments that P is not time dependent.

The robust value function is defined as $V_{\text{R}}^{\pi}(s) = \inf_{p\in \mathcal{P}}\mathbb{E}^{p,\pi}[\sum_{t=0}^{\infty} \gamma^t r_t | s_{0}=s]$ and the robust action value function as $Q^{\pi}_{\text{R}}(s,a) = r(s,a) + \gamma \inf_{p\in \mathcal{P}} \mathbb{E}_{s' \sim p(\cdot | s,a)}[V^{\pi}_{\text{R}}(s')]$. Both the robust Bellman operator $T_{\text{R}}^\pi: \mathcal{R}^{|S|} \rightarrow \mathcal{R}^{|S|}$ for a fixed policy and the optimal robust Bellman operator $T_{\text{R}}v(s) = \max_\pi T_{\text{R}}^\pi v(s)$ have previously been shown to be contractions \citep{iyengar2005robust}. A rectangularity assumption on the uncertainty set \citep{iyengar2005robust} ensures that ``nature'' can choose a worst-case transition function independently for every state $s$ and action $a$.
%\nir{should we explicitly present the robust Bellman operator?} No, it will add a bunch of unnecessary notation.

%\nir{how about saying something along: for ease of notation, from now on we drop some specifics whenever the intent is clear from the context?}

\textbf{Maximum A-Posteriori Policy Optimization (MPO)} \citep{abbas2018a,abdolmaleki2018maximum} is a continuous control RL algorithm that performs an expectation maximization form of policy iteration. There are two steps comprising \textbf{policy evaluation} and \textbf{policy improvement}. The \textit{policy evaluation} step receives as input a policy $\pi_k$ and evaluates an action-value function $Q^{\pi_k}_\theta(s,a)$ by minimizing the squared TD error: 
    $\min_{\theta} (r_t + \gamma Q_{\hat{\theta}}^{\pi_{k}}(s_{t+1} \sim P(\cdot | s_t, a_t), a_{t+1}\sim \pi_{k}(\cdot \vert s_{t+1})) - Q_{\theta}^{\pi_{k}}(s_t, a_t) )^2$, 
where $\hat{\theta}$ denotes the parameters of a target network \citep{mnih2015human} that are periodically updated from $\theta$. In practice we use a replay-buffer of samples in order to perform the policy evaluation step. The second step comprises a policy improvement step. The \textit{policy improvement} step consists of optimizing the objective $\bar{J}(s,\pi) = \mathbb{E}_\pi[Q_\theta^{\pi_{k}}(s,a)]$ for states $s$ drawn from a state distribution $\mu(s)$. In practice the state distribution samples are drawn from an experience replay. By improving $\bar{J}$ in all states $s$, we improve our objective. To do so, a two step procedure is performed. 

First, we construct a non-parametric estimate $q$ such that $\bar{J}(s,q) \geq \bar{J}(s,\pi_k)$. This is done by maximizing $\bar{J}(s,q)$ while ensuring that the solution, locally, stays close to the current policy $\pi_k$; i.e. $\mathbb{E}_{\mu(s)}[\text{KL}(q(\cdot|s), \pi_{k}(\cdot|s))] < \epsilon$. This optimization has a closed form solution given as $q(a\vert s)\propto \pi_k(a \vert s) \exp{\nicefrac{Q_{\theta}^{\pi_k}(s,a)}{\eta}},$ where $\eta$ is a temperature parameter that can be computed by minimizing a convex dual function (\cite{abdolmaleki2018maximum}). 
Second, we project this non-parametric representation back onto a parameterized policy by solving the optimization problem 
${\pi_{k+1} = \arg\min_{\pi} \mathbb{E}_{\mu(s)}[\text{KL}(q(a|s)\Vert \pi(a\vert s)]}$, where $\pi_{k+1}$ is the new and improved policy and where one typically employs additional regularization \citep{abbas2018a}. Note that this amounts to supervised learning with samples drawn fron $q(a|s)$; see \citet{abbas2018a} for details.

\section{Robust MPO}
To incorporate robustness into MPO, we focus on learning a worst-case value function in the policy evaluation step. Note that this policy evaluation step can be incorporated into any actor-critic algorithm. In particular, instead of optimizing the squared TD error, we optimize the worst-case squared TD error, which is defined as:
\vspace{-0.1cm}
\begin{equation}
\min_{\theta} \biggl(r_t + \gamma \inf_{p \in \mathcal{P}(s_t, a_t)} \biggl[Q_{\hat{\theta}}^{\pi_{k}}(s_{t+1} \sim p(\cdot | s_t, a_t), a_{t+1}\sim \pi_{k}( \cdot \vert s_{t+1}))  \biggr] - Q_{\theta}^{\pi_{k}}(s_t, a_t) \biggr)^2 \enspace ,
\label{eqn:rmpotd}
\end{equation}

%\nir{too many appearances of $p$, let's talk about it - in general i don't like $\mathcal{P}(s,a)$, and think we should just do $\mathcal{P}$.} - we can talk about it but it is notationally accurate.

where $\mathcal{P}(s_t,a_t)$ is an uncertainty set for the current state $s_t$ and action $a_t$; $\pi_{k}$ is the current network's policy, and $\hat{\theta}$ denotes the target network parameters. It is in this policy evaluation step (Line 3 in Algorithms \ref{alg:1},\ref{alg:2} and \ref{alg:3} in Appendix \ref{app:algorithm}) that the Bellman operators in the previous sections are applied.

%This policy evaluation step is equivalent to applying the robust Bellman operator for a policy $\pi_k$ to the current estimate of the action value function. 

\textbf{Relation to MPO:} In MPO, this replaces the current policy evaluation step. The robust Bellman operator \citep{iyengar2005robust} ensures that this process converges to a unique fixed point for the policy $\pi_k$. This is achieved by repeated application of the robust Bellman operator during the policy evaluation step until convergence to the fixed point. Since the proposal policy $q(a|s)$ (see Section \ref{sec:background}) is proportional to the robust action value estimate $Q^{\pi_k}_\theta(s,a)$, it intuitively yields a robust policy as the policy is being generated from a worst-case value function. The fitting of the policy network to the proposal policy yields a robust network policy $\pi_{k+1}$.  

\textbf{Entropy-regularized MPO:} Entropy-regularization encourages exploration and helps prevent early convergence to sub-optimal policies \citep{nachum2017bridging}. To incorporate these advantages, we extended the Robust Bellman operator \citep{iyengar2005robust} to  robust and soft-robust entropy-regularized versions (See Appendix \ref{app:rempo-bo} and \ref{app:srempo-bo} respectively for a detailed overview and the corresponding derivations) and show that these operators are contraction mappings (Theorem \ref{one} below and Theorem \ref{app:one} in Appendix \ref{app:proofs}) and yield a well-known value-iteration bound with respect to the max norm. 

\begin{customthm}{1}
The robust \textbf{entropy-regularized} Bellman operator $\mathcal{T}^\pi_{\text{R-KL}}$ for a fixed policy $\pi$ is a contraction operator. Specifically:  $\forall U,V \in \mathbb{R}^{|S|}$ and $\gamma \in \left(0,1\right)$, we have, $\Vert \mathcal{T}^\pi_{\text{R-KL}}U - \mathcal{T}^\pi_{\text{R-KL}}V \Vert \leq \gamma \Vert U - V \Vert
$.

\label{one}
%\label{thm:contraction}
\end{customthm}

In addition, we extended MPO to \textit{Robust} Entropy-regularized MPO (RE-MPO) and  Soft RE-MPO (SRE-MPO) (see Appendix \ref{app:rempo-pe} for a detailed overview and derivations) and show that they perform at least as well as R-MPO and in some cases significantly better. All the derivations have been deferred to the Appendix. The corresponding algorithms for R-MPO, RE-MPO and SRE-MPO can be found in Appendix \ref{app:algorithm}.

\section{Experiments}
We now present experiments on nine different continuous control domains (four of which we show in the paper and the rest can be found in Appendix \ref{app:mainexperiments}) from the DeepMind control suite \citep{Tassa2018}. In addition, we present an experiment on a high-dimensional dexterous, robotic hand called Shadow hand \citep{Shadow2005}. In our experiments, we found that the entropy-regularized version of Robust MPO had similar performance and in some cases, slightly better performance than the expected return version of Robust MPO \textit{without} entropy-regularization. We therefore decided to include experiments of our agent optimizing the entropy-regularized objective (non-robust, robust and soft-robust versions). This corresponds to (a) non-robust E-MPO baseline, (b) Robust E-MPO (RE-MPO) and (c) Soft-Robust E-MPO (SRE-MPO). From hereon in, it is assumed that the algorithms optimize for the entropy-regularized objective unless otherwise stated.

\textit{Appendix}: In Appendix \ref{app:mainexperiments}, we present results of our agent optimizing for the expected return objective \textit{without} entropy regularization   (for the non-robust, robust and soft-robust versions). This corresponds to (a') non-robust MPO baseline, (b') R-MPO and (c') SR-MPO.

The experiments are divided into three sections. The first section details the setup for robust and soft-robust training. The next section compares robust and soft-robust performance to the non-robust MPO baseline in each of the domains. The final section is a set of investigative experiments to gain additional insights into the performance of the robust and soft-robust agents. 
%This includes incorporating robustness into the Stochastic Value Gradients (SVG) algorithm \citep{heess2015learning}.

%\nir{is there a way to motivate the test set to not look completely cherry-picked? maybe something from different papers that show that it's a common practice?}

\paragraph{Setup:} For each domain, the robust agent is trained using a pre-defined uncertainty set consisting of three task perturbations \footnote{We did experiments on a larger set with similar results, but settled on three for computational efficiency.}. Each of the three perturbations corresponds to a particular perturbation of the Mujoco domain. For example, in Cartpole, the uncertainty set consists of three different pole lengths. Both the robust and non-robust agents are evaluated on a test set of three unseen task perturbations. In the Cartpole example, this would correspond to pole lengths that the agent has not seen during training. The chosen values of the uncertainty set and evaluation set for each domain can be found in Appendix \ref{app:hyperparameters}. Note that it is common practice to manually select the pre-defined uncertainty set and the unseen test environments. Practitioners often have significant domain knowledge and can utilize this when choosing the uncertainty set \citep{derman2019,derman2018soft,Dicastro2012,mankowitz2018learning,tamar2014scaling}.  
%\todd{can we say anything about why this is still robust? Anecdotally say that the agent works well with values in some range?}

%\todd{it feels like an easy criticism of the paper will be that the exact perturbation values were cherry-picked. Instead of saying we chose 3 values, can we instead sample them from some distribution, with new samples for each seed? Shows more robustness to hyper selection too}

During training, the robust, soft-robust and non-robust agents act in an unperturbed environment which we refer to as the \textit{nominal} environment. During the TD learning update, the robust agent calculates an infimum between Q values from each next state realization for each of the uncertainty set task perturbations (the soft-robust agent computes an average, which corresponds to a uniform distribution over $\mathcal{P}$, instead of an infimum). Each transition model is a different instantiation of the Mujoco task. The robust and soft-robust agents are exposed to more state realizations than the non-robust agent. However, as we show in our ablation studies, significantly increasing the number of samples and the diversity of the samples for the non-robust agent still results in poor performance compared to the robust and soft-robust agents. 

%\nir{should we say something stronger like that it actually doesn't improve the performance at all? becuase this sounds like the nonrobust agent still improves..} - it does improve a bit for domain randomization

%\todd{we should make it clear that the uncertainty set are more instantiations of mujoco, not learned transition models} \nir{I think it's worth making tis point stronger, as this is very important to the amount of additional knowledge the robust agent ic exposed to. Specifically, the other environments only supply the next state on which we take infimum - or minimum since it's discrete}

\subsection{Main Experiments}
\begin{figure*}
\centering
\newcommand{\scl}{0.2}
 \subfigure{
 \includegraphics[scale=\scl]{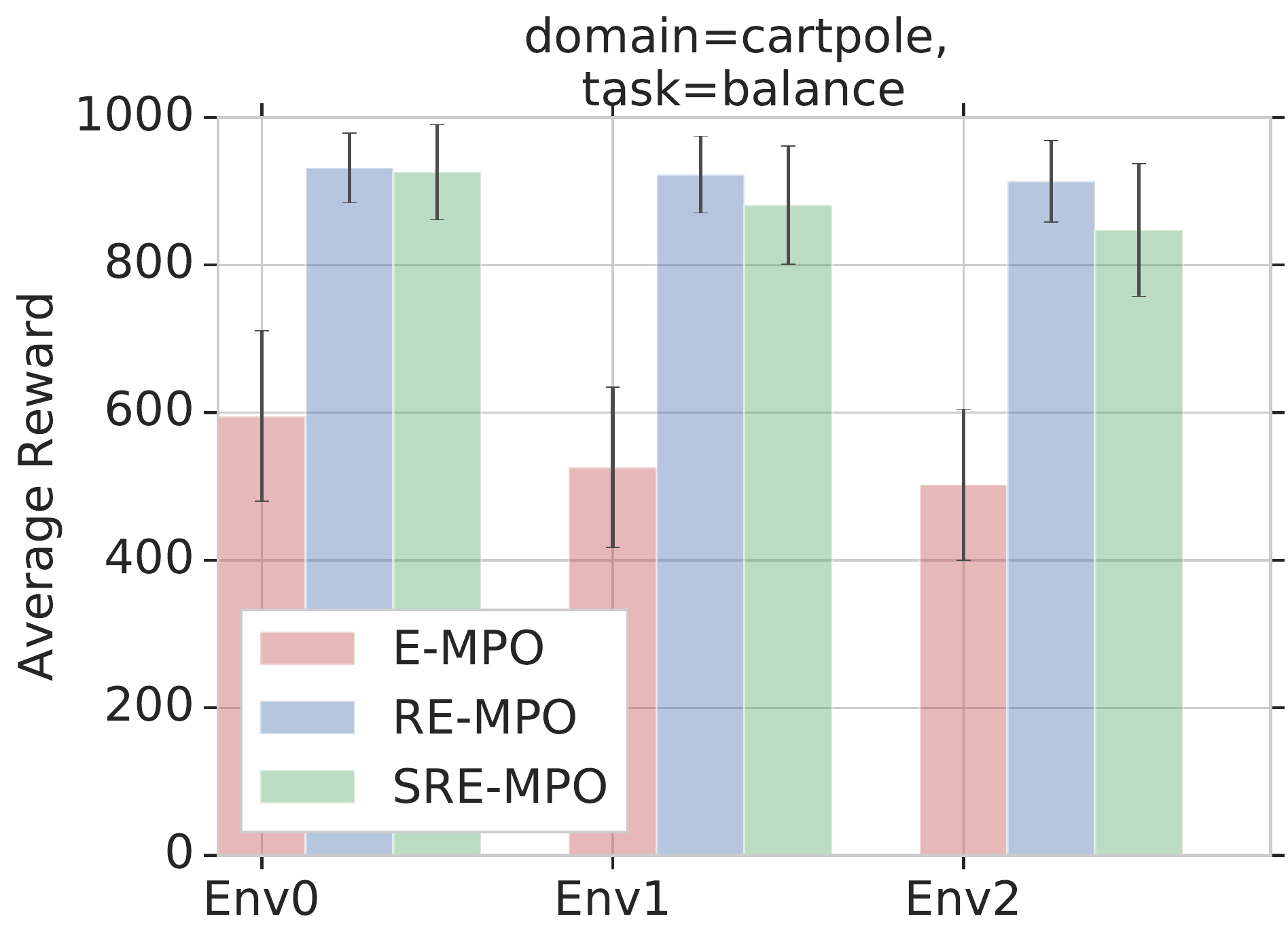}
 }
%   \subfigure{
%  \includegraphics[scale=\scl]{./figures/mpo_kl/pendulum_swingup}
%  }
  \subfigure{
 \includegraphics[scale=\scl]{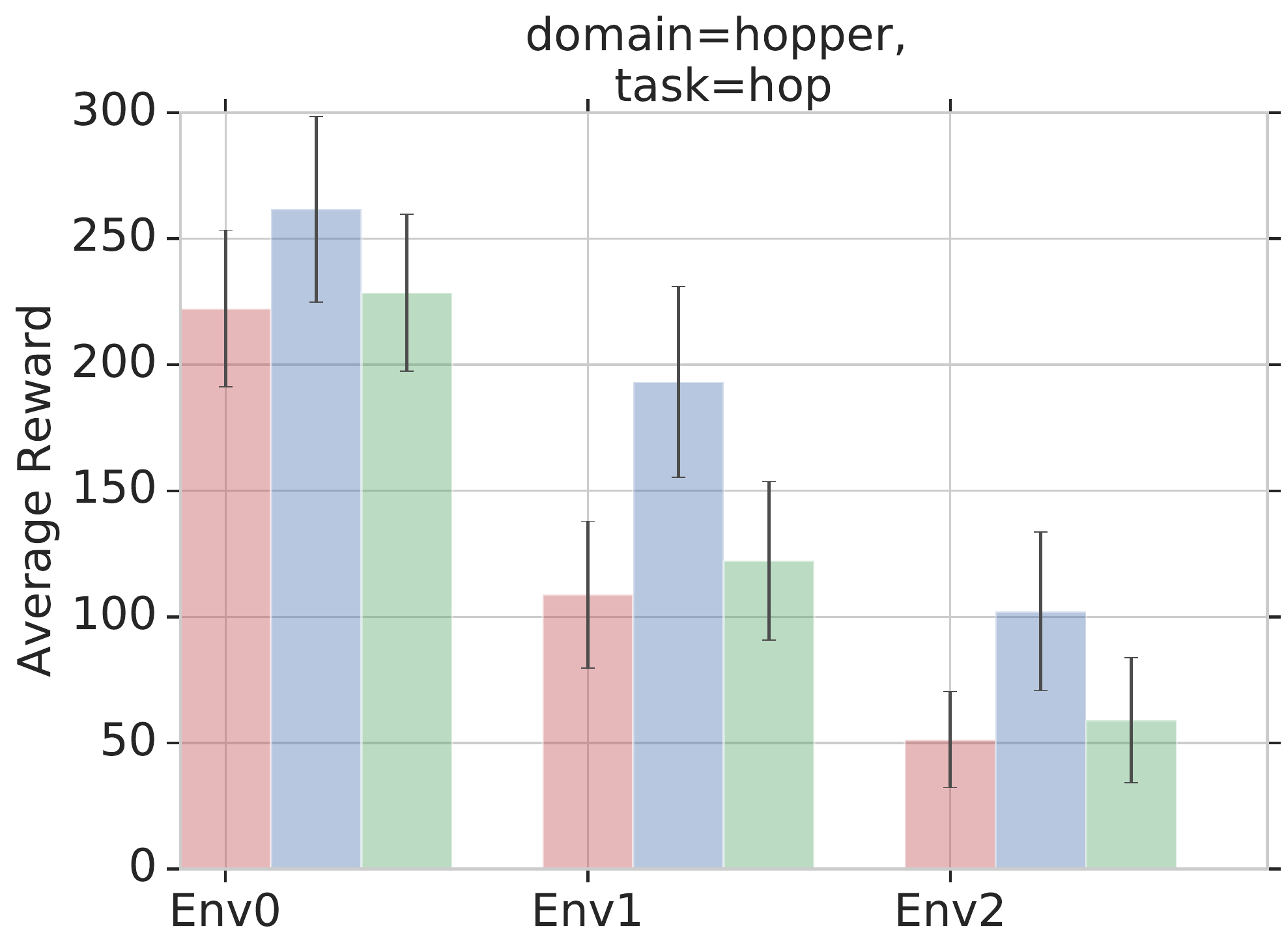}
 }
 \subfigure{
 \includegraphics[scale=\scl]{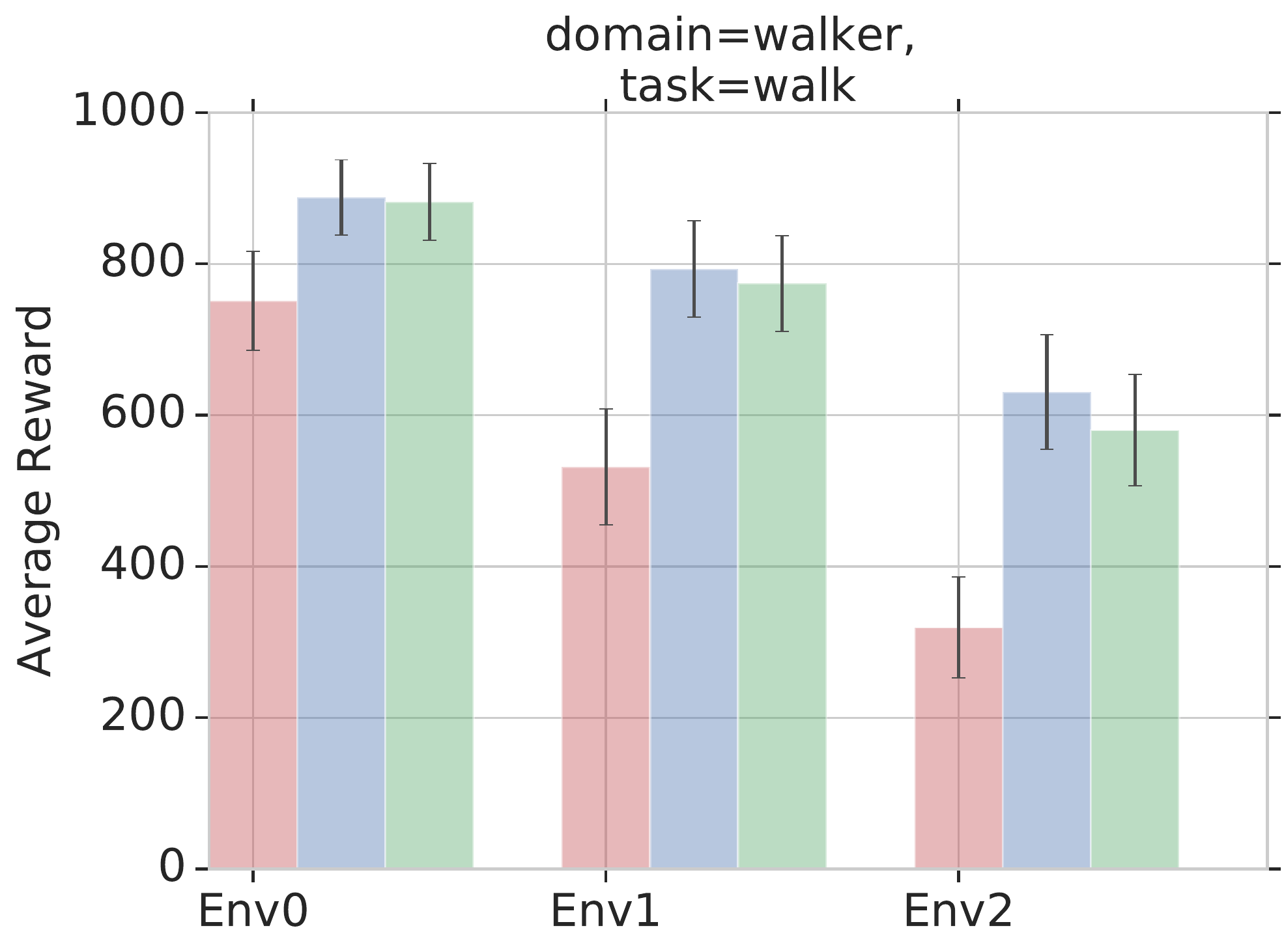}
 }
%  \subfigure{
%  \includegraphics[scale=0.16]{./figures/mpo_kl/shadowhand}
%  }
%   \subfigure{
%  \includegraphics[scale=\scl]{./figures/mpo_kl/shadowhand_orientate_shadowhand}
%  }

 \vspace{-0.4cm}
\caption{
Three domains showing RE-MPO (blue), SRE-MPO (green) and E-MPO (red). The addition six domains can be found in the appendix. In addition, the results for R-MPO, SR-MPO and MPO can be found in Appendix \ref{app:mainexperiments} with similar results.}
\label{fig:mpo_kl_main}
\end{figure*}

\textbf{Mujoco Domains:} We compare the performance of non-robust MPO to the robust and soft-robust variants. Each training run consists of $30k$ episodes and the experiments are repeated $5$ times. In the bar plots, the y-axis indicates the average reward (with standard deviation) and the x-axis indicates different unseen evaluation environment perturbations starting from the first perturbation (Env0) onwards. Increasing environment indices correspond to increasingly large perturbations. For example, in Figure \ref{fig:mpo_kl_main} (top left), Env0, Env1 and Env2 for the Cartpole Balance task represents the pole perturbed to lengths of $2.0, 2.2$ and $2.3$ meters respectively. Figure \ref{fig:mpo_kl_main} shows the performance of three Mujoco domains (The remaining six domains are in Appendix \ref{app:mainexperiments}). The bar plots indicate the performance of E-MPO (red), RE-MPO (blue) and SRE-MPO (green) on the held-out test perturbations. This color scheme is consistent throughtout the experiments unless otherwise stated. As can be seen in each of the figures, RE-MPO attains improved performance over E-MPO. This same trend holds true for all nine domains. SRE-MPO outperforms the non-robust baseline in all but the Cheetah domain, but is not able to outperform RE-MPO. An interesting observation can be seen in the video for the Walker walk task (\url{https://sites.google.com/view/robust-rl}), where the RE-MPO agent learns to `drag' its leg which is a fundamentally different policy to that of the non-robust agent which learns a regular gait movement.

\textit{Appendix:} The appendix contains additional experiments with the \textit{non} entropy-regularized versions of the algorithms where again the robust (R-MPO) and soft robust (SR-MPO) versions of MPO outperform the non-robust version (MPO).

\begin{figure*}
\centering
\newcommand{\scl}{0.16}
\subfigure{
 \includegraphics[scale=0.14]{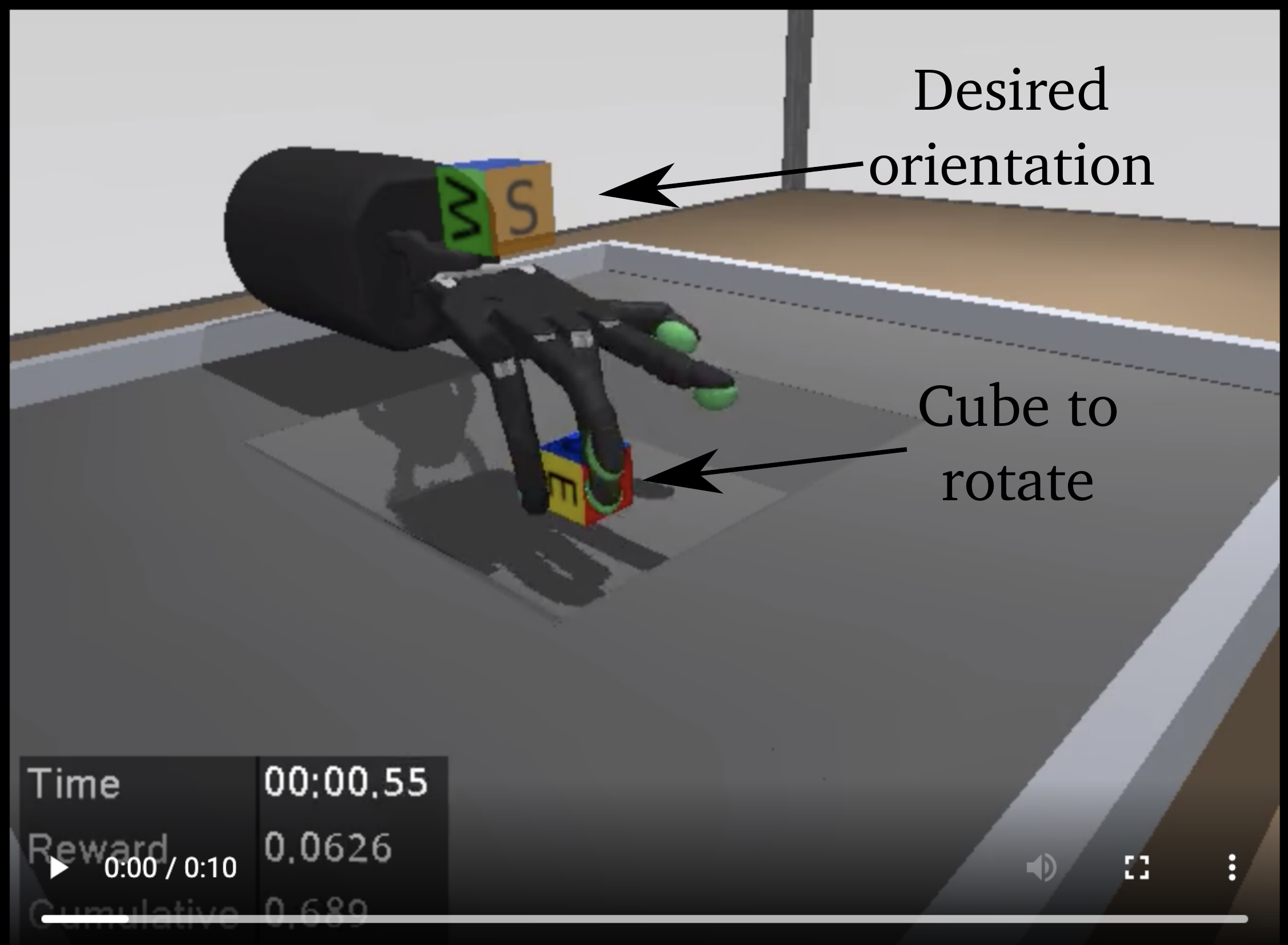}
 }
  \subfigure{
 \includegraphics[scale=\scl]{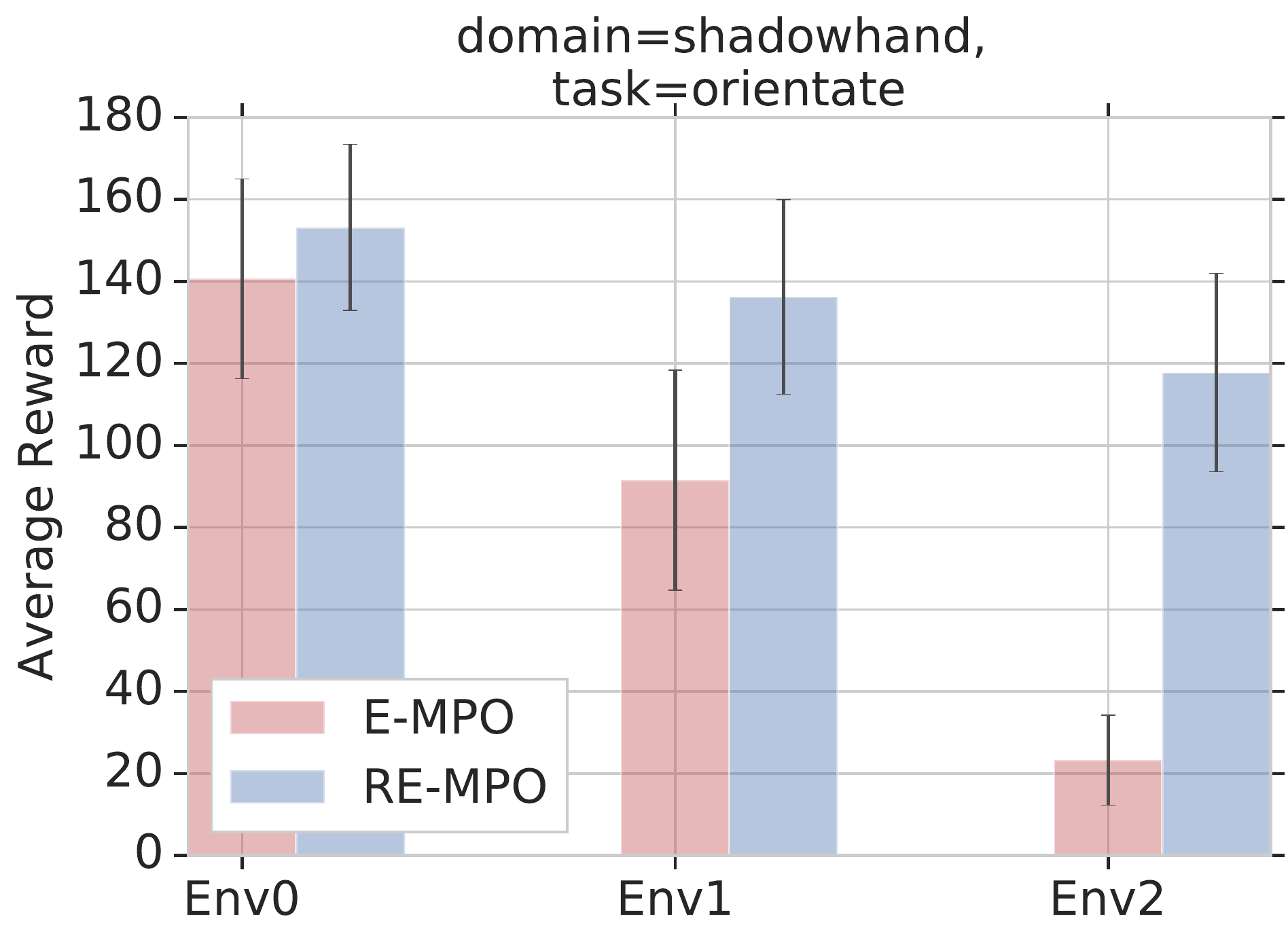}
 }
  \subfigure{
 \includegraphics[scale=\scl]{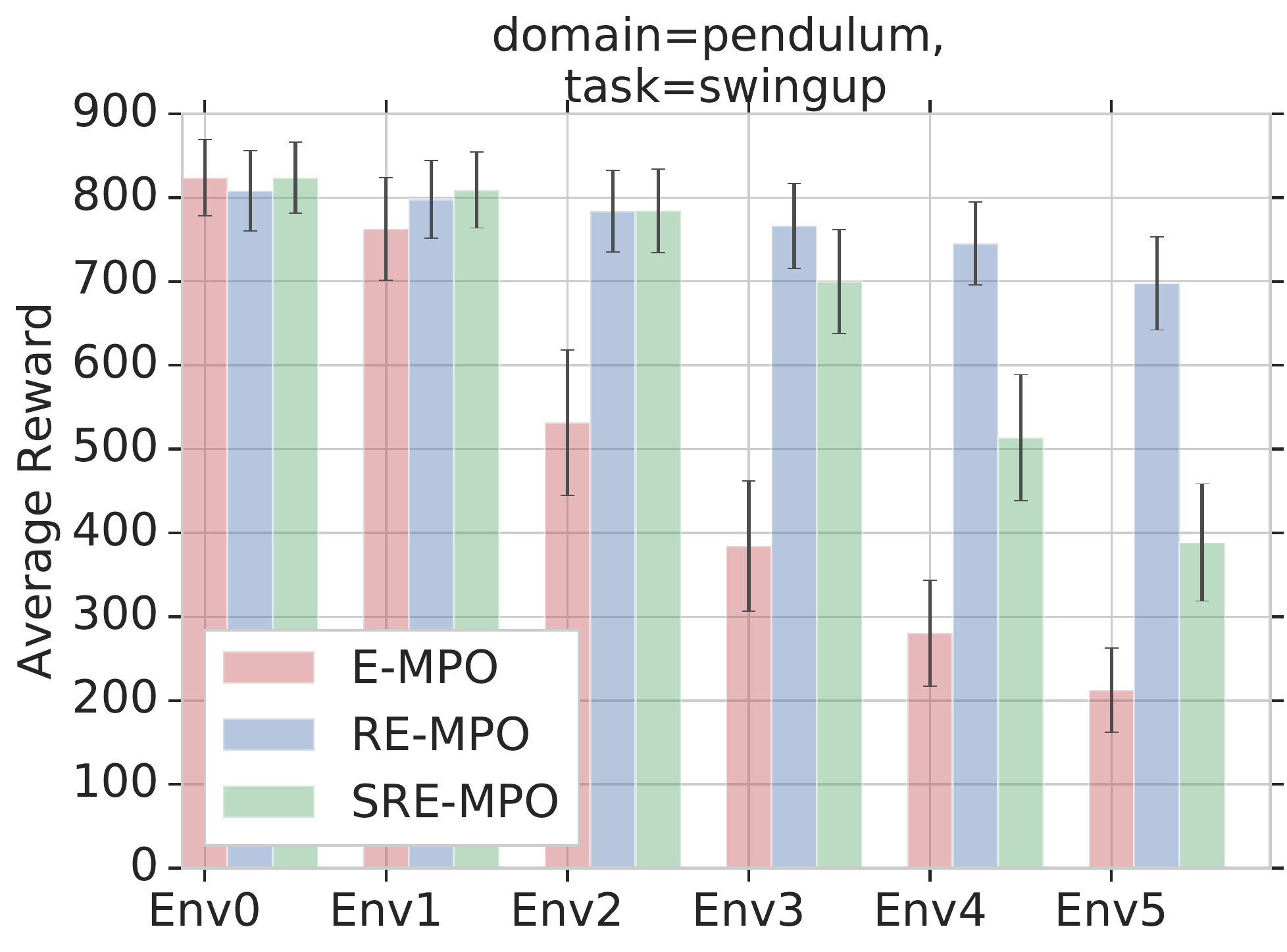}
 }
 \subfigure{
 \includegraphics[scale=\scl]{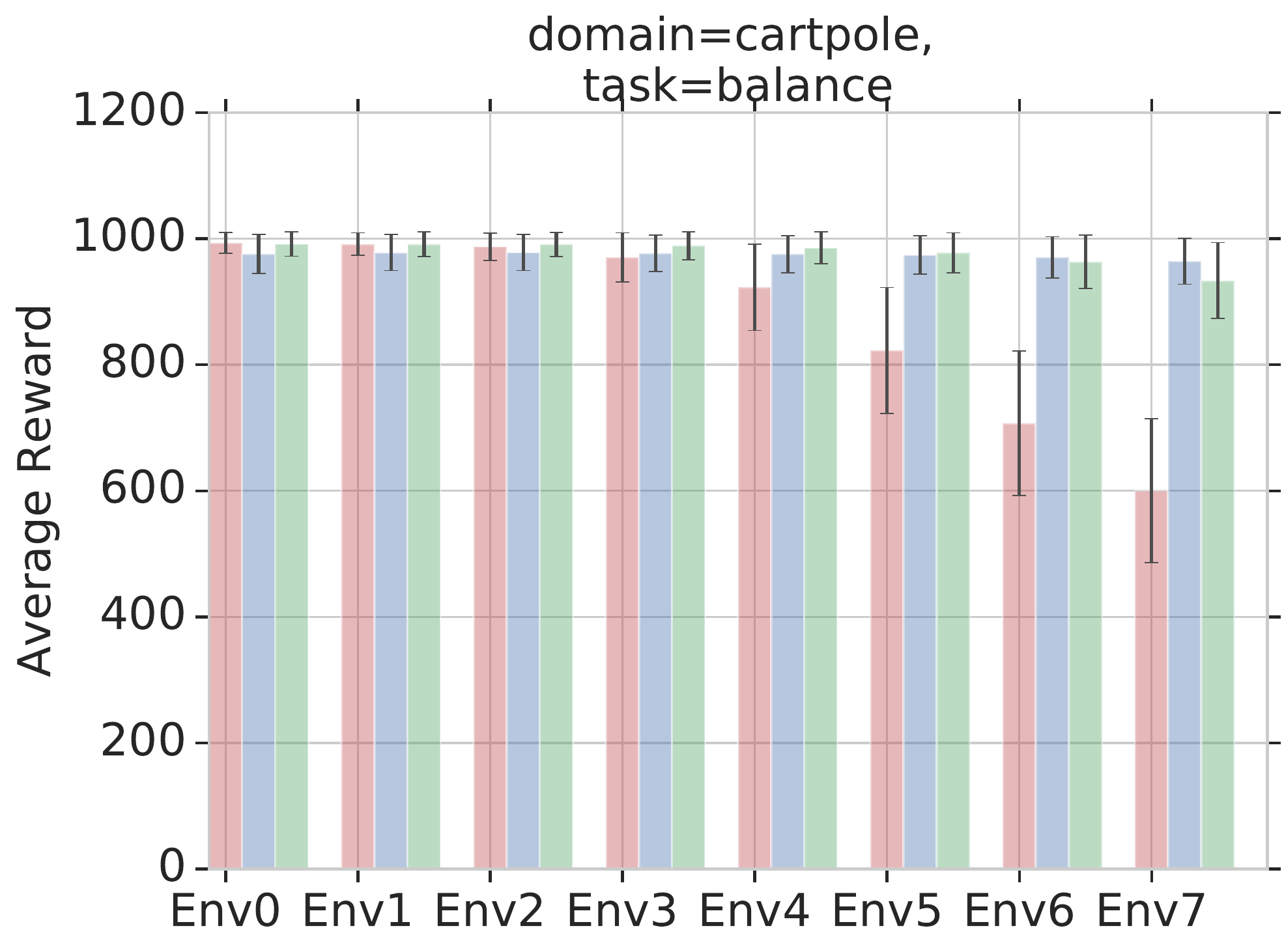}
 }
 \vspace{-0.3cm}
\caption{
(1) The Shadow hand domain (left) and results for RE-MPO and E-MPO (middle left). (2) A larger test set: the figures show the performance of RE-MPO (blue), SRE-MPO  (green) and E-MPO (red) for a test set that extends from the nominal environment to significant perturbations outside the training set for Cartpole Balance (middle right) and Pendulum Swingup (right).}
\label{fig:ablation1}
\end{figure*}

\textbf{Shadow hand domain:} This domain consists of a dexterous, simulated robotic hand called Shadow hand whose goal is to rotate a cube into a pre-defined orientation \citep{Shadow2005}. The state space is a $79$ dimensional vector and consisting of angular positions and velocities, the cube orientation and goal orientation. The action space is a $20$ dimensional vector and consisting of the desired angular velocity of the hand actuators. The reward is a function of the current orientation of the cube relative to the desired orientation. The uncertainty set consists of three models which correspond to increasingly smaller sizes of the cube that the agent needs to orientate. The agent is evaluated on a different, unseen holdout set. The values can be found in Appendix \ref{app:hyperparameters}. We compare RE-MPO to E-MPO trained agents. Episodes are $200$ steps long corresponding to approximately $10$ seconds of interaction. Each experiment is run for $6k$ episodes and is repeated $5$ times. As seen in Figure \ref{fig:ablation1}, RE-MPO outperforms E-MPO, especially as the size of the cube decreases (from Env0 to Env2). This is an especially challenging problem due to the high-dimensionality of the task. As seen in the videos (\url{https://sites.google.com/view/robust-rl}), the RE-MPO agent is able to manipulate significantly smaller cubes than it had observed in the nominal simulator.

%\todd{can we change the labels from saying env0, env1, env2, to what the perturbed values are?} - it is a bit difficult at this point.

%These results are consistent with the expected return objective variants: MPO, R-MPO and SR-MPO for each of the nine domains which can be seen in Appendix \ref{app:experiments}.

%\todd{we could only introduce the non entropy regularized experiments that are in the appendix here: The appendix contains additional experiments with the non entropy regularized versions of the algorithms, where again the robust and soft robust versions of MPO outperform the non-robust version.}
%\todd{anything we can say about entropy regularized vs not?} - we have an ablation study for this

% \todd{can we say anything about the amount of perturbation?} - we have tested it on more than 3 and it works. Probably results in a more robust solution, but cant say anything concrete

% \todd{if we're not going to plot the non-ER versions, we should at least say something about it}

% \todd{any pattern across envs? Is env2 more perturbed than env0?}
% \todd{I think we should really show results on the un-perturbed case too}
%}

\subsection{Investigative Experiments}
\label{sec:ablation}
This section aims to investigate and try answer various questions that may aid in explaining the performance of the robust and non-robust agents respectively. Each investigative experiment is conducted on the Cartpole Balance and Pendulum Swingup domains. 

\paragraph{What if we increase the number of training samples?} One argument is that the robust agent has access to more samples since it calculates the Bellman update using the infimum of three different environment realizations. To balance this is effect, the non-robust agent was trained for three times more episodes than the robust agents. Training with significantly more samples does not increase the performance of the non-robust agent and, can even decreases the performance, as a result of overfitting to the nominal domain. See  Appendix \ref{app:investigative}, Figure \ref{fig:app:90k} for the results.

\begin{figure*}
\centering
\newcommand{\scl}{0.16}
 \subfigure{
 \includegraphics[scale=\scl]{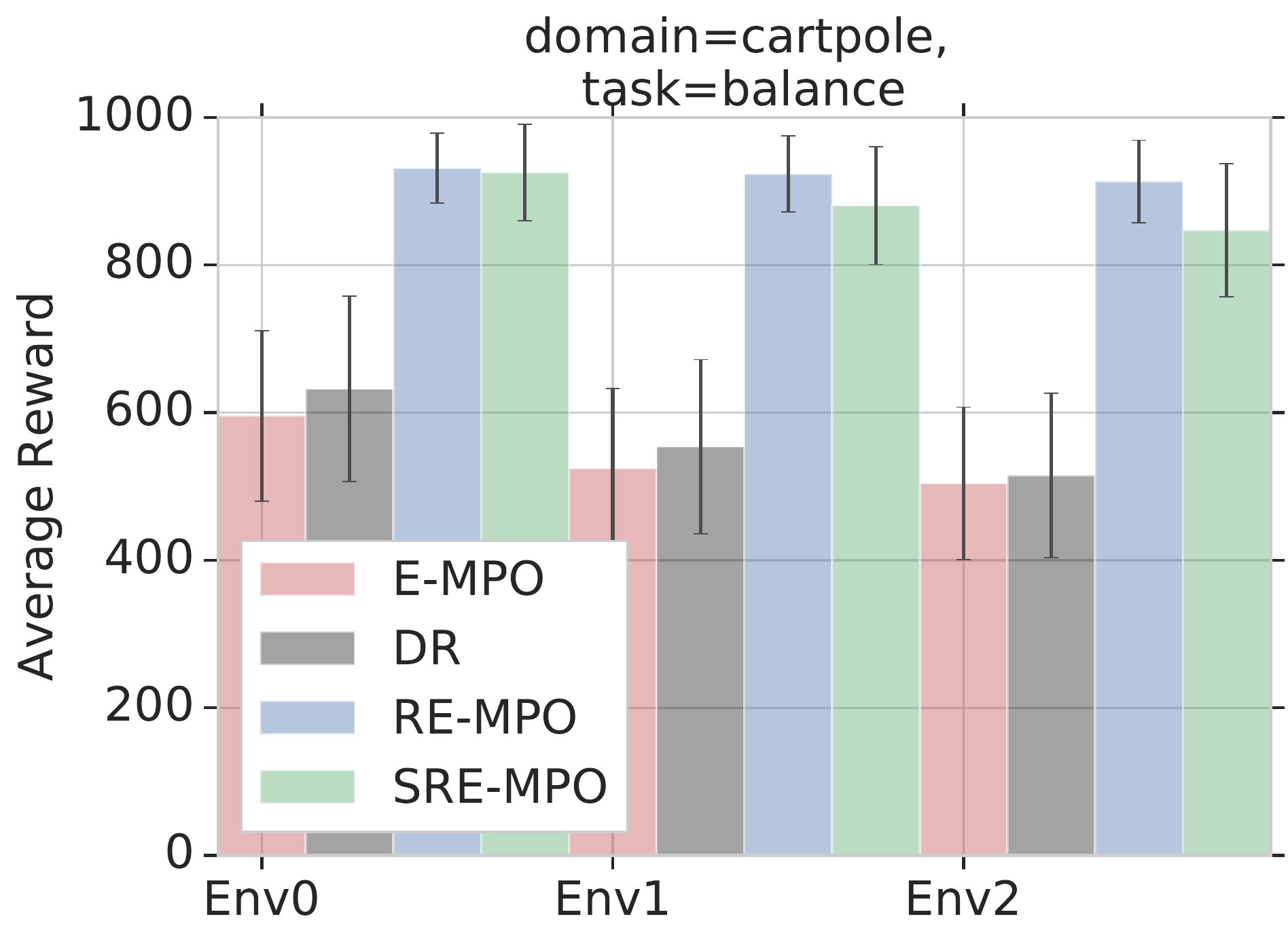}
 }
  \subfigure{
 \includegraphics[scale=\scl]{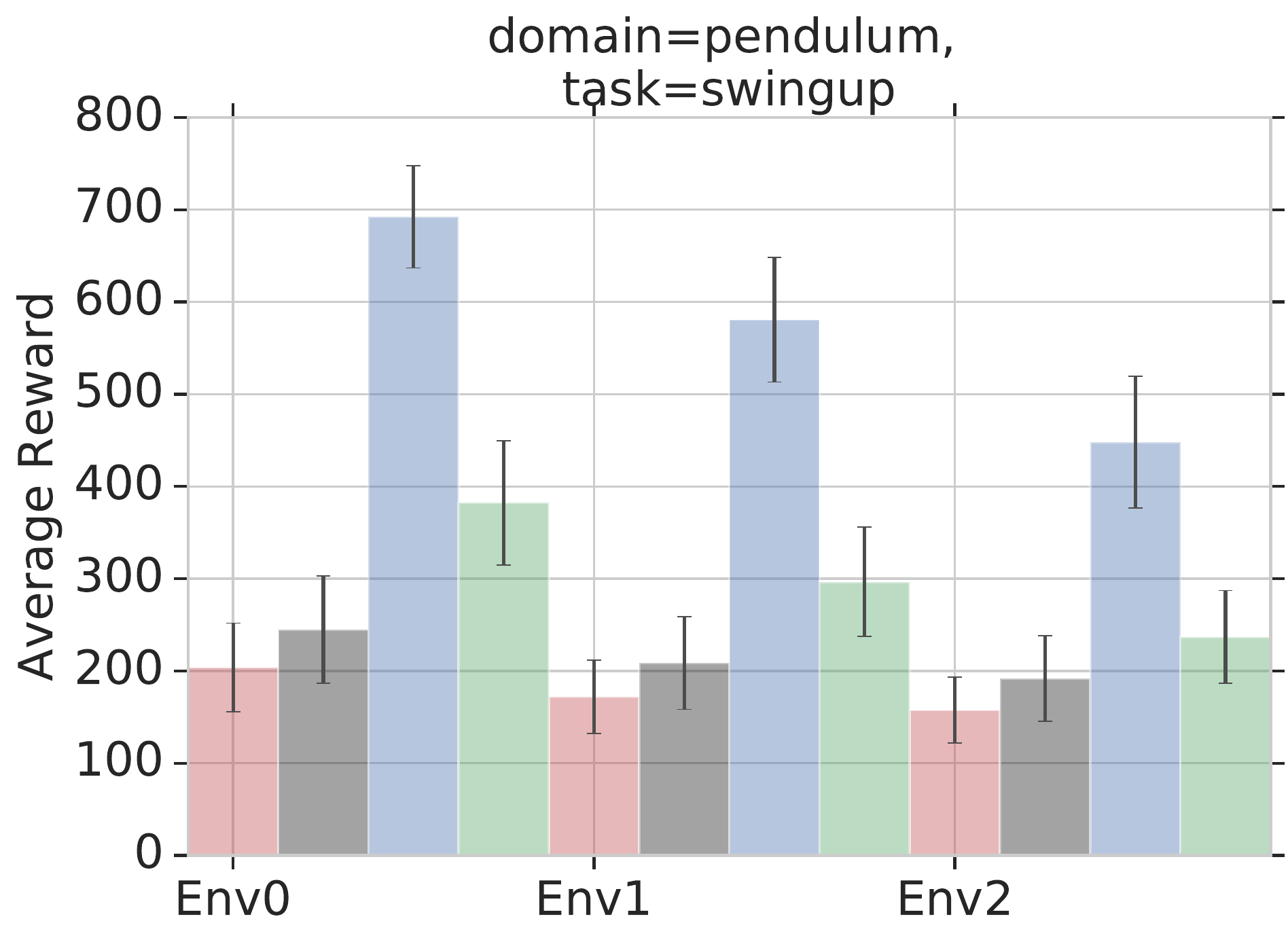}
 }
   \subfigure{
 \includegraphics[scale=\scl]{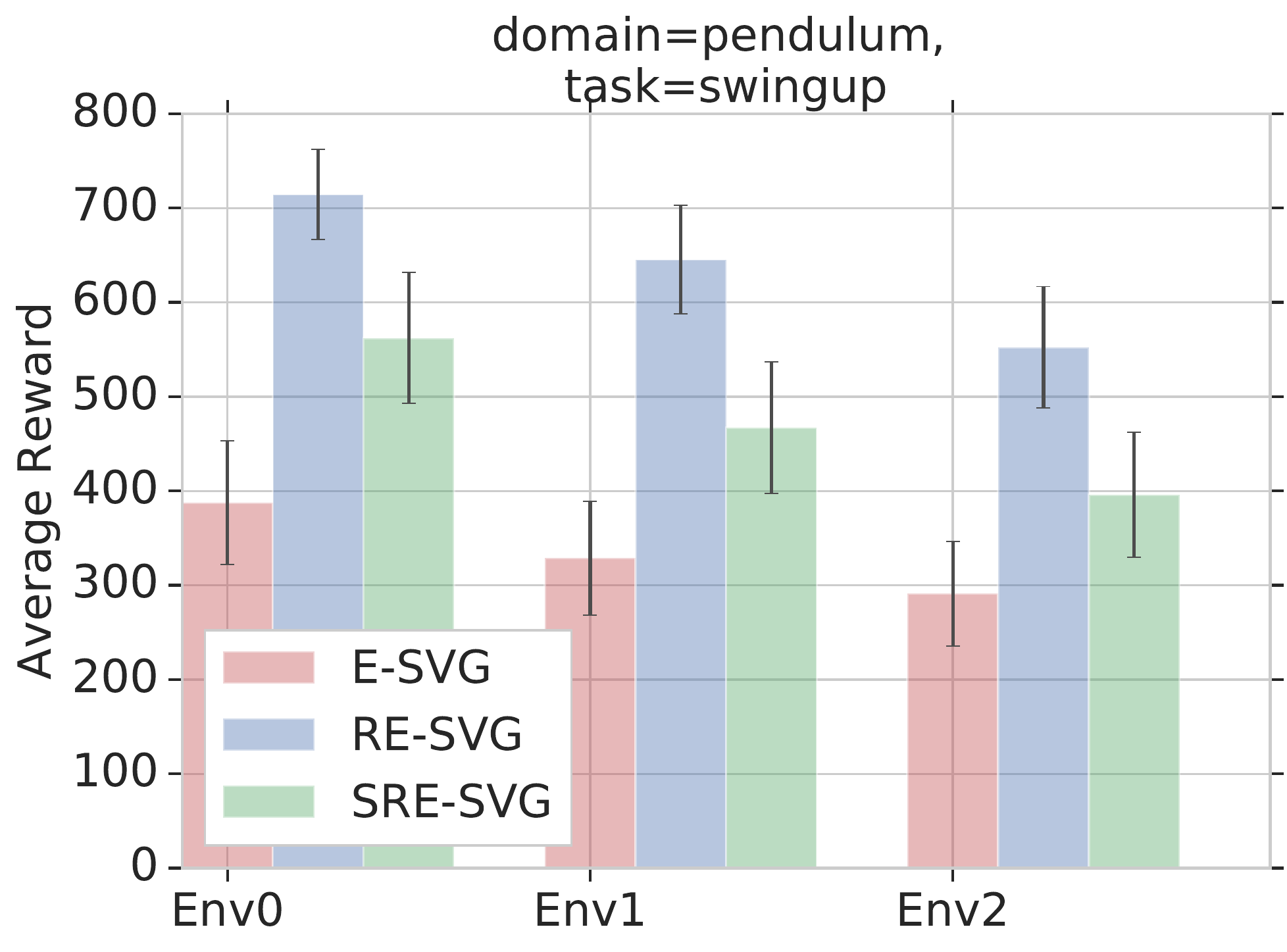}
 }
  \subfigure{
 \includegraphics[scale=\scl]{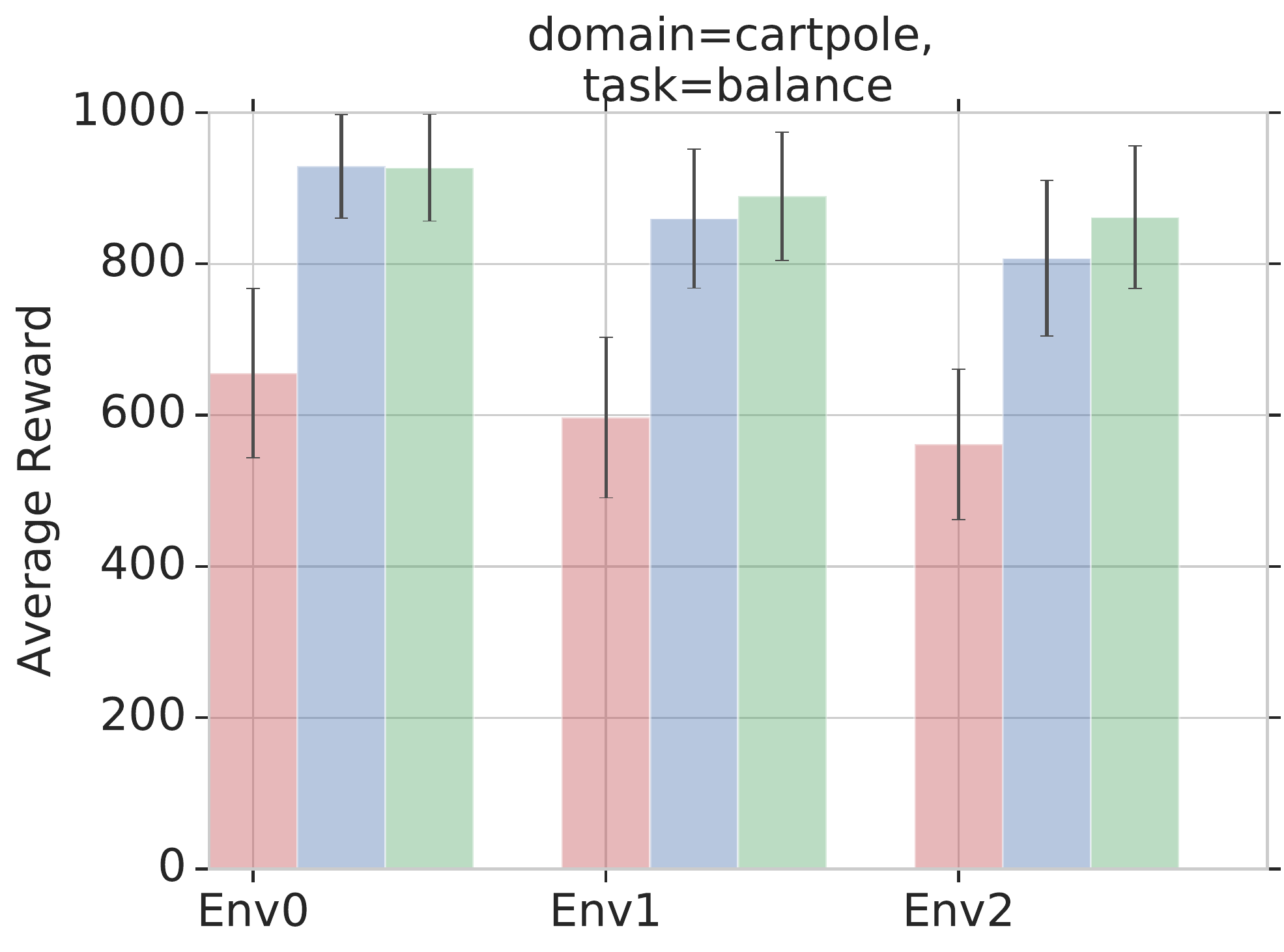}
 }
 \vspace{-0.3cm}
\caption{(1) Domain Randomization (DR): Domain randomization performance for the Cartpole balance (left) and Pendulum swingup (middle left) tasks. It should also be noted that our technique is learning a fundamentally different policy to domain randomization. See, for example, the Walker Walk task\url{https://sites.google.com/view/robust-rl}. (2) Stochastic Value Gradients (SVG): Two right images show the performance of Robust Entropy-regularized SVG (RE-SVG) and SRE-SVG compared to E-SVG for Pendulum and Cartpole respectively.}
\vspace{-0.5cm}
\label{fig:dr_svg}
\end{figure*}

\paragraph{What about Domain Randomization?} 
A subsequent point would be that the robust agent sees more diverse examples compared to the non-robust agent from each of the perturbed environments. We therefore trained the non-robust agent in a domain randomization setting \citep{andrychowicz2018learning,peng2018sim}. We compare our method to two variants of DR. The first variant \textit{Limited-DR} uses the same perturbations as in the uncertainty set of RE-MPO.  Here, we compare which method better utilizes a limited set of perturbations to learn a robust policy. As seen in Figure \ref{fig:dr_svg} (left and middle left for Carpole Balance and Pendulum Swingup respectively), RE-MPO yields significantly better performance given the limited set of perturbations. The second variant \textit{Full-DR} performs regular DR on a significantly larger set of $100$ perturbations in the Pendulum Swingup task. In this setting, DR, which uses $30$ times more perturbations, improves but still does not outperform RE-MPO (which still only uses three perturbations). This result can be seen in Figure~\ref{fig:ablation3}, Appendix~\ref{app:investigative}.  

%How is our technique different from DR from an empirical perspective?
%, you will see that the robust MPO agent learns to drag its leg in order to be robust to varying thigh lengths. This is a fundamentally different behaviour to that learned by MPO with DR that behaves similarly to the MPO baseline (also on website). In addition, since DR is a data augmentation technique, the agent learns the average behaviour across all the data. Robust MPO is explicitly trying to find adversarial examples by optimizing with respect to the worst case.

\textbf{What is the intuitive difference between DR and RE-MPO/SRE-MPO?}
\textbf{DR} defines the loss to be the expectation of TD-errors over the uncertainty set. Each TD error is computed using a state, action, reward, next state $<s,a,r,s’>$ trajectory from a particular perturbed environment, (selected uniformly from the uncertainty set). These TD errors are then averaged together. This is a form of data augmentation and the resulting behaviour is the average across all of this data. \textbf{RE-MPO/SRE-MPO}: In the case of robustness, the TD error is computed such that the \textit{target} action value function is computed as a worst case value function with respect to the uncertainty set. This means that the learned policy is explicitly searching for adversarial examples during training to account for worst-case performance. In the soft-robust case, the subtle yet important difference (as seen in the experiments) with DR is that the TD loss is computed with the average \textit{target} action value function with respect to next states (as opposed to averaging the TD errors of each individual perturbed environment as in DR). This results in different gradient updates being used to update the action value function compared to DR.

\paragraph{A larger test set:} It is also useful to view the performance of the agent from the nominal environment to increasingly large perturbations in the unseen test set (see Appendix \ref{app:hyperparameters} for values). These graphs can be seen in Figure \ref{fig:ablation1} for Cartpole Balance and Pendulum Swingup respectively. As expected, the robust agent maintains a higher level of performance compared to the non-robust agent. Initially, the soft-robust agent outperforms the robust agent, but its performance degrades as the perturbations increase which is consistent with the results of \cite{derman2018soft}. In addition, the robust and soft-robust agents are competitive with the non-robust agent in the nominal environment. 

\begin{figure*}
\centering
\newcommand{\scl}{0.16}
  \subfigure{
 \includegraphics[scale=\scl]{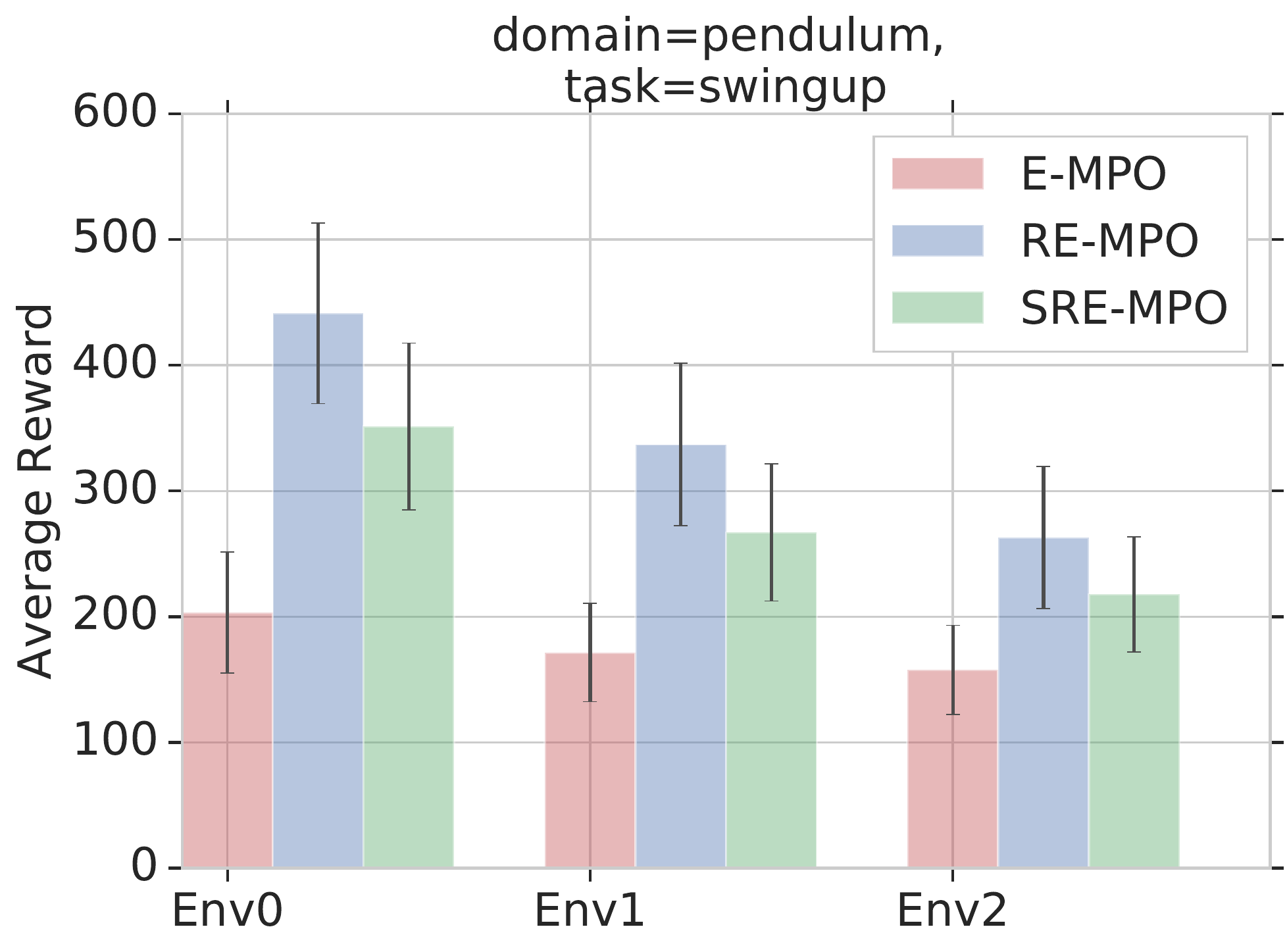}
 }
  \subfigure{
 \includegraphics[scale=\scl]{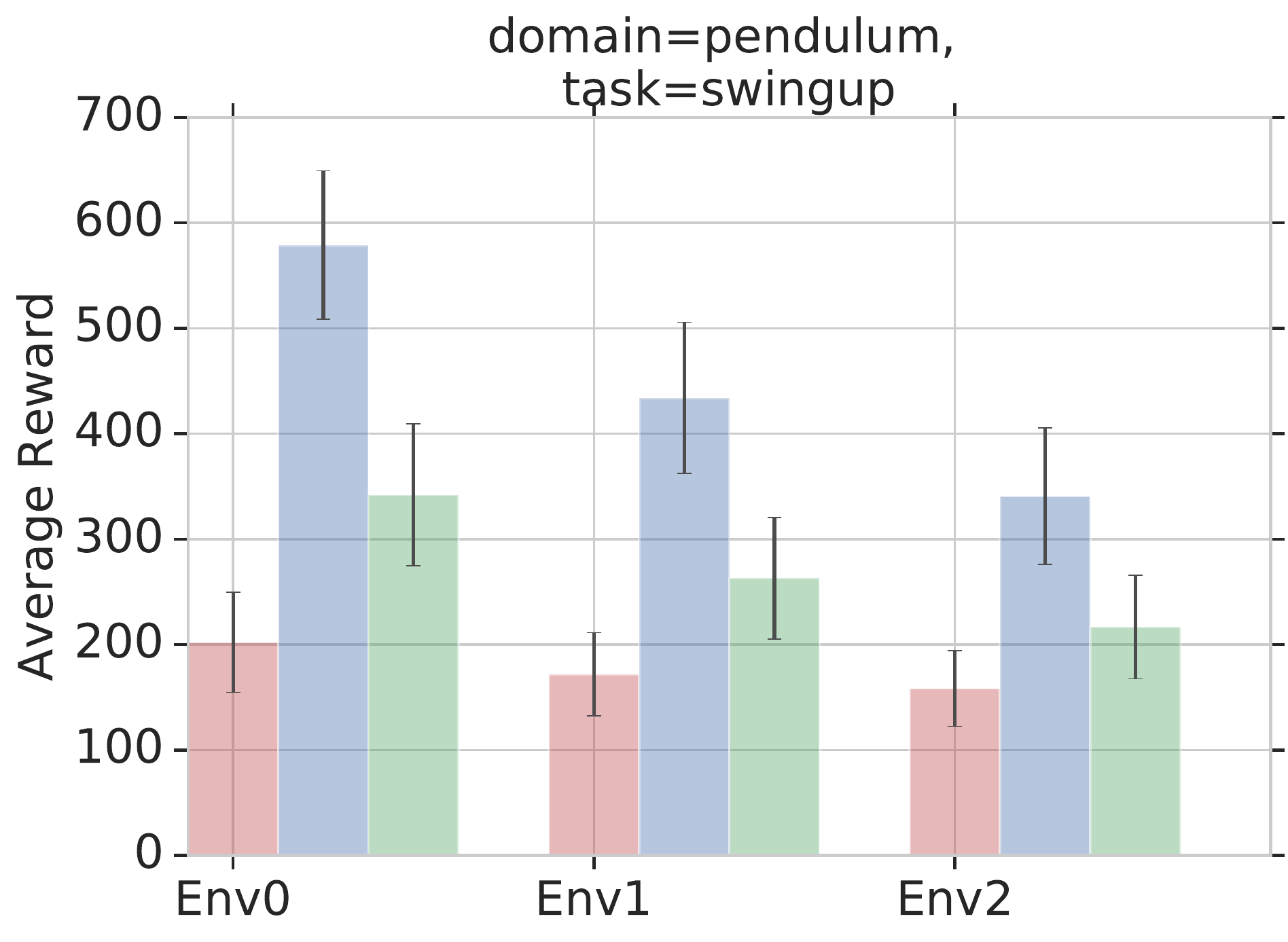}
 }
  \subfigure{
 \includegraphics[scale=\scl]{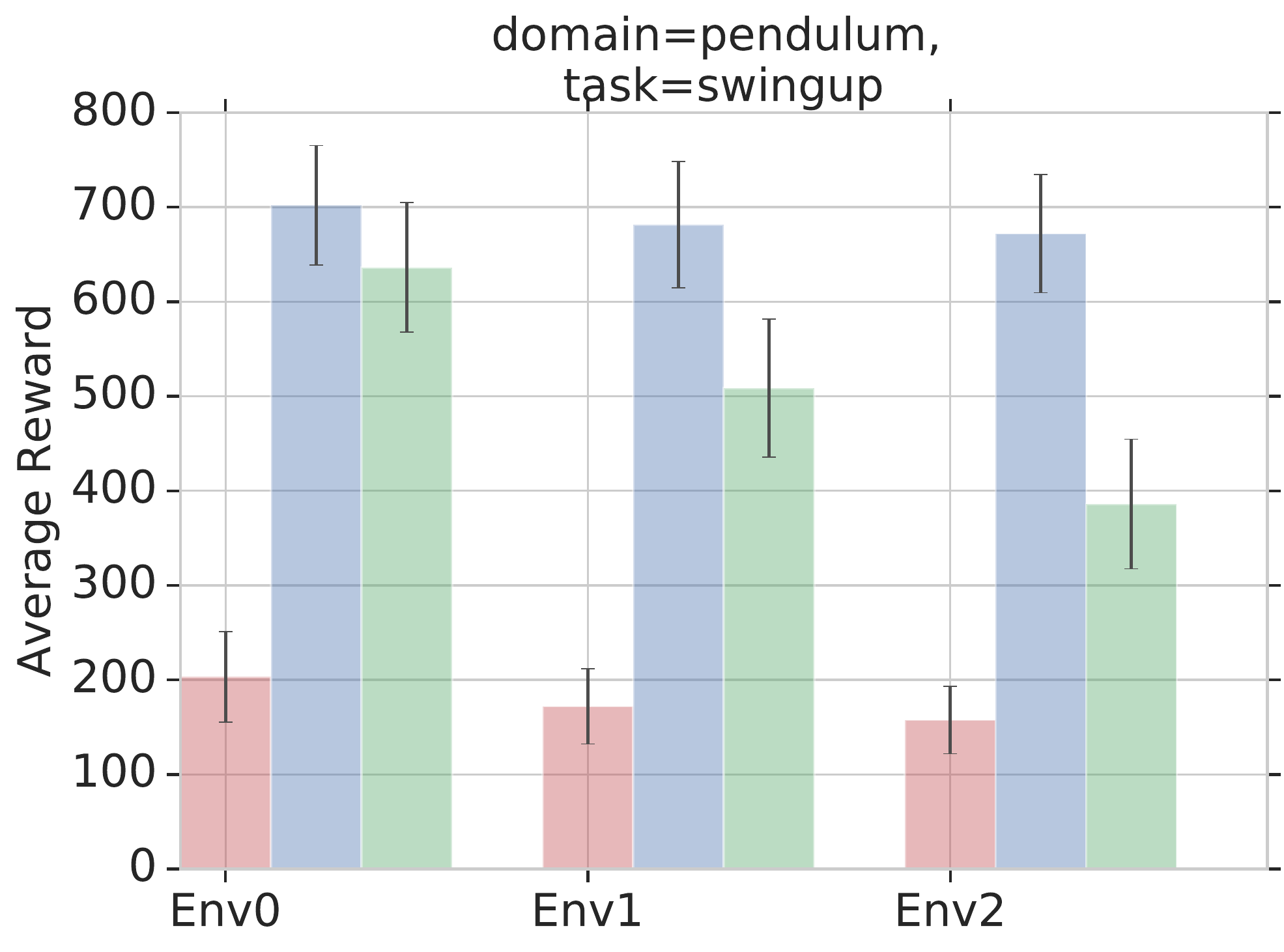}
 }
\caption{Modifying the uncertainty set: Pendulum Swingup when modifying the third perturbation of the uncertainty set to values of $1.2$ (left), $1.3$ (middle) and $2.0$ (right) meters respectively.}
\label{fig:ablation2}
\vspace{-0.5cm}
\end{figure*}

\paragraph{Modifying the uncertainty set:} We now evaluate the performance of the agent for different uncertainty sets. For Pendulum Swingup, the original uncertainty set values of the pendulum arm are $1.0, 1.1$ and $1.4$ meters. We modified the final perturbation to values of $1.2, 1.3$ and $2.0$ meters respectively. The agent is evaluated on unseen lengths of $1.5, 1.6$ and $1.7$ meters. An increase in performance can be seen in Figure \ref{fig:ablation2} as the third perturbation approaches that of the unseen evaluation environments. Thus it appears that if the agent is able to approximately capture the dynamics of the unseen test environments within the training set, then the robust agent is able to adapt to the unseen test environments. The results for cartpole balance can be seen in Appendix \ref{app:investigative}, Figure \ref{fig:app:uncertainty}.

% \todd{these numbers don't mean much without knowing the first two perturbation values, or the original third value.
% If we move to sampling from a distribution, these could be three different standard deviations, making the distribution wider. That would be clearer here too I think.} \nir{I agree on not reporting the numbers themselves. we should just say that we changed them and refer to the appendix for exact numbers}
% \nir{remove numbers}

%\nir{again with the order of figures..}
%\nir{it's a little hard to parse this sentence, as the reader don't really know the original parameters (and also the modified, once we remove them). I think we need to think of a way to pass the message clearer}

% \nir{remove numbers}

\paragraph{What about incorporating Robustness into other algorithms?}
To show the generalization of this robustness approach, we incorporate it into the critic of the Stochastic Value Gradient (SVG) continuous control RL algorithm (See Appendix \ref{app:svg}). As seen in Figure \ref{fig:dr_svg}, Robust Entropy-regularized SVG (RE-SVG) and Soft RE-SVG (SRE-SVG) significantly outperform the non-robust Entropy-regularized SVG (E-SVG) baseline in both Cartpole and Pendulum. 

\paragraph{Robust entropy-regularized return vs. robust expected return:} When comparing the robust entropy-regularized return performance to the robust expected return, we found that the entropy-regularized return appears to do no worse than the expected return. In some cases, e.g., Cheetah, the entropy-regularized objective performs significantly better (see Appendix \ref{app:investigative}, Figure \ref{fig:app:entropy}).

\paragraph{Different Nominal Models:} In this paper the nominal model was always chosen as the smallest perturbation parameter value from the uncertainty set. This was done to highlight the strong performance of robust policies to increasingly large environment perturbations. However, what if we set the nominal model as the median or largest perturbation with respect to the chosen uncertainty set for each agent? As seen in Appendix \ref{app:investigative}, Figure \ref{fig:app:nominal}, the closer (further) the nominal model is to (from) the holdout set, the better (worse) the performance of the non-robust agent. However, in all cases, the robust agent still performs at least as well as (and sometimes better than) the non-robust agent. 

\paragraph{What about learning the uncertainty set from offline data?} 

% \textit{Datacenters} \citep{gao2014machine}: In a datacenter, there are multiple cooling units that share similar dynamics. However, the exact dynamics of each cooling unit can be different due to slightly differing specifications from manufacturing inaccuracies. The units can also differ in their dynamics due to different wear and tear as well as different server loading. Given a nominal cooling unit simulator, a policy trained in the simulator may not transfer well to the individual cooling units.

% \textit{Robotics} \citep{andrychowicz2018learning,peng2018sim}: A nominal simulator is provided to learn a policy that solves a particular task. The learned policy may not accurately capture the real-world dynamics, due to differing robot specifications in the real world, robot upgrades, and robot degradation. This usually results in sub-optimal performance on the real robot.

In real-world settings, such as robotics and industrial control centers \citep{gao2014machine}, there may be a nominal simulator available as well as offline data captured from the real-world system(s). These data could be used to train transition models to capture the dynamics of the task at hand. For example, a set of robots in a factory might each be performing the same task, such as picking up a box. In industrial control cooling centers, there are a number of cooling units in each center responsible for cooling the overall system. In both of these examples, each individual robot and cooling unit operate with slightly different dynamics due to slight fluctuations in the specifications of the designed system, wear-and-tear as well as sensor calibration errors. As a result, an uncertainty set of transition models can be trained from data generated by each robot or cooling unit. 

\begin{figure*}
\centering
\newcommand{\scl}{0.24}
  \subfigure{
 \includegraphics[scale=\scl]{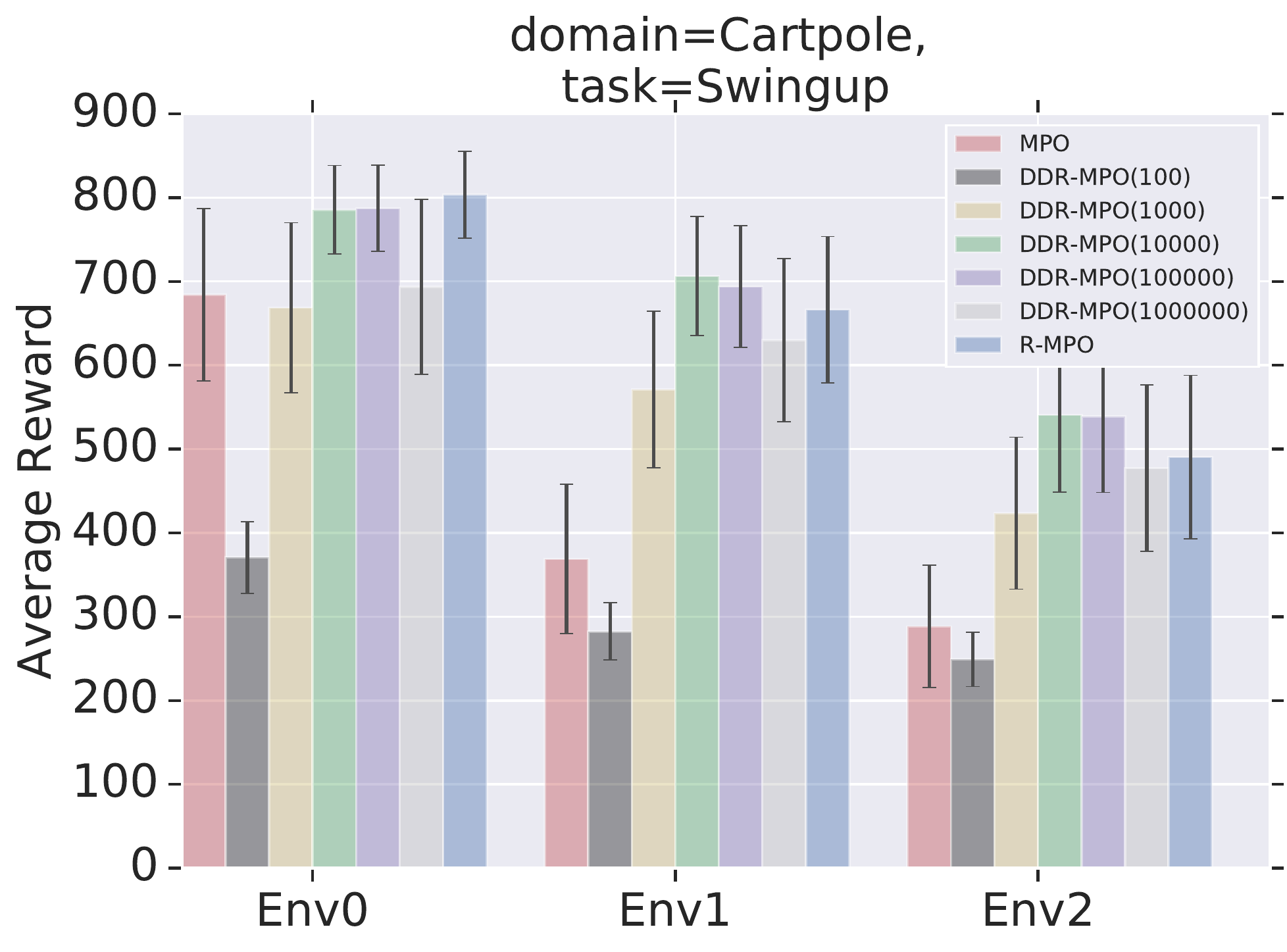}
 }
  \subfigure{
 \includegraphics[scale=\scl]{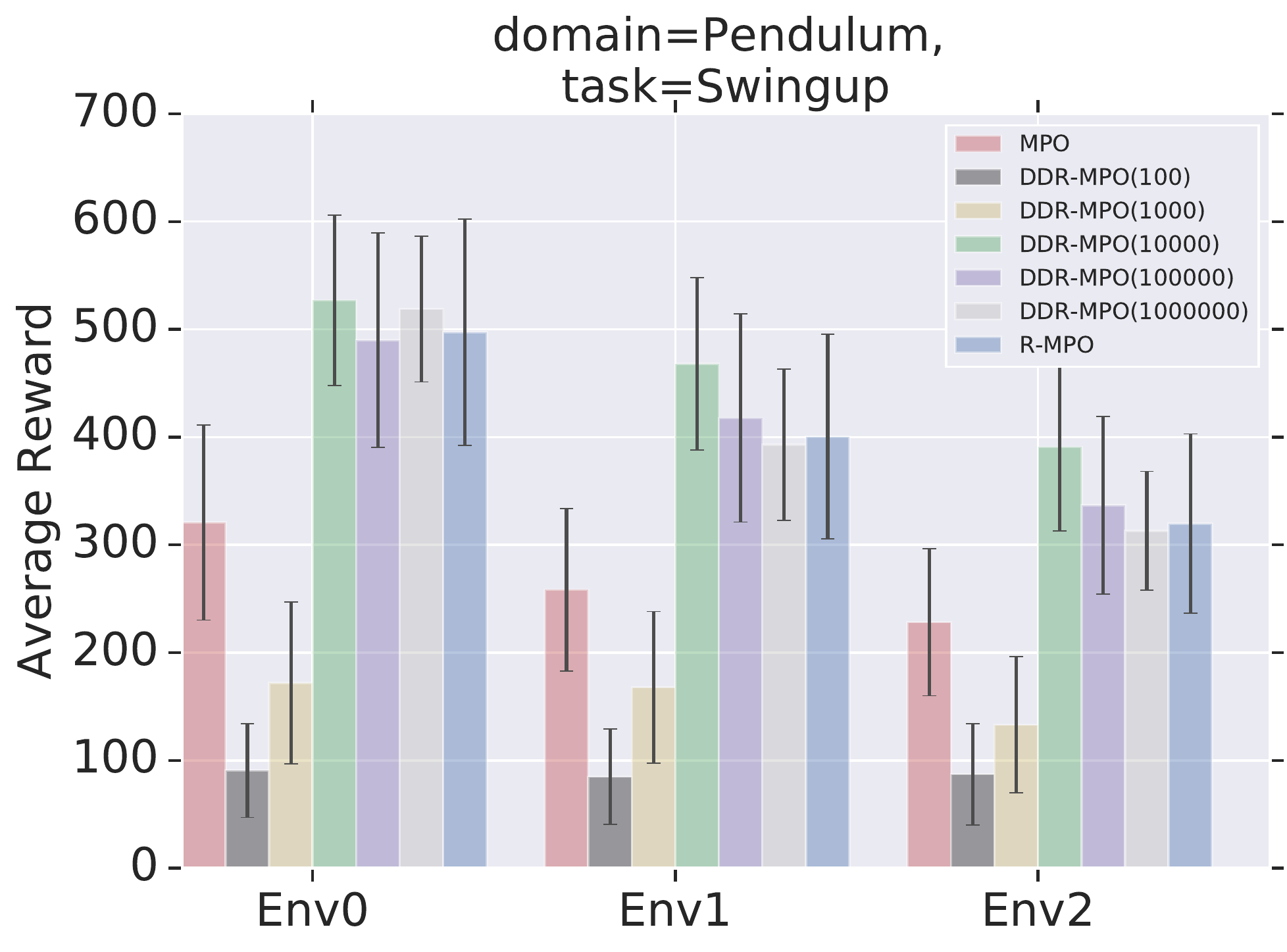}
 }
 \vspace{-0.2cm}
\caption{Training uncertainty sets of transition models on different batch sizes of offline data. The performance of Data Driven R-MPO (DDR-MPO) can be seen in the figures above for Cartpole Swingup (left) and Pendulum Swingup (right) respectively. }
\label{fig:offline}
\vspace{-0.5cm}
\end{figure*}

However, can we train a set of transition models from these data, utilize them as the uncertainty set in R-MPO and still yield robust performance when training on a nominal simulator? To answer this question, we mimicked the above scenarios by generating datasets for the Cartpole Swingup and the Pendulum swingup tasks. For Cartpole swingup, we varied the length of the pole and generated a dataset for each pole length. For Pendulum Swingup, we varied the mass of the pole and generated the corresponding datasets. We then trained transition models on increasingly large data batches ranging from $100$ to one million datapoints for each pole length and pole mass respectively.  We then utilized each set of transition models for different data batch sizes as the uncertainty set and ran R-MPO on each task. We term this variant of R-MPO, Data-Driven Robust MPO (DDR-MPO). The results can be seen in Figure \ref{fig:offline}. There are a number of interesting observations from this analysis. (1) As expected, on small batches of data, the models are too inaccurate and result in poor performance. (2) An interesting insight is that as the data batch size increases, DDR-MPO starts to \textit{outperform} R-MPO, especially for increasingly large perturbations. The hypothesis here is that due to the transition models being more accurate, but not perfect, adversarial examples are generated in a small region around the nominal next state observation, yielding an increasingly robust agent. (3) As the batch size increases further, and the transition models get increasingly close to the ground truth models, DDR-MPO converges to the performance of R-MPO.

%Two points to make: 
% 1. sometimes the uncertainty set needs to be set in and around the unseen evaluation environments and sometimes it doesn't.

% \textbf{Are there other ways to generate uncertainty sets?} It is possible to perturb the actions that an agent executes yield an uncertainty set. For this experiment, we generate an uncertainty set in the Cartpole swingup task ... \textbf{To do...} \nir{the epxeriments didn'y yield anything significant - we can take it offline}

% \textbf{TODO: any real robot experiments?}

% \nir{what about the ablation study of replacing the actor to be the middle or the other extreme of the perturbation set?}
% \rae{figure 4's ablation on dr is a bit unclear. is the red dr and purple svg?}

%\textbf{Add videos:}
% Four baselines - with/without entropy regularization, non-robust, soft-robust
% Applicable to all domains
% Run experiments with a different optimizer

\section{Related Work}
\label{sec:relatedwork}
% Previous robustness work from a theoretical and practical perspective
From a theoretical perspective, Robust Bellman operators were introduced in \citep{iyengar2005robust,Nilim2005, Wiesemann2013,Hansen2011,tamar2014scaling}. Our theoretical work extends this operator to the entropy regularized setting, for both the robust and soft-robust formulation, and modifies the MPO optimization formulation accordingly. A more closely related work from a theoretical perspective is that of \cite{Moya2016} who introduces a formulation for robustness to model misspecification. This work is a special case of robust MDPs where they introduce a robust Bellman operator that regularizes the \textit{immediate} reward with two KL terms; one entropy term capturing model uncertainty with respect to a base model, and the other term being entropy regularization with respect to a base policy. Our work differs from this work in a number of respects: (1) Their uncertainty set is represented by a KL constraint which has the effect of restricting the set of admissible transition models. Our setup does not have these restrictions. (2) The uncertainty set elements from \cite{Moya2016} output a probability distribution over model parameter space whereas the uncertainty set elements in our formulation output a distribution over next states. 

 \cite{mankowitz2018learning} learn robust options, also known as temporally extended actions \citep{sutton1999between}, using policy gradient. Robust solutions tend to be overly conservative. To combat this, \cite{derman2018soft} extend the actor-critic two-timescale stochastic approximation algorithm to a `soft-robust' formulation to yield a less, conservative solution. \cite{Dicastro2012} introduce a robust implementation of Deep Q Networks \citep{mnih2015human}. Domain Randomization (DR) \citep{andrychowicz2018learning,peng2018sim} is a technique whereby an agent trains on different perturbations of the environment. The agent batch averages the learning error of these different perturbed trajectories together to yield an agent that is robust to environment perturbations. This can be viewed as a data augmentation technique where the resulting behaviour is the average across all of the data. There are also works that look into robustness to action stochasticity \citep{Fox2015, Braun2011, Rubin2012}. 
 
 %\cite{tamar2014scaling} incorporate function approximation into the robust formulation to solve large-scale MDPs. They do so by introducing a robust dynamic programming technique based on a projected fixed point equation.

%\cite{morimoto2005robust} learn a disturbance value function by solving a min max game and augmenting the reward that the agent receives with a disturbance variable.

%\nir{if we want to avoid adversarial we need to re-word this part a bit}
%\nir{what specifically do you refer to in that section?}

%\nir{I think that the distinction between the two (and specifically why the robust method is making adversarial training) is very important. we should drill down on that.}
%\rae{agree with nir.}
\vspace{-0.3cm}
\section{Conclusion}
\vspace{-0.2cm}
We have presented a framework for incorporating robustness - to perturbations in the transition dynamics, which we refer to as model misspecification - into continuous control RL algorithms. This framework is suited to continuous control algorithms that learn a value function, such as an actor critic setup. 
We specifically focused on incorporating robustness into MPO as well as our entropy-regularized version of MPO (E-MPO). In addition, we presented an experiment which incorporates robustness into the SVG algorithm. From a \textit{theoretical} standpoint, we adapted MPO to an entropy-regularized version (E-MPO); we then incorporated robustness into the policy evaluation step of both algorithms to yield Robust MPO (R-MPO) and Robust E-MPO (RE-MPO) as well as the soft-robust variants (SR-MPO/SRE-MPO). This was achieved by deriving the corresponding robust and soft-robust entropy-regularized  Bellman operators to ensure that the policy evaluation step converges in each case. We have extensive experiments showing that the robust versions outperform the non-robust counterparts on nine Mujoco domains as well as a high-dimensional dexterous, simulated robotic hand called Shadow hand \citep{Shadow2005}. We also provide numerous investigative experiments to understand the robust and soft-robust policy in more detail. This includes an experiment showing improved robust performance over R-MPO when using an uncertainty set of transition models learned from offline data. 

\bibliography{iclr2020_conference}
\bibliographystyle{iclr2020_conference.bst}

\newpage
\appendix

\section{Background}
\label{app:entropy}
\textbf{Entropy-regularized Reinforcement Learning}: Entropy regularization encourages exploration and helps prevent early convergence to sub-optimal policies \citep{nachum2017bridging}. We make use of the relative entropy-regularized RL objective defined as $J_{\text{KL}}(\pi ; \bar{\pi})=\mathbb{E}^{\pi}[\sum_{t=0}^\infty \gamma^t (r_t - \tau \text{KL}\left(\pi(\cdot | s_t) \| \bar{\pi}(\cdot | s_t)\right))]$ where $\tau$ is a temperature parameter and $\text{KL}\left(\pi(\cdot | s_t) \| \bar{\pi}(\cdot | s_t)\right)$ is the Kullback-Leibler (KL) divergence between the current policy $\pi$ and a reference policy $\bar{\pi}$ given a state $s_t$ \citep{schulman2017equivalence}. The entropy-regularized value function is defined as $V_{\text{KL}}^{\pi}(s ; \bar{\pi}) = \mathbb{E}^{\pi}[\sum_{t=0}^\infty \gamma^t (r_t - \tau \text{KL}\left(\pi(\cdot | s_t) \| \bar{\pi}(\cdot | s_t)\right)) \vert s_0 = s]$. Intuitively, augmenting the rewards with the KL term  regularizes the policy by forcing it to be `close' in some sense to the base policy. 
% Entropy regularized value functions  - soft value functions

\section{Robust Entropy-Regularized Bellman Operator}
\label{app:rempo-bo}
(Relative-)Entropy regularization has been shown to encourage exploration and prevent early convergence to sub-optimal policies \citep{nachum2017bridging}. To take advantage of this idea when developing a robust RL algorithm we extend the robust Bellman operator to a robust \textit{entropy regularized} Bellman operator and prove that it is a contraction.\footnote{Note that while MPO already bounds the per step relative entropy we, in addition, want to regularize the action-value function to obtain a robust regularized algorithm.}
We also show that well-known value iteration bounds can be attained using this operator. We first define the \textbf{robust} entropy-regularized value function as ${V_{\text{R-KL}}^{\pi}(s ; \bar{\pi}) =  \mathbb{E}_{a \sim \pi(\cdot|s)} [r(s,a) - \tau\log\frac{\pi(\cdot|s)}{\bar{\pi}(\cdot|s)} + \gamma\inf_{p \in \mathcal{P}} \mathbb{E}_{s' \sim p(\cdot|s,a)}[ V_{\text{R-KL}}^{\pi}(s' ; \bar{\pi})]]}$. For the remainder of this section, we drop the sub-and superscripts, as well as the reference policy conditioning, from the value function $V^{\pi}_{\text{R-KL}}(s ; \bar{\pi})$, and simply represent it as $V(s)$ for brevity. We define the  robust entropy-regularized Bellman operator for a fixed policy $\pi$ in Equation \ref{eqn:re-mpo}, and show it is a max norm contraction (Theorem \ref{one}).
\vspace{-0.5cm}

\begin{eqnarray}
\mathcal{T}^\pi_{\text{R-KL}} V(s) &=&  \mathbb{E}_{a \sim \pi(\cdot|s)} [r(s, a) - \tau\log\frac{\pi(\cdot|s)}{\bar{\pi}(\cdot|s)} + \gamma \inf_{p \in \mathcal{P}} \mathbb{E}_{s' \sim p(\cdot|s,a)} [V(s')]] \enspace ,
\label{eqn:re-mpo}
\end{eqnarray}

% \nir{note that I changed the definitions of the bellman operators here and next section to be $\widetilde\pi$ instead of $\pi$, because it's confusing, as $\pi$ is associated with the value function before}

% \nir{I also changed $r(s,a)$ to be $r(s, \widetilde\pi(a|s))$ to emphasize the policy}

%We show that this is a contraction mapping in the max norm as detailed in Theorem \ref{one}.
% \begin{customthm}{1}\label{eight}
% Every theorem must be numbered by hand.
% \end{customthm}
\begin{customthm}{1}
The robust \textbf{entropy-regularized} Bellman operator $\mathcal{T}^\pi_{\text{R-KL}}$ for a fixed policy $\pi$ is a contraction operator. Specifically:  $\forall U,V \in \mathbb{R}^{|S|}$ and $\gamma \in \left(0,1\right)$, we have, $\Vert \mathcal{T}^\pi_{\text{R-KL}}U - \mathcal{T}^\pi_{\text{R-KL}}V \Vert \leq \gamma \Vert U - V \Vert
$.

%For $U,V \in \mathbf{V}$, and $\gamma \in (0,1)$, the robust entropy-regularized Bellman operator $\mathcal{T}_{\text{R-KL}}$ for a deterministic policy $\pi$ is a $\gamma$-contraction in the sup norm:
%\nir{how about: The robust entropy-regularized Bellman operator $\mathcal{T}_{\text{R-KL}}$ is a contraction operator. Specifically: forall $U,V \in \mathbb{R}^{|s|}$ and $\gamma \in \left(0,1\right)$, we have}

\label{one}
%\label{thm:contraction}
\end{customthm}
The proof can be found in the (Appendix \ref{app:proofs}, Theorem \ref{app:onefixedpolicy}). Using the optimal robust entropy-regularized Bellman operator ${T}_{\text{R-KL}} = \sup_\pi {T}^\pi_{\text{R-KL}}$, which is shown to also be a contraction operator in Appendix \ref{app:proofs}, Theorem \ref{app:one}, a standard value iteration error bound can be derived (Appendix \ref{app:proofs}, Corollary \ref{app:corrone}).

%\nir{should we explain the policy iteration algorithm before showing this result? Just how to obtain $\pi_n$ and $V^{\pi_n}$.}
% \begin{customcorollary}{1}\label{corrone}
% Let $\pi_N$ be the greedy policy after applying $N$ value iteration steps. The bound between the optimal value function $V^*$ and $V^{\pi_{N}}$, the value function that is induced by $\pi_{N}$, is given by,
% $\Vert V^* - V^{\pi_{N}} \Vert \leq \frac{2\gamma \epsilon }{(1-\gamma)^2} + \frac{2\gamma^{N+1}}{(1-\gamma)} \Vert V^* - V_{0} \Vert$, where $\epsilon=\max_{0\leq k \leq N}\Vert \mathcal{T}_{\text{R-KL}}V_{k} - V_{k+1} \Vert$ is the function approximation error, and $V_0$ is the initial value function. The proof is in the Appendix \ref{app:proofs}.
% \label{thm:2}
% \end{customcorollary}
% Robust and Soft-Robust theoretical results.

\section{\textit{Soft-}Robust Entropy-Regularized Bellman Operator}
\label{app:srempo-bo}
In this section, we derive a soft-robust entropy-regularized Bellman operator and show that it is also a $\gamma$-contraction in the max norm. First, we define the average transition model as $\bar{p} = \mathbb{E}^{p\sim w}[p]$ which corresponds to the average transition model distributed according to some distribution $w$ over the uncertainty set $\mathcal{P}$. This average transition model induces an average stationary distribution (see \cite{derman2018soft}). The \textit{soft-robust} entropy-regularized value function is defined as ${V_{\text{SR-KL}}^{\pi}(s ; \bar{\pi}) = \mathbb{E}_{a \sim \pi(\cdot|s)} [r(s,a) - \tau\log\frac{\pi(\cdot|s)}{\bar{\pi}(\cdot|s)}] + \gamma \mathbb{E}_{s' \sim \bar{p}(\cdot | s,a)}[ V_{\text{SR-KL}}^{\pi}(s' ; \bar{\pi})]}$. Again, for ease of notation, we denote $V_{\text{SR-KL}}^{\pi}(s ; \bar{\pi}) = V(s)$ for the remainder of the section. The soft-robust entropy-regularized Bellman operator for a fixed policy $\pi$ is defined as:
\begin{eqnarray}
\mathcal{T}^\pi_{\text{SR-KL}} V(s) &=& \mathbb{E}_{a \sim \pi(\cdot | s)} [r(s, a) - \tau\log\frac{\pi(\cdot|s)}{\bar{\pi}(\cdot|s)} + \gamma \mathbb{E}_{s' \sim \bar{p}(\cdot | s,a)}[V(s')]] \enspace ,
\end{eqnarray}

which is also a contraction mapping (see Appendix \ref{app:softrobust}, Theorem \ref{app:twofixedpolicy}) and yields the same bound as Corollary~\ref{app:corrone} for the optimal soft-robust Bellman operator derived in Appendix \ref{app:softrobust}, Theorem \ref{app:two}.

\section{Robust Entropy-Regularized Policy Evaluation}
\label{app:rempo-pe}
% \nir{do we want to mention the update for the soft-robust version as well?}

% \nir{something doesn't work in this whole flow (including what I put in), because we actually want to work with $Q^q_{\theta}$ all the time, and not with $Q^{\pi_k}_\theta$ (think abuot the original optimization problem from MPO). We want to compute $KL(q \| \pi_k)$, and this doesn't coincide with $Q^{\pi_k}$. We need to solve this.}

To extend Robust policy evaluation to robust \textit{entropy-regularized} policy evaluation, two key steps need to be performed: (1) optimize for the entropy-regularized expected return as opposed to the regular expected return and modify the TD update accordingly; (2) Incorporate robustness into the entropy-regularized expected return and modify the entropy-regularized TD update. To achieve (1), we define the entropy-regularized expected return as $Q_{\text{KL}}^{\pi_k}(s,a ; \bar{\pi}) = r(s,a)  - \tau \text{KL}(\pi_k(\cdot | s) \| \bar{\pi}(\cdot | s)) + \mathbb{E}_{s' \sim p(\cdot|s,a)}[V^{\pi_k}_{\text{KL}}(s' ; \bar{\pi})]$, and show in Appendix \ref{app:empo} that performing policy evaluation with the entropy-regularized value function is equivalent to optimizing the entropy-regularized squared TD error (same as Eq.~\eqref{eqn:robust_td_update}, only omitting the $\inf$ operator). To achieve (2), we optimize for the robust entropy regularized expected return objective defined as ${Q^{\pi_k}_{\text{R-KL}}(s,a; \bar{\pi}) = r(s,a)  - \tau \text{KL}(\pi_k(\cdot | s) \| \bar{\pi}(\cdot | s)) + \inf_{p \in \mathcal{P}} \mathbb{E}_{s' \sim p(\cdot|s,a)}[V^{\pi_k}_{\text{R-KL}}(s' ; \bar{\pi})]}$, yielding the robust entropy-regularized squared TD error:
\begin{equation}
\begin{aligned}
    \min_{\theta} \biggl(r_t + & \gamma \inf_{p \in \mathcal{P}(s_t,a_t)} \bigg[\widetilde Q_{\text{R-KL},\hat{\theta}}^{\pi_k}(s_{t+1} \sim p(\cdot | s_t, a_t), a_{t+1}\sim \pi_k(\cdot \vert s_{t+1}) ; \bar{\pi})\\
    & - \tau \text{KL}(\pi_k(\cdot | s_{t+1} \sim p(\cdot | s_t, a_t)) \| \bar{\pi}(\cdot | s_{t+1} \sim p(\cdot | s_t, a_t))) \bigg]  - \widetilde Q_{\text{R-KL},\theta}^{\pi_k}(s_t, a_t ; \bar{\pi}) \biggr)^2,
\end{aligned}
\label{eqn:robust_td_update}
\end{equation}

where $Q^{\pi_k}_{\text{R-KL}}(s,a ; \bar{\pi}) = \widetilde Q^{\pi_k}_{\text{R-KL}}(s,a ; \bar{\pi}) - \tau \text{KL}(\pi_k(\cdot | s) \| \bar{\pi}(\cdot | s))$. For the \textit{soft-robust} setting, we remove the infimum from the TD update and replace the next state transition function $p(\cdot | s_t, a_t)$ with the average next state transition function $\bar{p}(\cdot|s_t, a_t)$. 

\textbf{Relation to MPO:} As in the previous section, this step replaces the policy evaluation step of MPO. Our robust \textit{entropy-regularized} Bellman operator $T^{\pi_{k}}_{\text{R-KL}}$ and soft-robust \textit{entropy-regularized} Bellman operator $T^{\pi_{k}}_{\text{SR-KL}}$ ensures that this process converges to a unique fixed point for the policy $\pi_k$ for the robust and soft-robust cases respectively. We use $\pi_{k-1}$ as the reference policy $\bar{\pi}$. The pseudo code for the R-MPO, RE-MPO and Soft-Robust Entropy-regularized MPO (SRE-MPO) algorithms can be found in Appendix \ref{app:algorithm} (Algorithms \ref{Alg:GradientFree}, \ref{Alg:GradientFreeEntropy} and \ref{Alg:GradientFreeEntropySoft} respectively).

\section{Proofs}
\label{app:proofs}

\begin{customthm}{1}\label{app:onefixedpolicy}

\end{customthm}

\begin{proof}
We follow the proofs from \citep{tamar2014scaling,iyengar2005robust}, and adapt them to account for the additional entropy regularization for a fixed policy $\pi$. Let $U,V \in \mathbb{R}^{|S|}$, and $s \in S$ an arbitrary state. Assume $\mathcal{T}^{\pi}_{\text{R-KL}} U(s) \geq \mathcal{T}^\pi_{\text{R-KL}} V(s)$. Let $\epsilon > 0$ be an arbitrary positive number. 

By the definition of the $\inf$ operator, there exists $p_{s} \in \mathcal{P}$ such that,

\begin{multline}
    \mathbb{E}_{a \sim \pi(\cdot |s)} [r(s, a) - \tau\log\frac{\pi(\cdot|s)}{\bar{\pi}(\cdot|s)} + \gamma \mathbb{E}_{s' \sim p_{s}(\cdot|s,a)} [V(s')]] \\
    < \inf_{p \in \mathcal{P}} \mathbb{E}_{a \sim \pi(\cdot |s)} [r(s, a) - \tau\log\frac{\pi(\cdot|s)}{\bar{\pi}(\cdot|s)} + \gamma \mathbb{E}_{s' \sim p(\cdot|s,a)} [V(s')]] + \epsilon    
\end{multline}

In addition, we have by definition that:

\begin{multline}
    \mathbb{E}_{a \sim \pi(\cdot |s)} [r(s, a) - \tau\log\frac{\pi(\cdot|s)}{\bar{\pi}(\cdot|s)} + \gamma \mathbb{E}_{s' \sim p_{s}(\cdot|s,a)} [U(s')]] \\
    \geq  \inf_{p \in \mathcal{P}} \mathbb{E}_{a \sim \pi(\cdot |s)} [r(s, a) - \tau\log\frac{\pi(\cdot|s)}{\bar{\pi}(\cdot|s)} + \gamma \mathbb{E}_{s' \sim p(\cdot|s,a)} [U(s')]]   
\end{multline}

Thus, we have,

\begin{equation}
\begin{split}
0 &\leq \mathcal{T}^{\pi}_{\text{R-KL}} U(s) - \mathcal{T}^{\pi}_{\text{R-KL}} V(s)\\
&< \mathbb{E}_{a \sim \pi(\cdot |s)} [r(s, a) - \tau\log\frac{\pi(\cdot|s)}{\bar{\pi}(\cdot|s)} + \gamma \mathbb{E}_{s' \sim p_{s}(\cdot|s,a)} [U(s')]] \\
& \quad - \mathbb{E}_{a \sim \pi(\cdot |s)} [r(s, a) - \tau\log\frac{\pi(\cdot|s)}{\bar{\pi}(\cdot|s)} + \gamma \mathbb{E}_{s' \sim p_{s}(\cdot|s,a)} [V(s')]] + \epsilon\\
&= \mathbb{E}_{a \sim \pi(\cdot |s), s' \sim p_s(\cdot|s,a)} [ \gamma U(s')] - \mathbb{E}_{a \sim \pi(\cdot |s), s' \sim p_s(\cdot|s,a)} [ \gamma V(s')] + \epsilon\\
&\leq \gamma \Vert U - V \Vert + \epsilon
\end{split}
\end{equation}

Applying a similar argument for the case $\mathcal{T}_{\text{R-KL}} U(s) \leq \mathcal{T}_{\text{R-KL}} V(s)$ results in

\begin{equation}
| \mathcal{T}^{\pi}_{\text{R-KL}} U - \mathcal{T}^{\pi}_{\text{R-KL}} V | < \gamma \Vert U - V \Vert + \epsilon .
\end{equation}

Since $\epsilon$ is an arbitrary positive number, we establish the result, i.e.,

\begin{equation}
| \mathcal{T}^{\pi}_{\text{R-KL}} U - \mathcal{T}^{\pi}_{\text{R-KL}} V | \leq \gamma \Vert U - V \Vert.
\end{equation}
\end{proof}

\newpage
\begin{customthm}{2}\label{app:one}

\end{customthm}

\begin{proof}
We follow a similar argument to the proof of Theorem~\ref{app:onefixedpolicy}. Let $U,V \in \mathbb{R}^{|S|}$, and $s \in S$ an arbitrary state. Assume $\mathcal{T}_{\text{R-KL}} U(s) \geq \mathcal{T}_{\text{R-KL}} V(s)$. Let $\epsilon > 0$ be an arbitrary positive number. By definition of the $\sup$ operator, there exists $\hat{\pi} \in \Pi$ such that,

\begin{equation}
    \inf_{p \in \mathcal{P}} \mathbb{E}_{a \sim \hat{\pi}(\cdot |s)} [r(s, a) - \tau\log\frac{\hat{\pi}(\cdot|s)}{\bar{\pi}(\cdot|s)} + \gamma \mathbb{E}_{s' \sim p(\cdot|s,a)} [U(s')]] > \mathcal{T}_{\text{R-KL}} U(s) - \epsilon
\end{equation}

In addition, by the definition of the $\inf$ operator, there exists $p_{s} \in \mathcal{P}$ such that,

\begin{multline}
    \mathbb{E}_{a \sim \hat{\pi}(\cdot |s)} [r(s, a) - \tau\log\frac{\hat{\pi}(\cdot|s)}{\bar{\pi}(\cdot|s)} + \gamma \mathbb{E}_{s' \sim p_{s}(\cdot|s,a)} [V(s')]] \\
    < \inf_{p \in \mathcal{P}} \mathbb{E}_{a \sim \hat{\pi}(\cdot |s)} [r(s, a) - \tau\log\frac{\hat{\pi}(\cdot|s)}{\bar{\pi}(\cdot|s)} + \gamma \mathbb{E}_{s' \sim p(\cdot|s,a)} [V(s')]] + \epsilon    
\end{multline}

Thus, we have,

\begin{equation}
\begin{split}
0 &\leq \mathcal{T}_{\text{R-KL}} U(s) - \mathcal{T}_{\text{R-KL}} V(s)\\
&< (\inf_{p \in \mathcal{P}} \mathbb{E}_{a \sim \hat{\pi}(\cdot |s)} [r(s, a) - \tau\log\frac{\hat{\pi}(\cdot|s)}{\bar{\pi}(\cdot|s)} + \gamma \mathbb{E}_{s' \sim p(\cdot|s,a)} [U(s')]] + \epsilon) \\
& \quad - (\inf_{p \in \mathcal{P}} \mathbb{E}_{a \sim \hat{\pi}(\cdot |s)} [r(s, a) - \tau\log\frac{\hat{\pi}(\cdot|s)}{\bar{\pi}(\cdot|s)} + \gamma \mathbb{E}_{s' \sim p(\cdot|s,\hat{\pi}(a|s))} [V(s')]])\\
&< (\mathbb{E}_{a \sim \hat{\pi}(\cdot |s)} [r(s, a) - \tau\log\frac{\hat{\pi}(\cdot|s)}{\bar{\pi}(\cdot|s)} + \gamma \mathbb{E}_{s' \sim p_{s}(\cdot|s,a)} [U(s')]] + \epsilon)\\
& \quad - (\mathbb{E}_{a \sim \hat{\pi}(\cdot |s)} [r(s, a) - \tau\log\frac{\hat{\pi}(\cdot|s)}{\bar{\pi}(\cdot|s)} + \gamma \mathbb{E}_{s' \sim p_{s}(\cdot|s,a)} [V(s')]] - \epsilon)\\
&= \mathbb{E}_{a \sim \hat{\pi}(\cdot |s), s' \sim \bar{p}(\cdot|s,a)} [ \gamma U(s')] - \mathbb{E}_{a \sim \hat{\pi}(\cdot |s), s' \sim \bar{p}(\cdot|s,a)} [ \gamma V(s')] + 2\epsilon\\
&\leq \gamma \Vert U - V \Vert + 2\epsilon
\end{split}
\end{equation}

Applying a similar argument for the case $\mathcal{T}_{\text{R-KL}} U(s) \leq \mathcal{T}_{\text{R-KL}} V(s)$ results in

\begin{equation}
| \mathcal{T}_{\text{R-KL}} U - \mathcal{T}_{\text{R-KL}} V | < \gamma \Vert U - V \Vert + 2\epsilon .
\end{equation}

Since $\epsilon$ is an arbitrary positive number, we establish the result, i.e.,

\begin{equation}
| \mathcal{T}_{\text{R-KL}} U - \mathcal{T}_{\text{R-KL}} V | \leq \gamma \Vert U - V \Vert.
\end{equation}
\end{proof}

\newpage
\begin{customcorollary}{1}\label{app:corrone}
Let $\pi_N$ be the greedy policy after applying $N$ value iteration steps. The bound between the optimal value function $V^*$ and $V^{\pi_{N}}$, the value function that is induced by $\pi_{N}$, is given by,
$\Vert V^* - V^{\pi_{N}} \Vert \leq \frac{2\gamma \epsilon }{(1-\gamma)^2} + \frac{2\gamma^{N+1}}{(1-\gamma)} \Vert V^* - V_{0} \Vert$, where $\epsilon=\max_{0\leq k \leq N}\Vert \mathcal{T}_{\text{R-KL}}V_{k} - V_{k+1} \Vert$ is the function approximation error, and $V_0$ is the initial value function.
\end{customcorollary}
\begin{proof}
From Berteskas (1996), we have the following proposition:

\begin{customlemma}
Let $V^*$ be the optimal value function, $V$ some arbitrary value function, $\pi$ the greedy policy with respect to $V$, and $V^\pi$ the value function that is induced by $\pi$. Thus,

\begin{equation}
\Vert V^* - V^\pi \Vert \leq \frac{2\gamma}{(1-\gamma)} \Vert V^* - V \Vert
\end{equation}
\label{lem:1}
\end{customlemma}

Next, define the maximum projected loss to be: 
\begin{equation}
\epsilon = \max_{0\leq k \leq N}\Vert \mathcal{T}_{\text{R-KL}}V_k - V_{k+1} \Vert
\end{equation}

We can now derive a bound on the loss between the optimal value function $V^*$ and the value function obtained after $N$ updates of value iteration (denoted by $V_N$) as follows:

\begin{equation}
\begin{split}
\Vert V^* - V_{N} \Vert &\leq \Vert V^* - \mathcal{T}_{\text{R-KL}}V_{N-1} \Vert + \Vert \mathcal{T}_{\text{R-KL}}V_{N-1} - V_{N} \Vert\\
&= \Vert \mathcal{T}_{\text{R-KL}}V^* - \mathcal{T}_{\text{R-KL}}V_{N-1} \Vert + \Vert \mathcal{T}_{\text{R-KL}}V_{N-1} - V_{N} \Vert\\
&\leq \gamma \Vert V^* - V_{N-1} \Vert + \Vert \mathcal{T}_{\text{R-KL}}V_{N-1} - V_{N} \Vert\\
&\leq \gamma \Vert V^* - V_{N-1} \Vert + \epsilon\\
&\leq (1 + \gamma + \cdots + \gamma^{N-1})\epsilon + \gamma^{N} \Vert V^* - V_0 \Vert\\
&\leq \frac{\epsilon}{(1-\gamma)} + \gamma^{N} \Vert V^* - V_0 \Vert\\
\end{split}
\end{equation}

Then, using Lemma \ref{lem:1}, we get:

\begin{equation}
\begin{split}
\Vert V^* - V^{\pi_{N}} \Vert &\leq \frac{2\gamma}{(1-\gamma)} \Vert V^* - V_{N} \Vert\\
&\leq \frac{2\gamma}{(1-\gamma)} \frac{\epsilon}{(1-\gamma)} +\frac{2\gamma}{(1-\gamma)} \gamma^{N} \Vert V^* - V_0 \Vert\\
&= \frac{2\gamma\epsilon}{(1-\gamma)^2} + \frac{2\gamma^{N+1}}{(1-\gamma)} \Vert V^* - V_0 \Vert
\end{split}
\end{equation}

which establishes the result.
\end{proof}

% \begin{proof}
% From Berteskas (1996), we have the following proposition:

% \begin{lemma}
% The bound between the optimal value function $V^*$ and the value function $V^\pi$ that is greedy with respect to the policy $\pi$ is:

% \begin{equation}
% \Vert V^* - V^\pi \Vert \leq \frac{2\gamma}{(1-\gamma)} \Vert V^* - V \Vert
% \end{equation}
% \label{lem:1}
% \end{lemma}

% In addition, define the maximum projected loss to be: 
% \begin{equation}
% \epsilon = \max_{0\leq k \leq N}\Vert TV_k - V_{k+1} \Vert
% \end{equation}

% Then, we derive a bound on the loss between the optimal value function $V^*$ and the value function after $TM$ updates of value iteration. That is,

% \begin{eqnarray*}
% \Vert V^* - V_{N} \Vert &\leq& \Vert V^* - TV_{N-1} \Vert + \Vert TV_{N-1} - V_{N} \Vert\\
% &=& \Vert TV^* - TV_{N-1} \Vert + \Vert TV_{N-1} - V_{N} \Vert\\
% &\leq& \gamma \Vert V^* - V_{N-1} \Vert + \Vert TV_{N-1} - V_{N} \Vert\\
% &\leq& \gamma \Vert V^* - V_{N-1} \Vert + \epsilon\\
% &\leq& (1 + \gamma + \cdots + \gamma^{N-1})\epsilon + \gamma^{N} \Vert V^* - V_0 \Vert\\
% &\leq& \frac{\epsilon}{(1-\gamma)} + \gamma^{N} \Vert V^* - V_0 \Vert\\
% \end{eqnarray*}

% Then, using Lemma \ref{lem:1}, we get:

% \begin{eqnarray*}
% \Vert V^* - V^{\pi_{N}} \Vert &\leq& \frac{2\gamma}{(1-\gamma)} \Vert V^* - V_{N} \Vert\\
% &\leq& \frac{2\gamma}{(1-\gamma)} \frac{\epsilon}{(1-\gamma)} +\frac{2\gamma}{(1-\gamma)} \gamma^{N} \Vert V^* - V_0 \Vert\\
% &=& \frac{2\gamma\epsilon}{(1-\gamma)^2} + \frac{2\gamma^{N+1}}{(1-\gamma)} \Vert V^* - V_0 \Vert
% \end{eqnarray*}

% \end{proof}

\newpage
\section{Soft-Robust Entropy-Regularized Bellman Operator}
\label{app:softrobust}

\begin{customthm}{3}\label{app:twofixedpolicy}
% For $U,V \in \mathbf{V}$, and $\gamma \in (0,1)$, the soft-robust entropy regularized robust Bellman operator $\mathcal{T}_{\pi}$ for a deterministic policy $\pi$ is a $\gamma$-contraction in the sup norm:

% \begin{equation}
% \Vert \mathcal{T}U - \mathcal{T}V \Vert \leq \gamma \Vert U - V \Vert    
% \end{equation}
\end{customthm}

\begin{proof}
For an arbitrary $U,V \in \mathbb{R}^{|S|}$ and for a fixed policy $\pi$:

\begin{eqnarray*}
\begin{split}
\Vert \mathcal{T}^\pi_{\text{SR-KL}}U(s) - \mathcal{T}^\pi_{\text{SR-KL}}V(s) \Vert_\infty & \\
&= \sup_s \bigg| \mathbb{E}_{a \sim \pi(\cdot |s)} [r(s, a) - \tau\log\frac{\pi(\cdot|s)}{\bar{\pi}(\cdot|s)} + \gamma \mathbb{E}_{s' \sim \bar{p}(\cdot|s,a)} [U(s')]] \\
& \quad - \mathbb{E}_{a \sim \pi(\cdot |s)} [r(s, a) - \tau\log\frac{\pi(\cdot|s)}{\bar{\pi}(\cdot|s)} + \gamma \mathbb{E}_{s' \sim \bar{p}(\cdot|s,a)} [V(s')]] \bigg|\\
&= \gamma \sup_s \vert  \sum_{s'} \bar{p}(s' \vert s,a) [U(s') -  V(s')] \vert\\
&\leq \gamma \sup_s  \sum_{s'} \bar{p}(s' \vert s,a) \vert U(s') -  V(s') \vert\\
&\leq \gamma \sup_s  \sum_{s'} \bar{p}(s' \vert s,a) \Vert U(s') -  V(s') \Vert_\infty\\
&\leq \gamma \Vert U - V \Vert_\infty
\end{split}
\end{eqnarray*}

\end{proof}

\begin{customthm}{4}\label{app:two}
% For $U,V \in \mathbf{V}$, and $\gamma \in (0,1)$, the soft-robust entropy regularized robust Bellman operator $\mathcal{T}_{\pi}$ for a deterministic policy $\pi$ is a $\gamma$-contraction in the sup norm:

% \begin{equation}
% \Vert \mathcal{T}U - \mathcal{T}V \Vert \leq \gamma \Vert U - V \Vert    
% \end{equation}
\end{customthm}

\begin{proof}
Let $U,V \in \mathbb{R}^{|S|}$, and $s \in S$ an arbitrary state. Assume $\mathcal{T}_{\text{SR-KL}} U(s) \geq \mathcal{T}_{\text{SR-KL}} V(s)$. Let $\epsilon > 0$ be an arbitrary positive number. By definition of the $\sup$ operator, there exists $\hat{\pi} \in \Pi$ such that,

\begin{equation}
    \mathbb{E}_{a \sim \hat{\pi}(\cdot |s)} [r(s, a) - \tau\log\frac{\hat{\pi}(\cdot|s)}{\bar{\pi}(\cdot|s)} + \gamma \mathbb{E}_{s' \sim \bar{p}(\cdot|s,a)} [U(s')]] > \mathcal{T}_{\text{SR-KL}} U(s) - \epsilon
\end{equation}

Thus, we have,

\begin{equation}
\vspace{-0.2in}
\begin{split}
0 & \leq \mathcal{T}_{\text{SR-KL}} U(s) - \mathcal{T}_{\text{SR-KL}} V(s)\\
&< (\mathbb{E}_{a \sim \hat{\pi}(\cdot |s)} [r(s, a) - \tau\log\frac{\hat{\pi}(\cdot|s)}{\bar{\pi}(\cdot|s)} + \gamma \mathbb{E}_{s' \sim \bar{p}(\cdot|s,a)} [U(s')]] + \epsilon) \\
& \quad - (\sup_{\pi \in \Pi} \mathbb{E}_{a \sim \pi(\cdot |s)} [r(s, a) - \tau\log\frac{\pi(\cdot|s)}{\bar{\pi}(\cdot|s)} + \gamma \mathbb{E}_{s' \sim \bar{p}(\cdot|s,a)} [V(s')]])\\
& \leq (\mathbb{E}_{a \sim \hat{\pi}(\cdot |s)} [r(s, a) - \tau\log\frac{\hat{\pi}(\cdot|s)}{\bar{\pi}(\cdot|s)} + \gamma \mathbb{E}_{s' \sim \bar{p}(\cdot|s,a)} [U(s')]] + \epsilon) \\
& \quad - (\mathbb{E}_{a \sim \hat{\pi}(\cdot |s)} [r(s, a) - \tau\log\frac{\hat{\pi}(\cdot|s)}{\bar{\pi}(\cdot|s)} + \gamma \mathbb{E}_{s' \sim \bar{p}(\cdot|s,a)} [V(s')]])\\
&= \mathbb{E}_{a \sim \hat{\pi}(\cdot |s), s' \sim \bar{p}(\cdot|s,a)} [ \gamma U(s')] - \mathbb{E}_{a \sim \hat{\pi}(\cdot |s), s' \sim \bar{p}(\cdot|s,a)} [ \gamma V(s')] + \epsilon\\
&\leq \gamma \Vert U - V \Vert + \epsilon    
\end{split}
\end{equation}

Applying a similar argument for the case $\mathcal{T}_{\text{SR-KL}} U(s) \leq \mathcal{T}_{\text{SR-KL}} V(s)$ results in

\begin{equation}
| \mathcal{T}_{\text{SR-KL}} U - \mathcal{T}_{\text{SR-KL}} V | < \gamma \Vert U - V \Vert + \epsilon .
\end{equation}

Since $\epsilon$ is an arbitrary positive number, we establish the result, i.e.,

\begin{equation}
| \mathcal{T}_{\text{SR-KL}} U - \mathcal{T}_{\text{SR-KL}} V | \leq \gamma \Vert U - V \Vert.
\end{equation}

\end{proof}

\newpage
\section{Entropy-regularized Policy Evaluation}
\label{app:empo}

This section describes: (1) modification to the TD update for the expected return to optimize for the entropy-regularized expected return, (2) additional modification to account for robustness.

We start with (1).

The entropy-regularized value function is defined as:

\begin{eqnarray}
V^{\pi}_{\text{KL}}(s ; \bar{\pi}) &=& \mathbb{E}^{\pi}[\sum_{t=0}^{\infty} \gamma^t (r_t - \tau\text{KL}(\pi(\cdot | s_t) \| \bar{\pi}(\cdot | s_t))) | s_0 = s]
\end{eqnarray}

and the corresponding entropy-regularized action value function is given by:

\begin{eqnarray}
Q^{\pi}_{\text{KL}}(s,a ; \bar{\pi}) &=& \mathbb{E}^{\pi}[\sum_{t=0}^{\infty} \gamma^{t} (r_t - \tau \text{KL}(\pi(\cdot | s_t) \| \bar{\pi}(\cdot | s_t))) | s_0=s, a_0=a]\\
&=& r(s, a)  - \tau \text{KL}(\pi(\cdot | s) \| \bar{\pi}(\cdot | s)) + \mathbb{E}_{s' \sim p(\cdot | s, a)}[V^{\pi}_{\text{KL}}(s' ; \bar{\pi})]
\end{eqnarray}

Next, we define:

\begin{equation}
    \widetilde{Q}^{\pi}_{\text{KL}}(s,a ; \bar{\pi}) = r(s, a) + \mathbb{E}_{s' \sim p(\cdot | s, a)}[V^{\pi}_{\text{KL}}(s' ; \bar{\pi})]
\end{equation}

thus,

\begin{equation}
    Q^{\pi}_{\text{KL}}(s,a ; \bar{\pi}) = \widetilde{Q}^{\pi}_{\text{KL}}(s,a ; \bar{\pi}) - \tau \text{KL}(\pi(\cdot | s) \| \bar{\pi}(\cdot | s)))
\end{equation}

Therefore, we have the following relationship:

\begin{equation}
    V^{\pi}_{\text{KL}}(s' ; \bar{\pi}) = \mathbb{E}_{a \sim \pi(\cdot | s)}\biggl[Q^{\pi}_{\text{KL}}(s,a ; \bar{\pi}) \biggr] = \mathbb{E}_{a \sim \pi(\cdot | s)}\biggl[\widetilde{Q}^{\pi}_{\text{KL}}(s,a ; \bar{\pi}) - \tau \text{KL}(\pi(\cdot | s) \| \bar{\pi}(\cdot | s))\biggr]
\end{equation}

We now retrieve the TD update for the entropy-regularized action value function:

\begin{equation}
\begin{aligned}
    \delta_{t} &= r_t - \tau \text{KL}(\pi(\cdot | s_t) \| \bar{\pi}(\cdot | s_t)) + \gamma Q^{\pi}_{\text{KL}}(s_{t+1} \sim P(\cdot | s_t, a_t),a_{t+1} \sim \pi(\cdot | s_{t+1}) ; \bar{\pi})\\
    & \quad - Q^{\pi}_{\text{KL}}(s_t,a_t ; \bar{\pi}) \\
    &= r_t - \tau \text{KL}(\pi(\cdot | s_t) \| \bar{\pi}(\cdot | s_t)) + \gamma Q^{\pi}_{\text{KL}}(s_{t+1} \sim P(\cdot | s_t, a_t),a_{t+1} \sim \pi(\cdot | s_{t+1}) ; \bar{\pi})  \\
    & \quad - \widetilde{Q}^{\pi}_{\text{KL}}(s_t,a_t ; \bar{\pi}) + \tau \text{KL} (\pi(\cdot | s_t) \| \bar{\pi}(\cdot | s_t)) \\
    &= r_t + \gamma Q^{\pi}_{\text{KL}}(s_{t+1} \sim P(\cdot | s_t, a_t),a_{t+1} \sim \pi(\cdot | s_{t+1}) ; \bar{\pi}) - \widetilde{Q}^{\pi}_{\text{KL}}(s_t,a_t ; \bar{\pi})\\
    &= r_t + \gamma \bigg[\widetilde{Q}_{\text{KL}}^{\pi}(s_{t+1} \sim P(\cdot | s_t, a_t), a_{t+1}\sim \pi(\cdot \vert s_{t+1}) ; \bar{\pi}) \\
    & \quad - \tau \text{KL}(\pi(\cdot | s_{t+1} \sim P(\cdot | s_t, a_t)) \| \bar{\pi}(\cdot | s_{t+1} \sim P(\cdot | s_t, a_t))) \bigg]  - \widetilde{Q}_{\text{KL}}^{\pi}(s_t, a_t ; \bar{\pi})\\
\end{aligned}
\end{equation}

Note that in the above TD update we replaced $Q^{\pi}_{\text{KL}}$ with $\widetilde{Q}^{\pi}_{\text{KL}}$.

Next, we move to (2).

Before extending the TD update to the robust case, we first consider the \textit{robust} entropy-regularized value function, which is defined as:

\begin{eqnarray}
V^{\pi}_{\text{R-KL}}(s ; \bar{\pi}) &=& \inf_{p \in \mathcal{P}} \mathbb{E}^{p, \pi}[\sum_{t=0}^{\infty} \gamma^t (r_t - \tau\text{KL}(\pi(\cdot | s_t) \| \bar{\pi}(\cdot | s_t))) | s_0 = s]
\end{eqnarray}

Applying similar steps as above yields the following TD update:

\begin{equation}
\begin{aligned}
    \delta_{t} &= r_t - \tau \text{KL}(\pi(\cdot | s_t) \| \bar{\pi}(\cdot | s_t)) + \gamma \inf_{p \in \mathcal{P}} Q^{\pi}_{\text{R-KL}}(s_{t+1} \sim p(\cdot | s_t, a_t),a_{t+1} \sim \pi(\cdot | s_{t+1}) ;\bar{\pi})\\
    & \quad - Q^{\pi}_{\text{R-KL}}(s_t,a_t ; \bar{\pi}) \\
    &= r_t - \tau \text{KL}(\pi(\cdot | s_t) \| \bar{\pi}(\cdot | s_t)) + \gamma \inf_{p \in \mathcal{P}} Q^{\pi}_{\text{R-KL}}(s_{t+1} \sim p(\cdot | s_t, a_t),a_{t+1} \sim \pi(\cdot | s_{t+1}) ; \bar{\pi})  \\
    & \quad - \widetilde{Q}^{\pi}_{\text{R-KL}}(s_t,a_t ; \bar{\pi}) + \tau \text{KL} (\pi(\cdot | s_t) \| \bar{\pi}(\cdot | s_t)) \\
    &= r_t + \gamma \inf_{p \in \mathcal{P}} Q^{\pi}_{\text{R-KL}}(s_{t+1} \sim p(\cdot | s_t, a_t),a_{t+1} \sim \pi(\cdot | s_{t+1}) ; \bar{\pi}) - \widetilde{Q}^{\pi}_{\text{R-KL}}(s_t,a_t ; \bar{\pi})\\
    &= r_t + \gamma \inf_{p \in \mathcal{P}} \bigg[\widetilde{Q}_{\text{R-KL}}^{\pi}(s_{t+1} \sim p(\cdot | s_t, a_t), a_{t+1}\sim \pi(\cdot \vert s_{t+1}) ; \bar{\pi}) \\
    & \quad - \tau \text{KL}(\pi(\cdot | s_{t+1} \sim p(\cdot | s_t, a_t)) \| \bar{\pi}(\cdot | s_{t+1} \sim p(\cdot | s_t, a_t))) \bigg]  - \widetilde{Q}_{\text{R-KL}}^{\pi}(s_t, a_t ; \bar{\pi})\\
\end{aligned}
\end{equation}

\newpage
\section{Experiments}
\label{app:experiments}

\subsection{Additional Details on the SVG baseline}
\label{app:svg}
For the stochastic value gradients SVG(0) baseline we use the same policy parameterization as for our algorithm, e.g. we have 
$$
\pi_\theta = \mathcal{N}(\mu_\theta(s),\sigma^2_\theta(s) I),
$$
where $I$ denotes the identity matrix and $\sigma_\theta(s)$ is computed from the network output via a softplus activation function. 

To obtain a baseline that is, in spirit, similar to our algorithm we used SVG in combination with Entropy regularization. That is, we optimize the policy via gradiend ascent, following the reparameterized gradient for a given state s sampled from the replay:
\begin{equation}
\begin{aligned}
\nabla_\theta \mathbb{E}_{\pi_\theta(a | s)}[Q(a, s)] + \alpha \mathrm{H}\Big(\pi_\theta(a | s)\Big),
\end{aligned}
\end{equation}
which can be computed, using the reparameterization trick, as 
\begin{equation}
\begin{aligned}
 \mathbb{E}_{\zeta \sim \mathcal{N}(0, I)}[\nabla_\theta g_\theta(s, \zeta) \nabla_g Q(g_\theta(s, \zeta), s)] + \alpha \nabla_\theta \mathrm{H}\Big(\pi_\theta(a | s)\Big),
\end{aligned}
\end{equation}
where $g_\theta(s, \zeta)$ is now a deterministic function of a sample from the standard multivariate normal distribution. See e.g. \cite{svg} (for SVG) as well as \cite{rezende14,kingma2013auto} (for the reparameterization trick) for a detailed explanation.

\subsection{Experiment Details for MPO and SVG}

In this section we outline the details on the hyperparameters used for the MPO and SVG algorithms. All experiments use a feed-forward two layer neural network with 50 neurons to map the current state of the network to the mean and diagonal covariance of the Gaussian policy. The policy is given by a Gaussian distribution with a diagonal covariance matrix. The neural network outputs the mean $\mu=\mu(s)$ and diagonal Cholesky factors $A=A(s)$, such that $\Sigma = AA^T$. The diagonal factor $A$ has positive diagonal elements enforced by the softplus transform $A_{ii} \leftarrow \log(1 + \exp(A_{ii}))$ to ensure positive definiteness of the diagonal covariance matrix. Tables \ref{table:mpo_hyperparameters} and \ref{table:svg_hyperparameters} show the hyperparameters used for the MPO and SVG algorithms. 

\begin{table}[t]
\begin{center}
 \begin{tabular}{c||c} 
 Hyperparameters & SVG \\
 \hline
 Policy net & 200-200-200 \\
 Q function net & 500-500-500 \\
 Discount factor ($\gamma$) & 0.99 \\
 Adam learning rate & 0.0003 \\
 Replay buffer size & 1000000 \\
 Target network update period & 200 \\
 Batch size & 1024 \\
 Activation function & elu\\
 Tanh on output of layer norm & Yes\\
 Layer norm on first layer & Yes\\
 Tanh on Gaussian mean & Yes \\
 Min variance & 0.1\\
 Max variance & unbounded 
\end{tabular}
\end{center}
\caption{Hyperparameters for SVG}
\label{table:svg_hyperparameters}
\end{table}

\begin{table}[t]
\begin{center}
 \begin{tabular}{c||c} 
 Hyperparameters & MPO \\
 \hline
 Policy net & 200-200-200 \\ 
 Number of actions sampled per state& 15\\
 Q function net & 500-500-500 \\
 $\epsilon$ & 0.1 \\
 $\epsilon_{\mu}$ & 0.01 \\
 $\epsilon_{\Sigma}$ & 0.00001\\
 Discount factor ($\gamma$) & 0.99 \\
 Adam learning rate & 0.0003 \\
 Replay buffer size & 1000000 \\
 Target network update period & 200\\
 Batch size & 1024\\
 Activation function & elu\\
 Layer norm on first layer & Yes\\
 Tanh on output of layer norm & Yes\\
 Tanh on Gaussian mean & No \\
 Min variance & Zero \\
 Max variance & unbounded 
\end{tabular}
\end{center}
\caption{Hyperparameters for MPO}
\label{table:mpo_hyperparameters}
\end{table}

\subsection{Uncertainty set parameters}
\label{app:hyperparameters}
Table \ref{tab:perturb_parameters} contains the chosen uncertainty set values for each of the domains and the corresponding holdout set perturbations. The final column of the table contains the parameter that was perturbed.

\begin{table}[]
\centering
\small
\caption{The parameters chosen for the uncertainty set perturbations as well as the holdout set perturbations. The final column contains the parameter that was perturbed.}
\begin{tabular}{|c|c|c|c|}
\hline
\textbf{Domain}           & \textbf{Uncertainty Set Perturbations} & \textbf{Hold-out Test Perturbations} & \textbf{Parameter} \\ \hline
\textbf{Acrobot}          & 1.0, 1.025, 1.05 meters                & 1.15, 1.2, 1.25 meters               & First pole length  \\ \hline
\textbf{Cartpole Balance} & 0.5, 1.9, 2.1 meters                   & 2.0, 2.2, 2.3 meters                 & Pole length        \\ \hline
\textbf{Cartpole Swingup} & 1.0, 1.4, 1.7 meters                   & 1.2, 1.5, 1.8 meters                 & Pole Length        \\ \hline
\textbf{Cheetah Run}      & 0.4, 0.45, 0.5 meters                  & 0.3, 0.325, 0.35 meters              & Torso Length       \\ \hline
\textbf{Hopper Hop}       & -0.32, -0.33, -0.34 meters             & -0.4, -0.45, -0.5 meters             & Calf Length        \\ \hline
\textbf{Hopper Stand}     & -0.32, -0.33, -0.34 meters             & -0.4, -0.475, -0.5 meters            & Calf Length        \\ \hline
\textbf{Pendulum Swingup} & 1.0, 1.1, 1.4 Kg                       & 1.5, 1.6, 1.7 Kg                     & Ball Mass          \\ \hline
\textbf{Walker Run}       & 0.225, 0.2375, 0.25 meters             & 0.35, 0.375, 0.4 meters              & Thigh Lengths      \\ \hline
\textbf{Walker Walk}      & 0.225, 0.2375, 0.25 meters             & 0.35, 0.375, 0.4 meters              & Thigh Lengths      \\ \hline
\textbf{Shadow hand}      & 0.025, 0.022, 0.02 meters             & 0.021, 0.018, 0.015 meters              & Half-cube width      \\ \hline
\textbf{Cartpole Balance:}      & 0.5, 1.9 ,2.1 meters             & 0.5, 0.7, 0.9, 1.1,              & Pole Length     \\ 
\textbf{Larger Test Set}      &              & 1.3, 1.5, 1.7, 1.9 meters              &       \\ \hline
\textbf{Pendulum Swingup:}      & 1.0, 1.1, 1.4 meters             & 1.0, 1.1, 1.2,              & Pole Length      \\ 
\textbf{Larger Test Set}      &              & 1.3, 1.4, 1.5 meters              &      \\ \hline
\textbf{Pendulum Swingup - Offline datasets} & 1.0, 1.1, 1.2 Kg                       & 1.5, 1.6, 1.7 Kg                     & Ball Mass          \\ \hline
\textbf{Cartpole Swingup - Offline datasets} & 1.0, 1.4, 1.7 meters                   & 1.2, 1.5, 1.8 meters                 & Pole Length  \\ \hline
\end{tabular}
\label{tab:perturb_parameters}
\end{table}

\subsection{Main Experiments}
\label{app:mainexperiments}
\begin{figure*}
\centering
\newcommand{\scl}{0.22}
 \subfigure{
 \includegraphics[scale=\scl]{./figures/mpo_kl/cartpole_balance}
 }
  \subfigure{
 \includegraphics[scale=\scl]{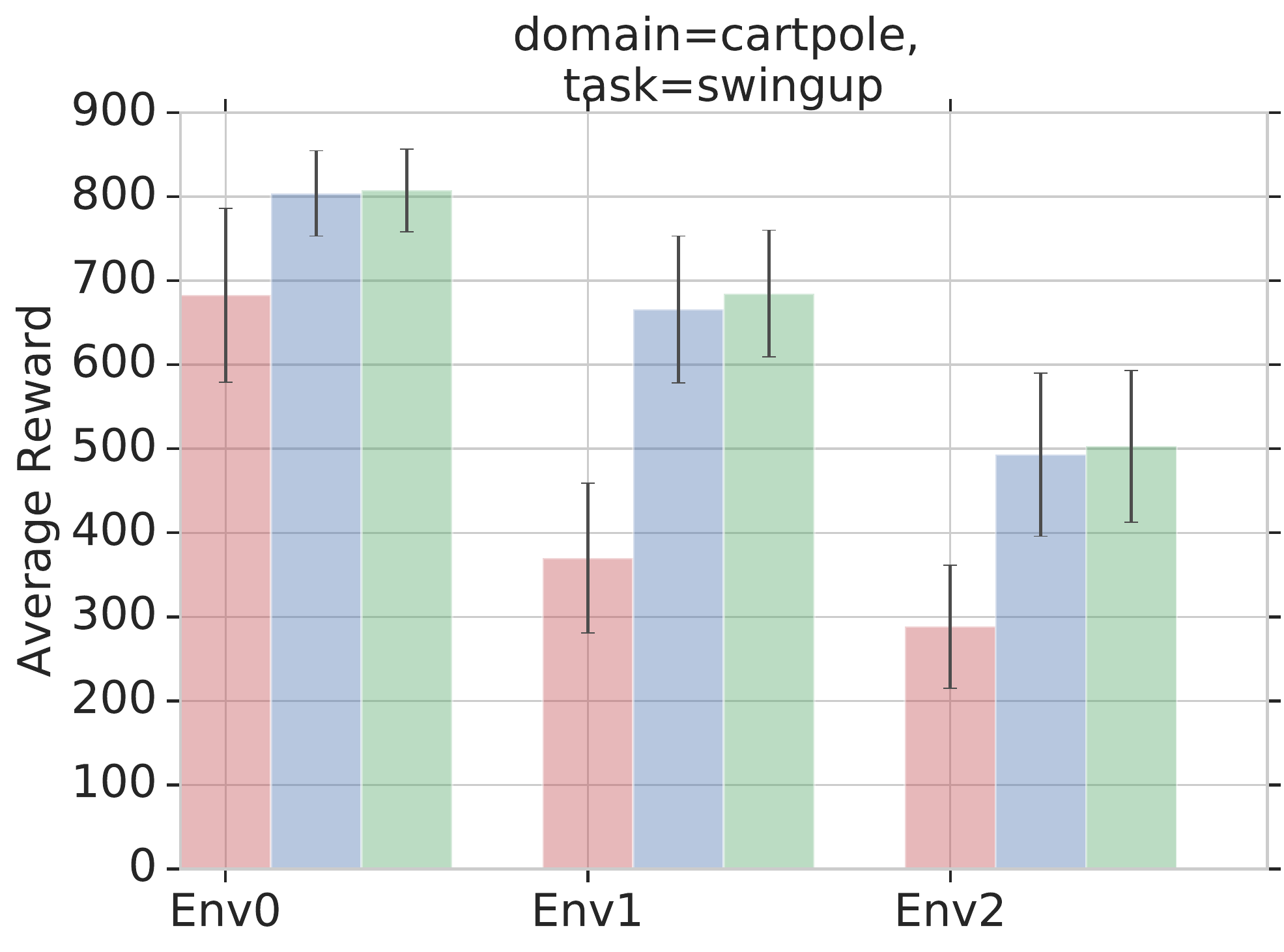}
 }
 \subfigure{
 \includegraphics[scale=\scl]{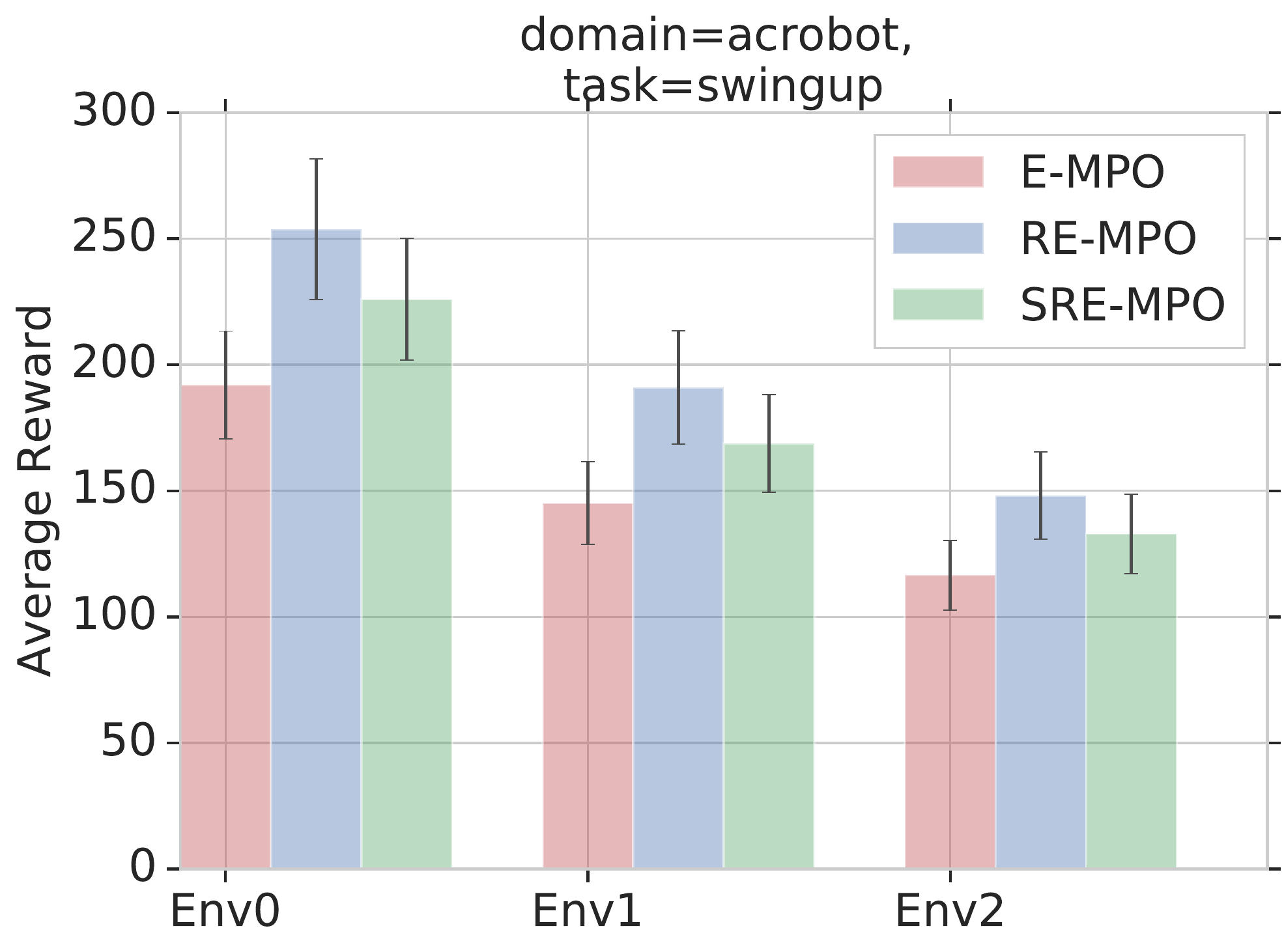}
 }
  \subfigure{
 \includegraphics[scale=\scl]{./figures/mpo_kl/hopper_hop}
 }
  \subfigure{
 \includegraphics[scale=\scl]{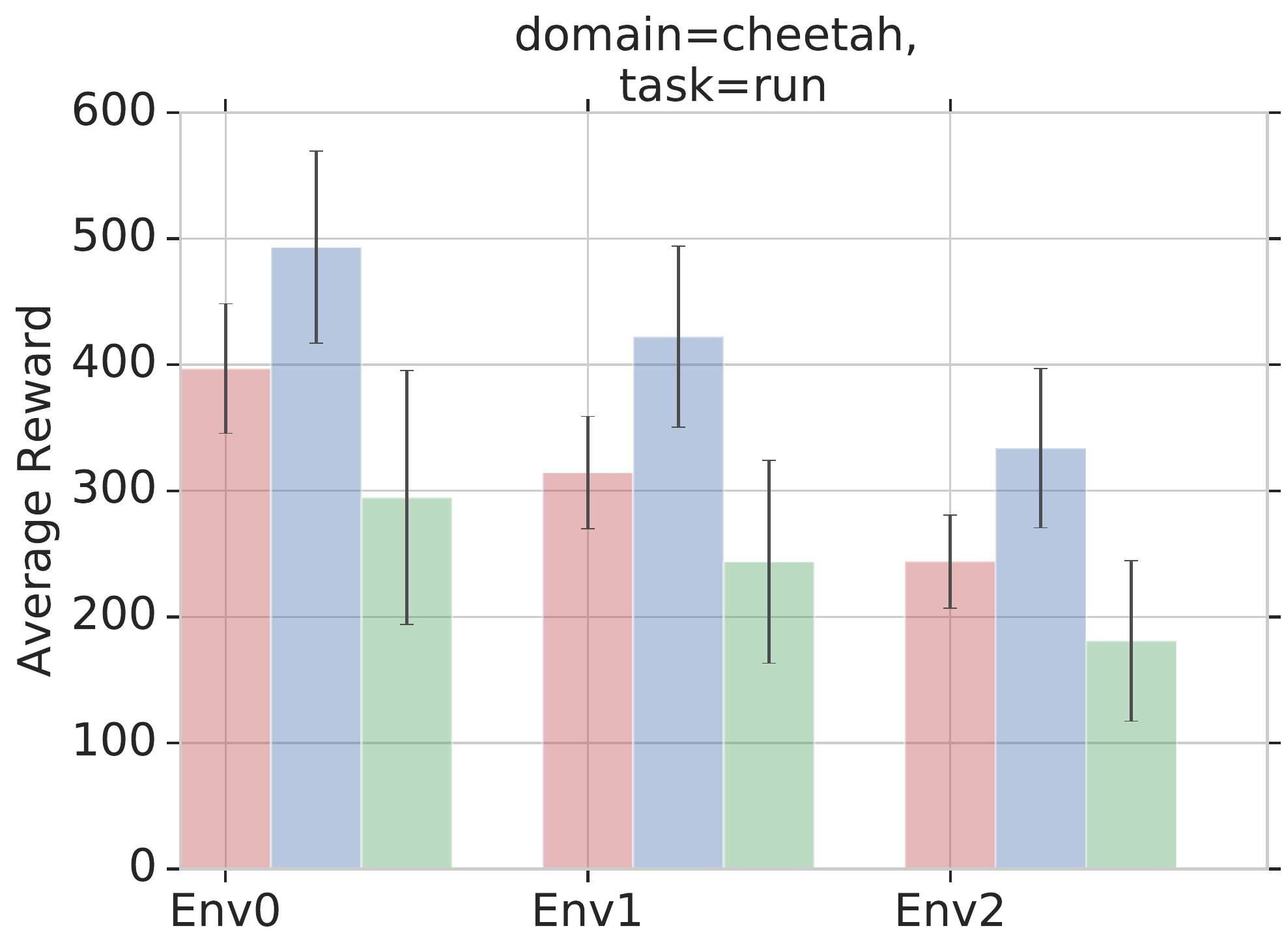}
 }
 \subfigure{
 \includegraphics[scale=\scl]{./figures/mpo_kl/walker_walk}
 }
   \subfigure{
 \includegraphics[scale=\scl]{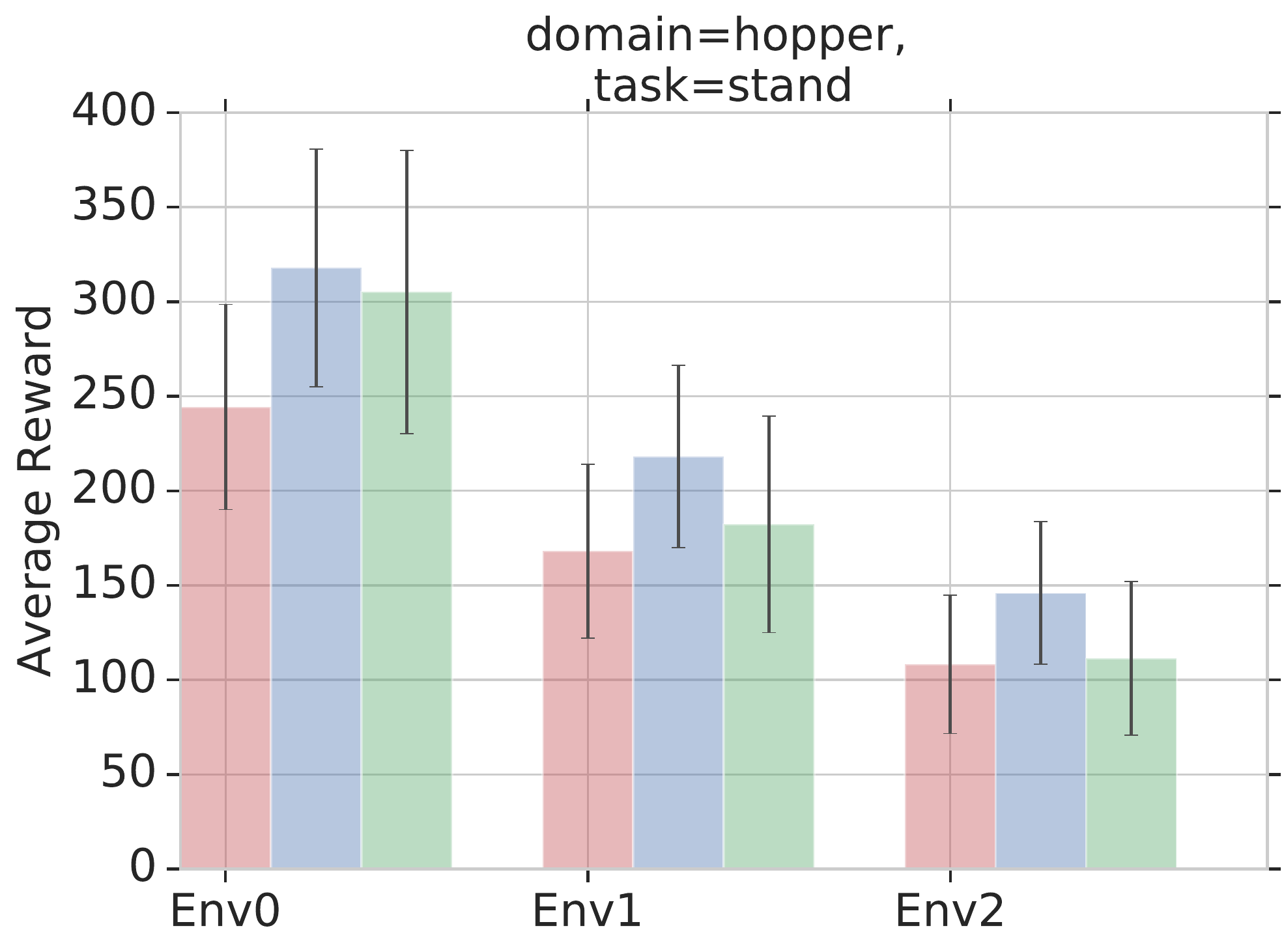}
 }
  \subfigure{
 \includegraphics[scale=\scl]{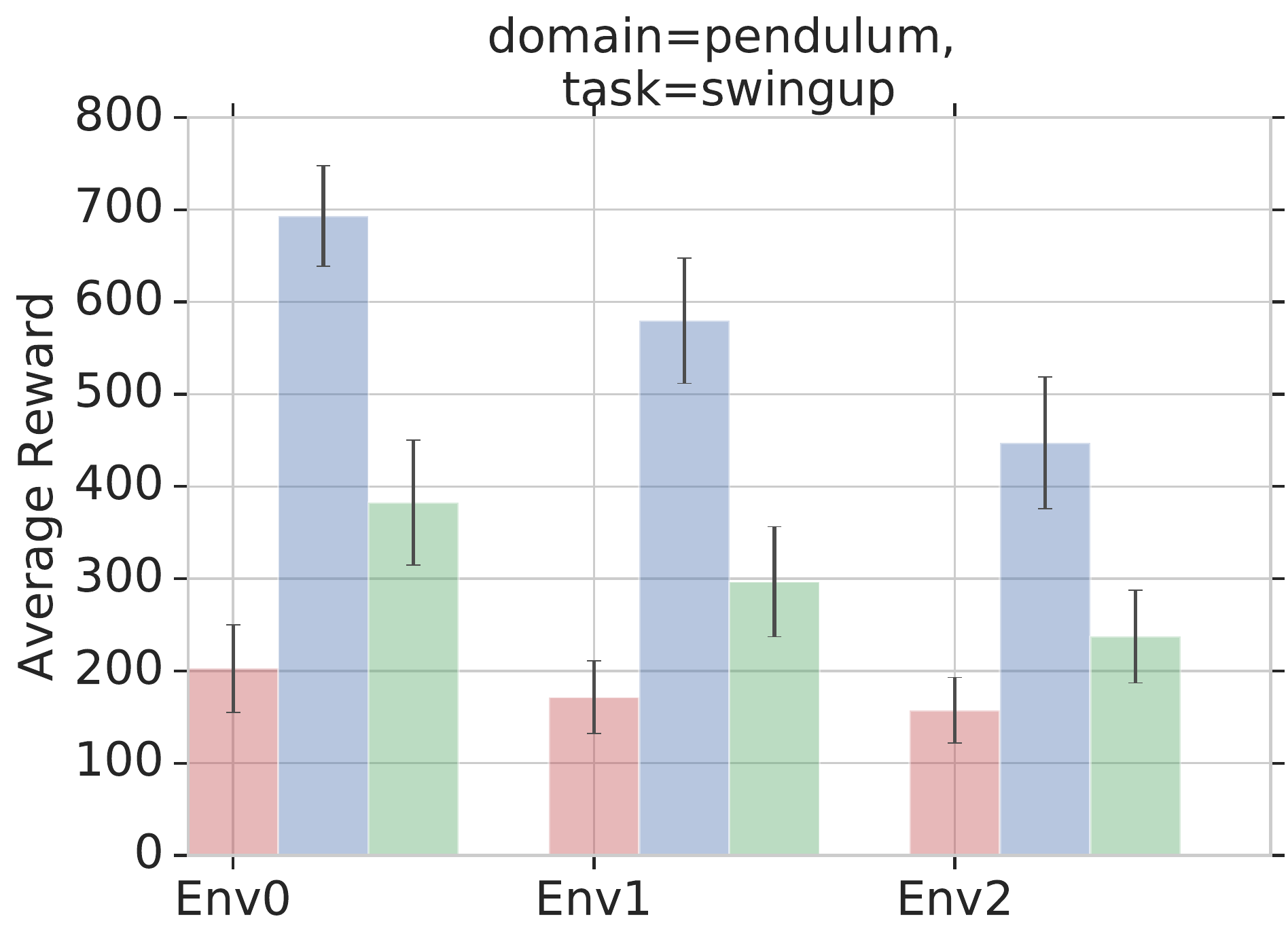}
 }
 \subfigure{
 \includegraphics[scale=\scl]{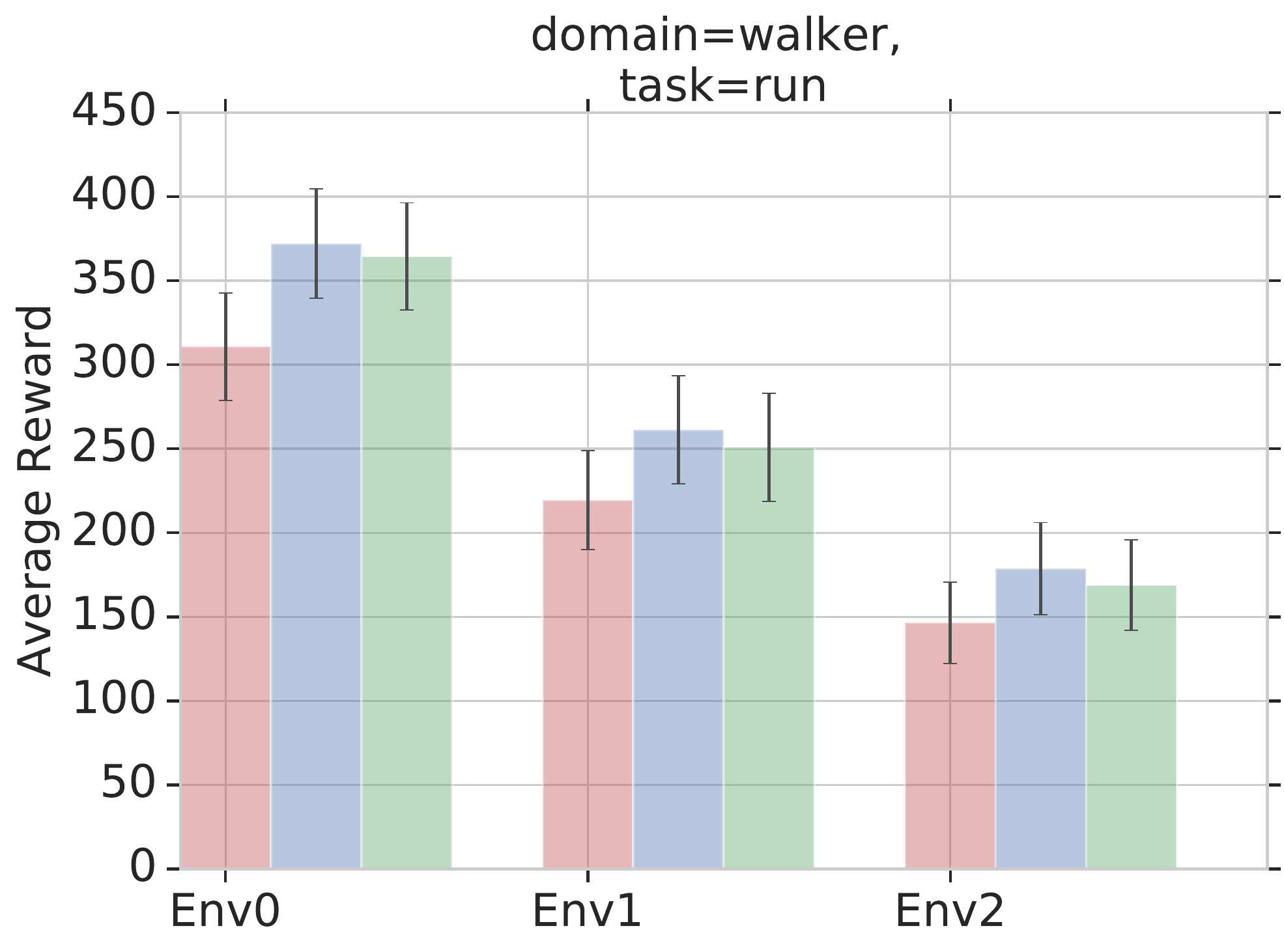}
 }
\caption{
All nine domains showing RE-MPO (blue), SRE-MPO (green) and E-MPO (red). 
}
\label{fig:mpo_kl}
\end{figure*}

\begin{figure*}
\centering
\newcommand{\scl}{0.2}
 \subfigure{
 \includegraphics[scale=\scl]{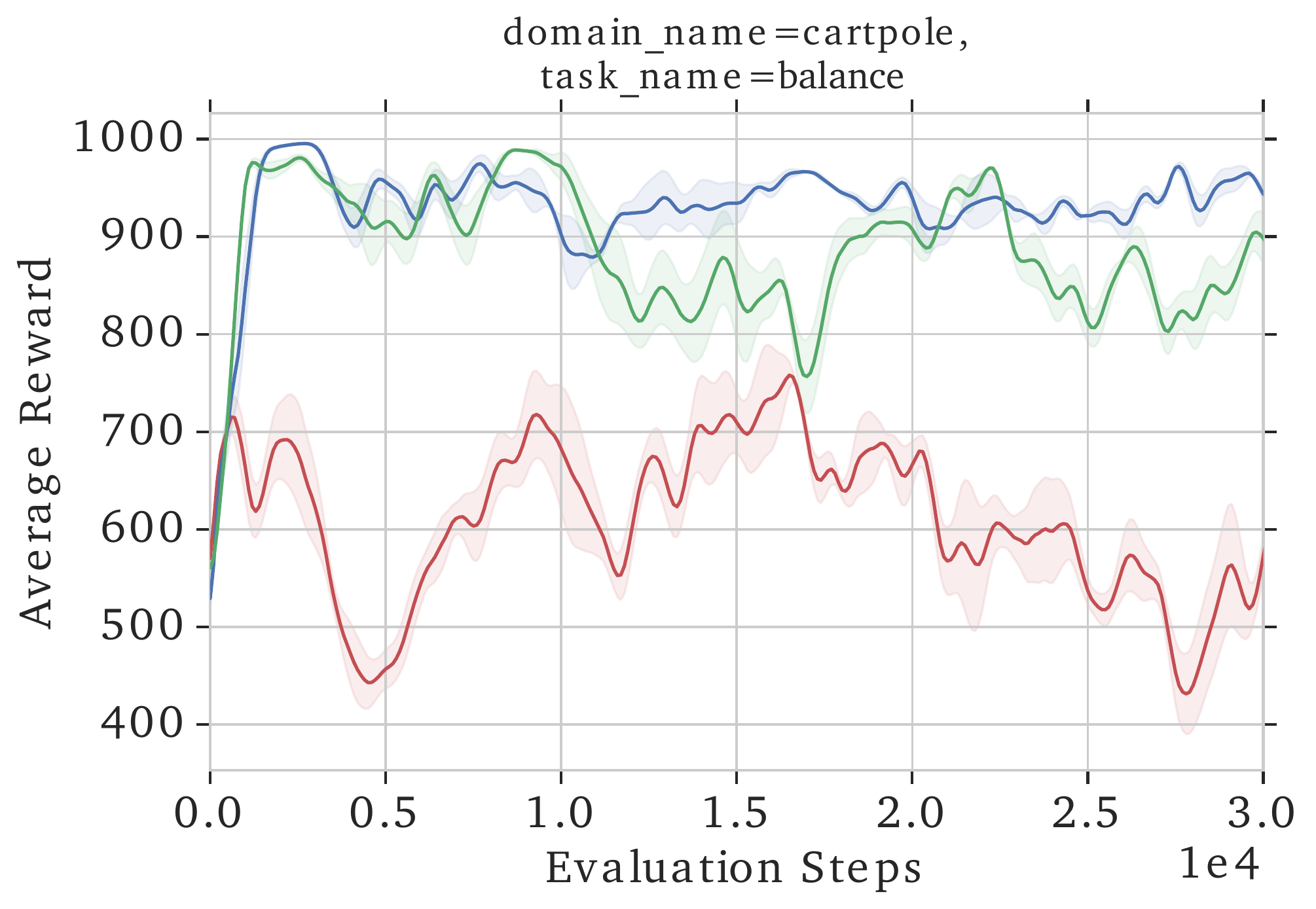}
 }
  \subfigure{
 \includegraphics[scale=\scl]{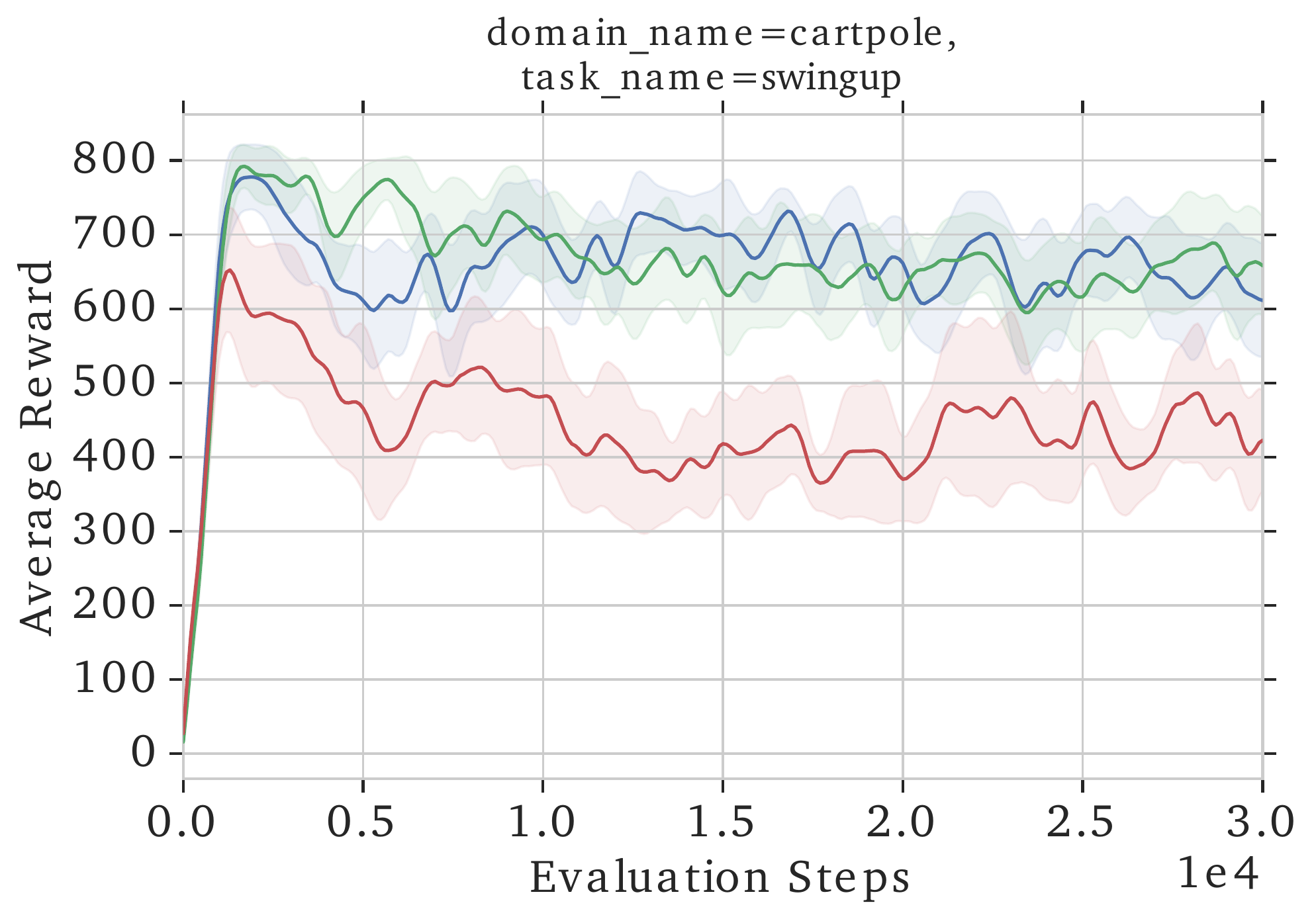}
 }
 \subfigure{
 \includegraphics[scale=\scl]{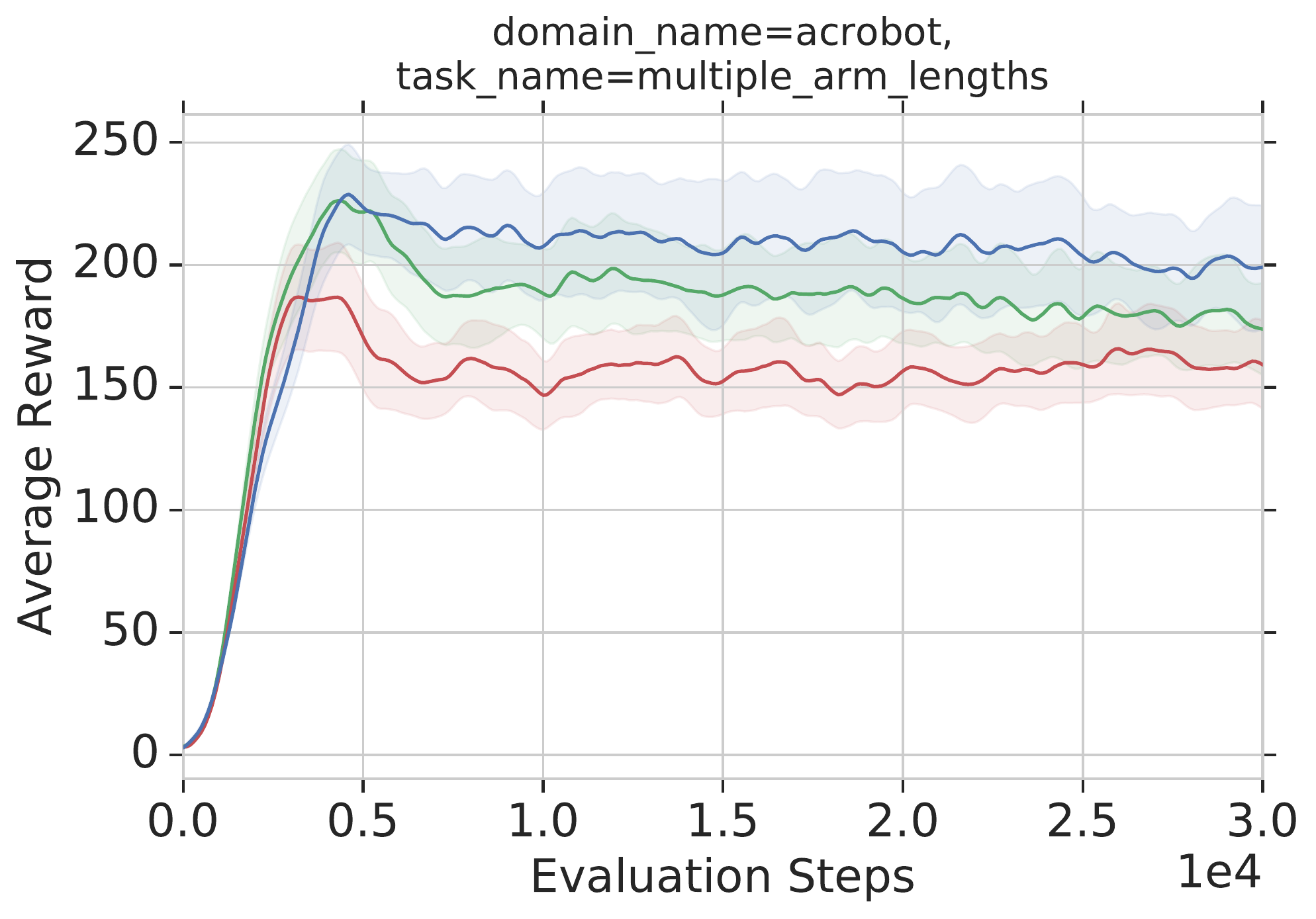}
 }
  \subfigure{
 \includegraphics[scale=\scl]{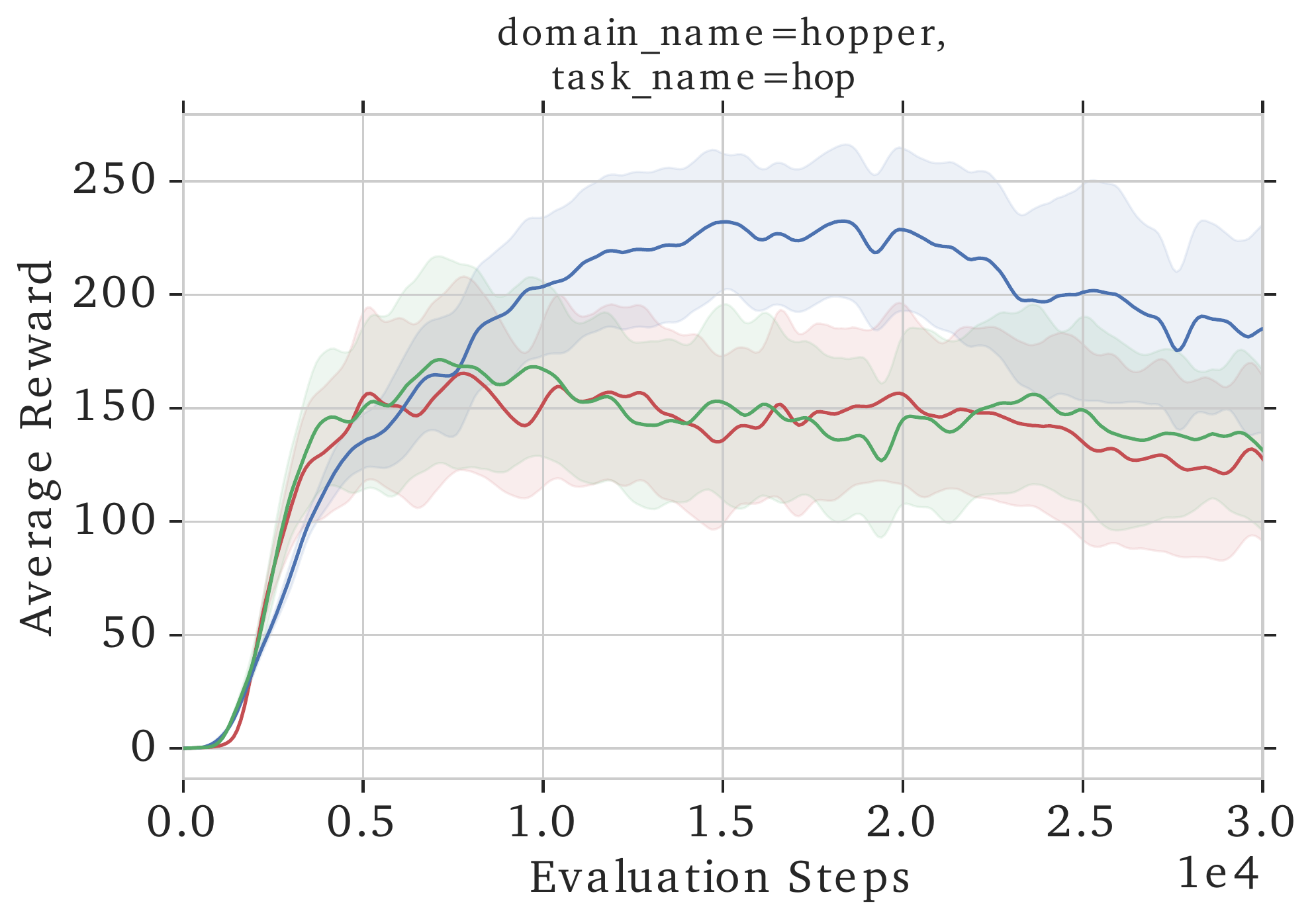}
 }
  \subfigure{
 \includegraphics[scale=\scl]{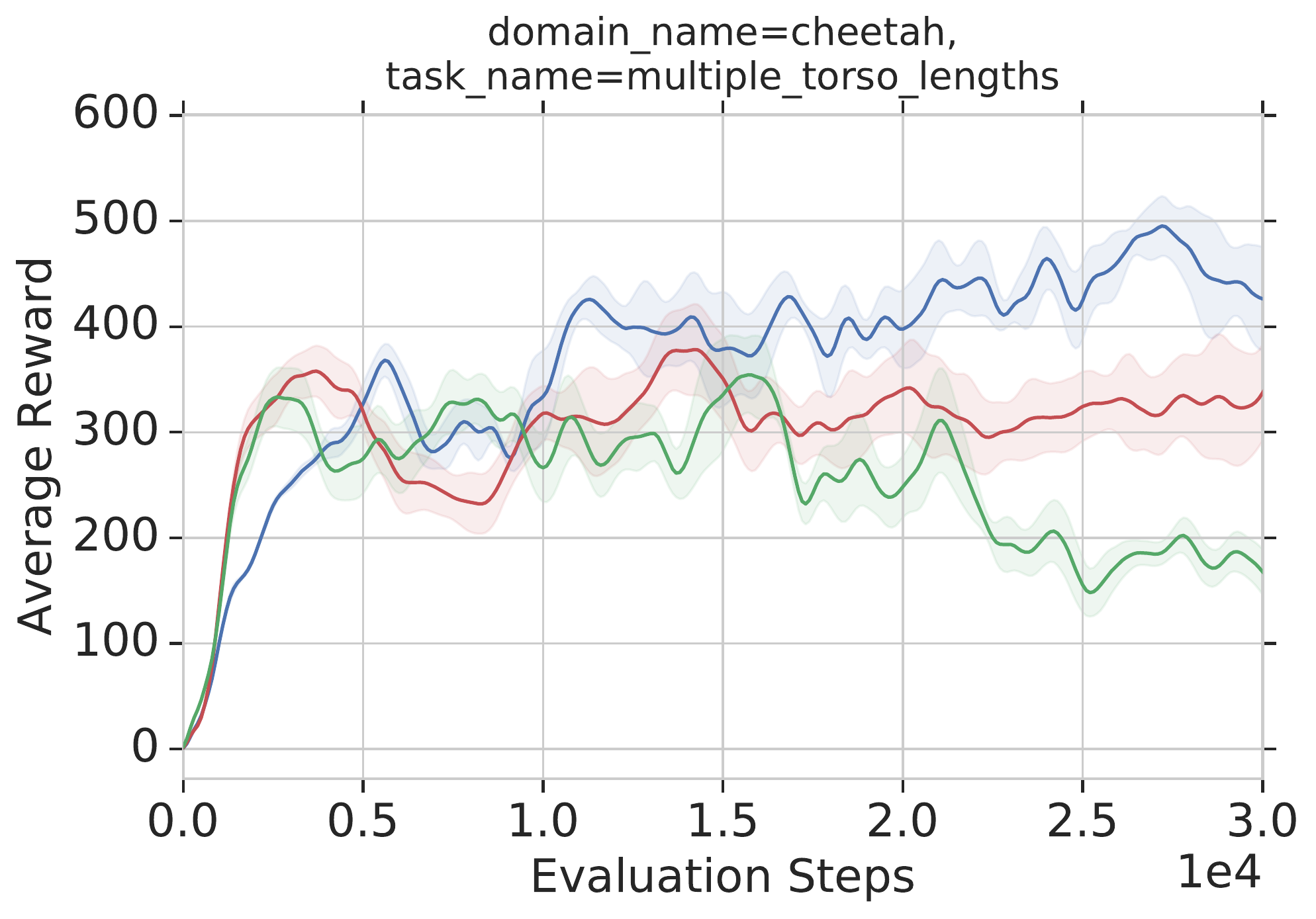}
 }
 \subfigure{
 \includegraphics[scale=\scl]{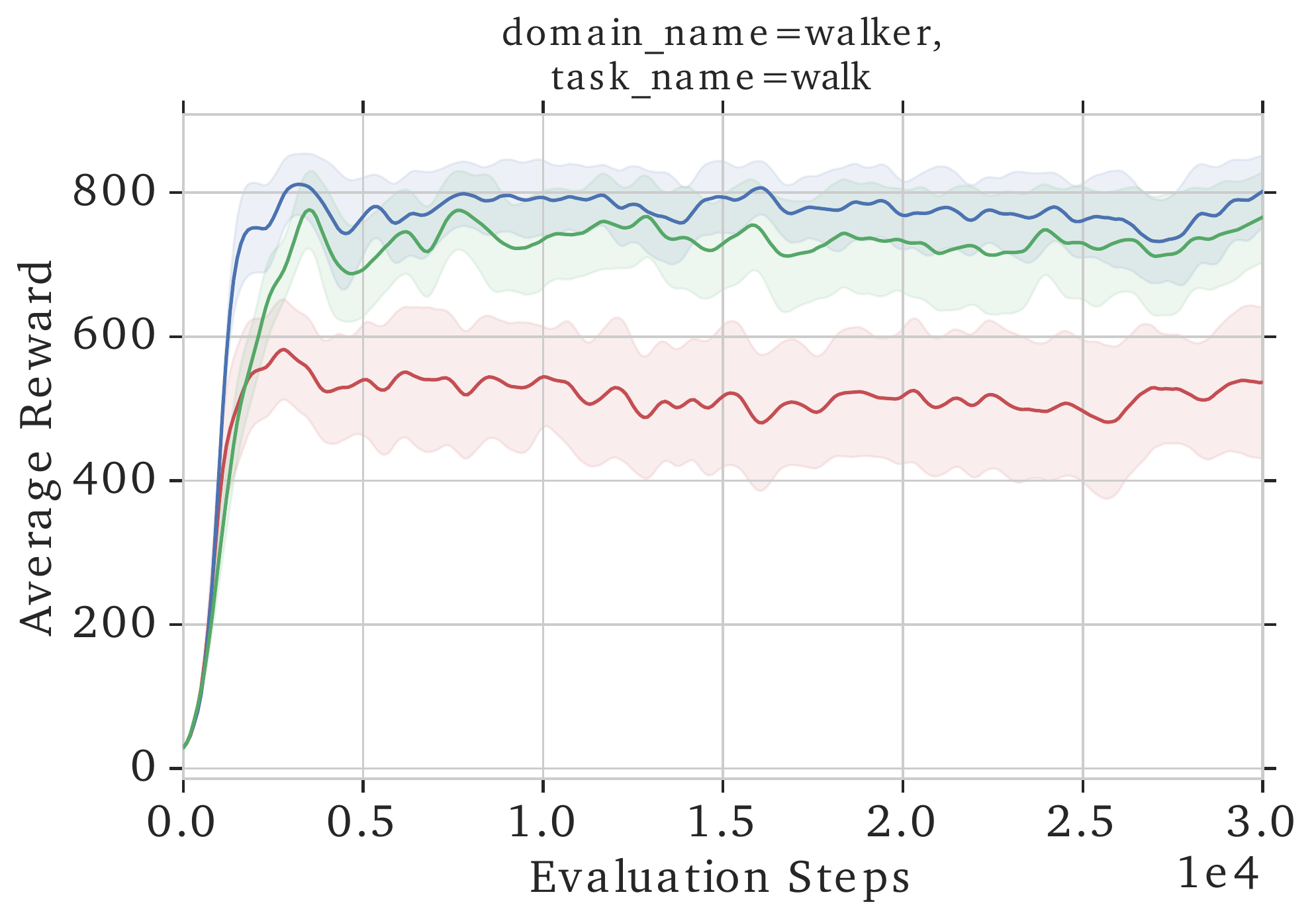}
 }
   \subfigure{
 \includegraphics[scale=\scl]{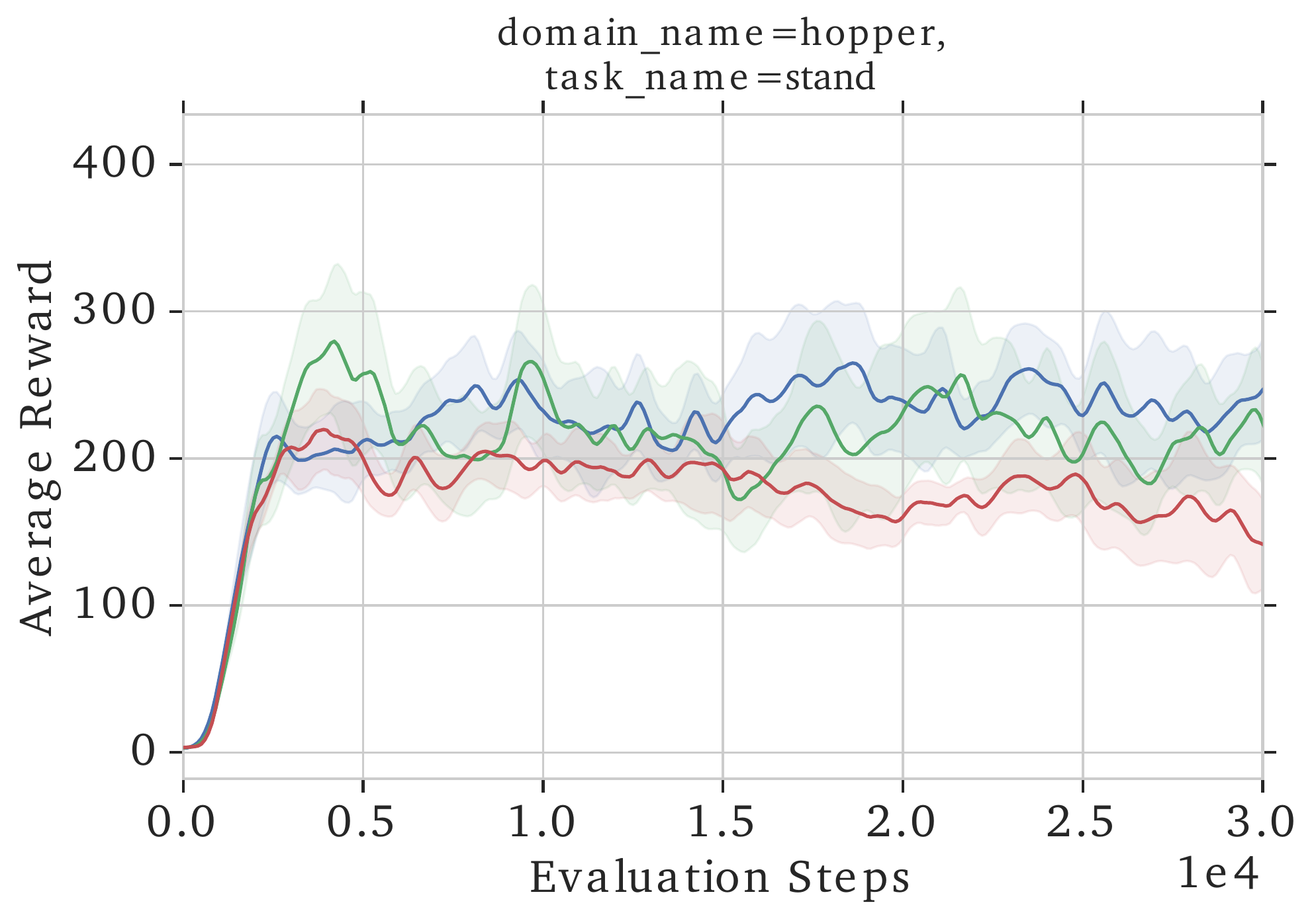}
 }
  \subfigure{
 \includegraphics[scale=\scl]{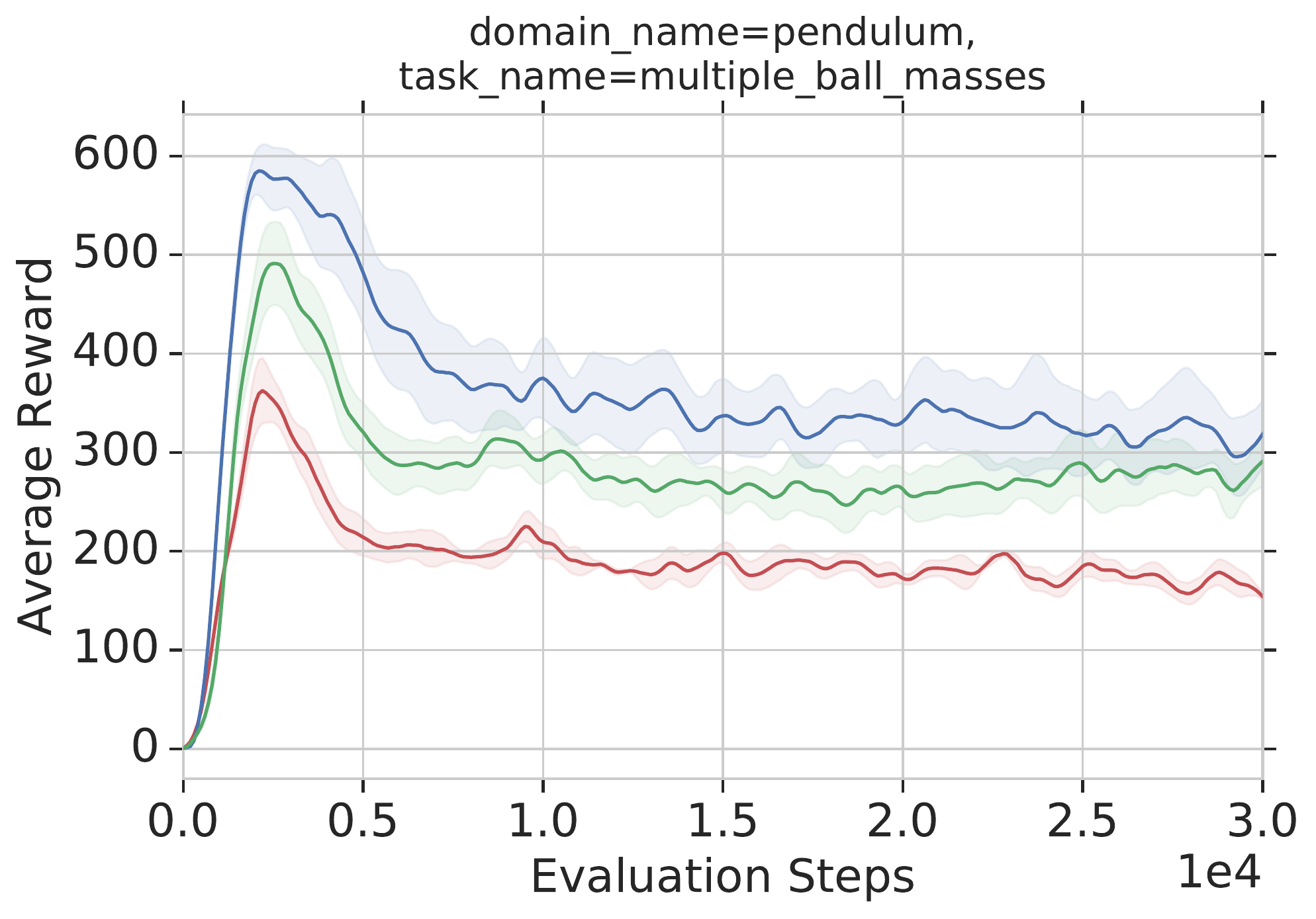}
 }
 \subfigure{
 \includegraphics[scale=\scl]{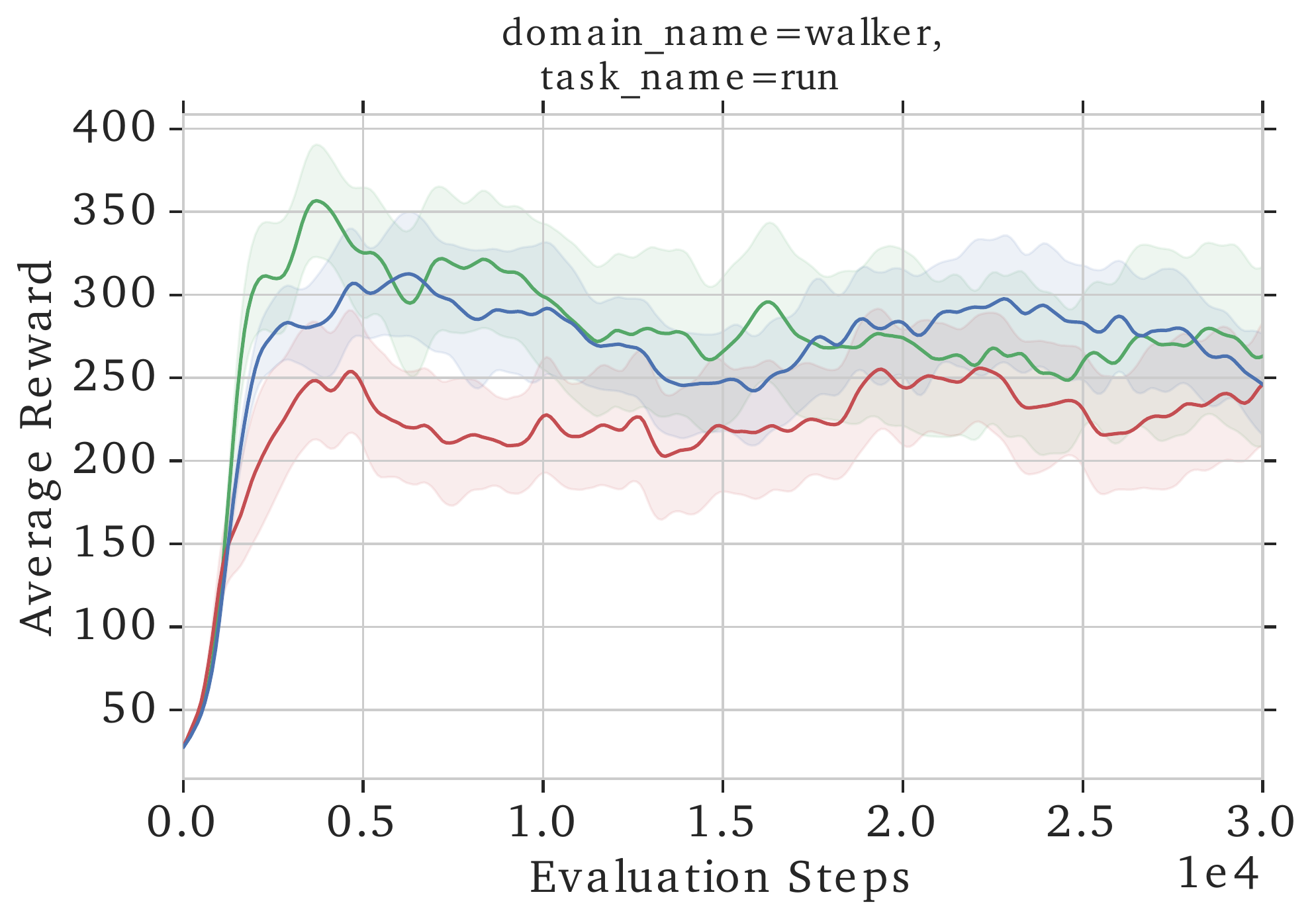}
 }
\caption{
All nine domains showing RE-MPO (blue), SRE-MPO (green) and E-MPO (red) as a function of evaluation steps during training. 
}
\label{fig:mpo_kl_eval_agg}
\end{figure*}

\begin{figure*}
\centering
\newcommand{\scl}{0.22}
 \subfigure{
 \includegraphics[scale=\scl]{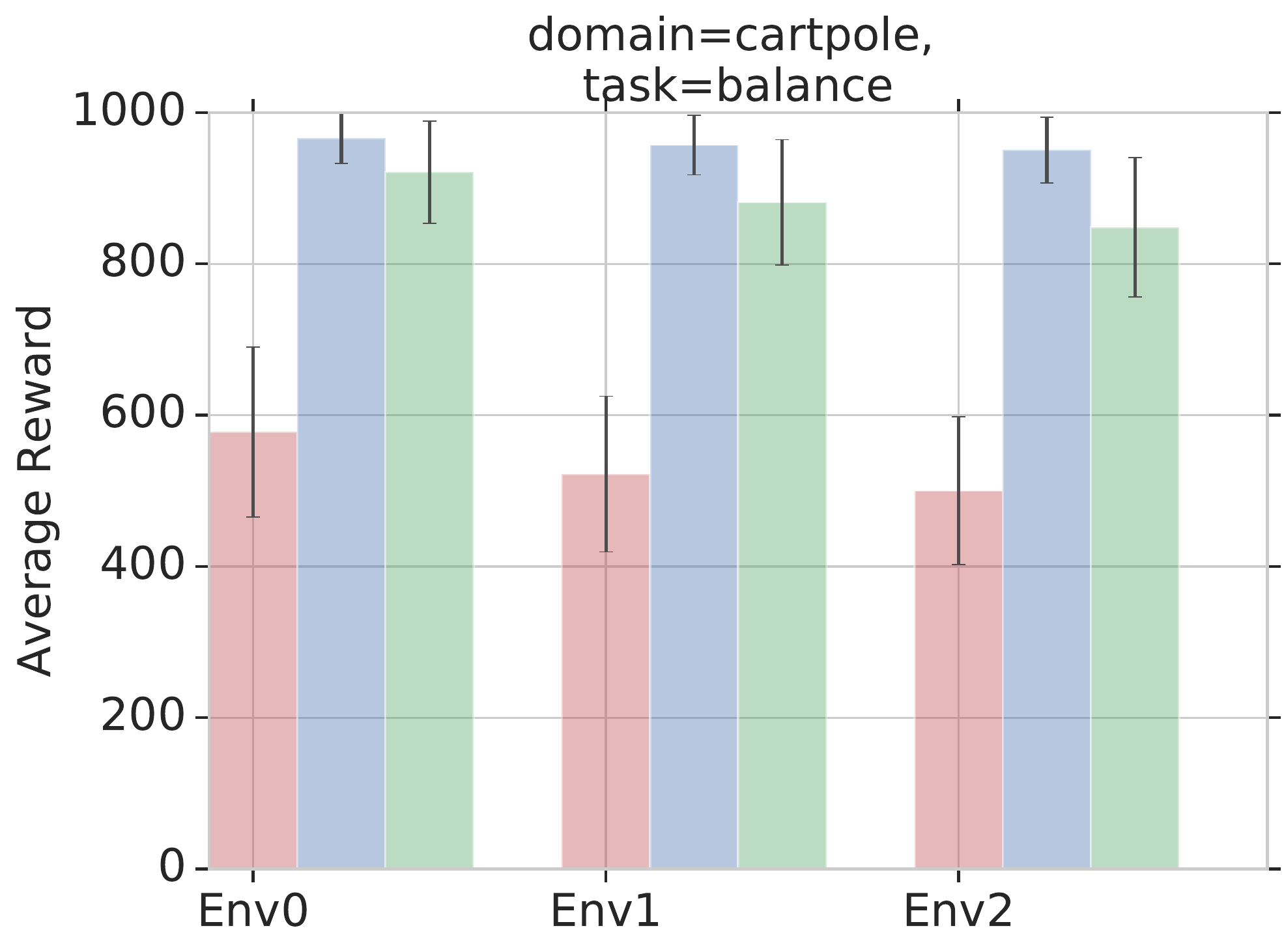}
 }
  \subfigure{
 \includegraphics[scale=\scl]{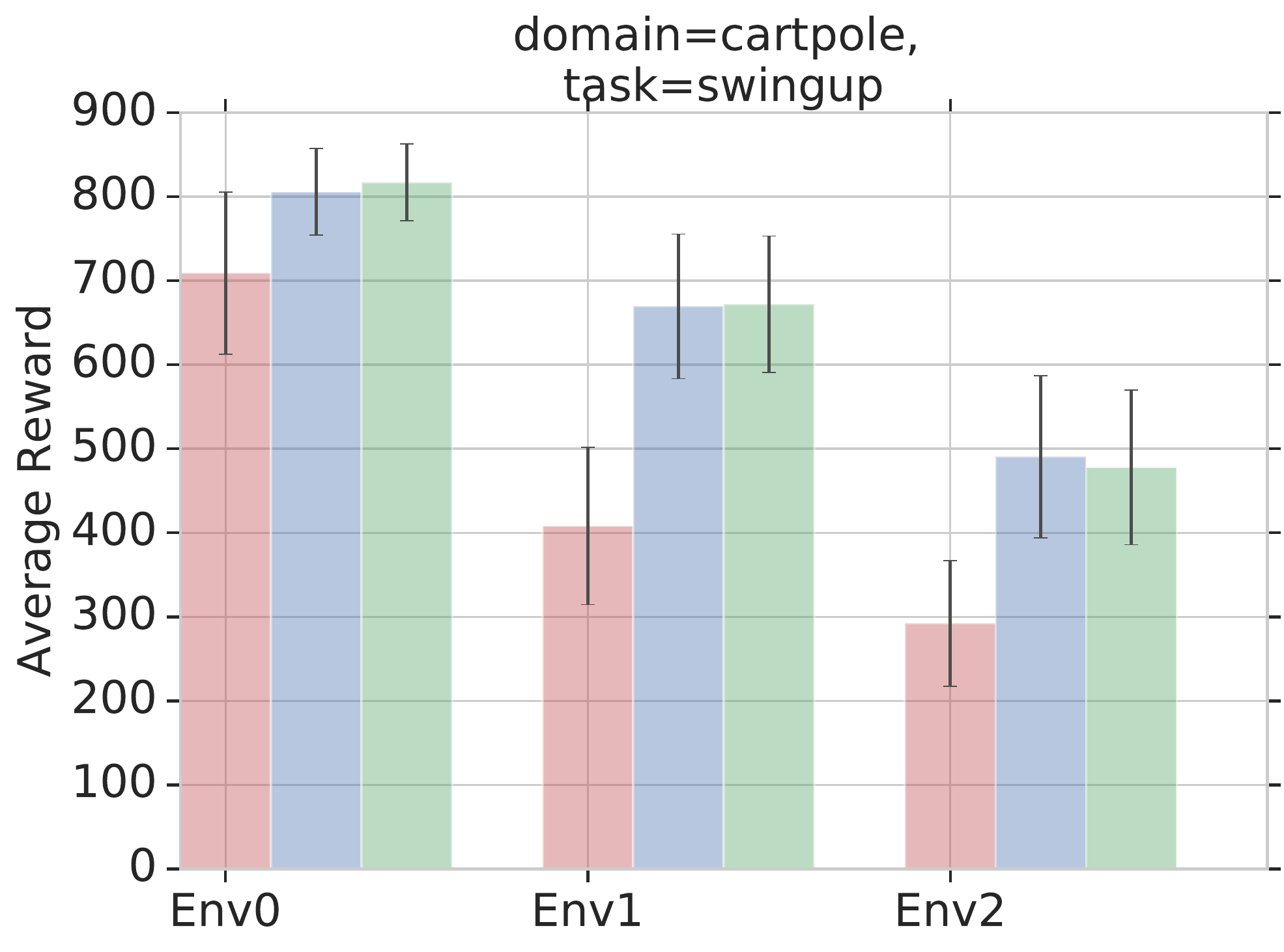}
 }
 \subfigure{
 \includegraphics[scale=\scl]{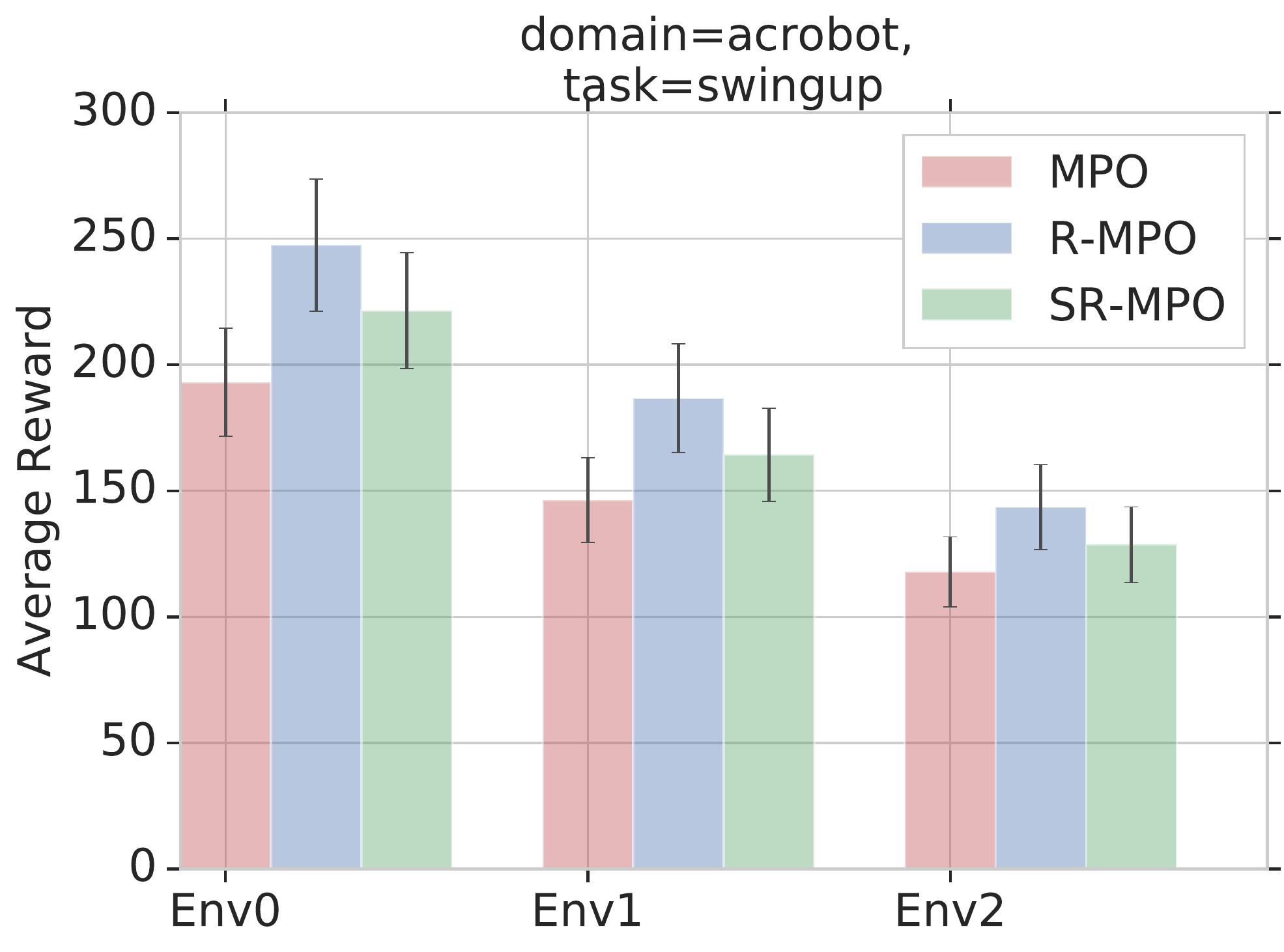}
 }
  \subfigure{
 \includegraphics[scale=\scl]{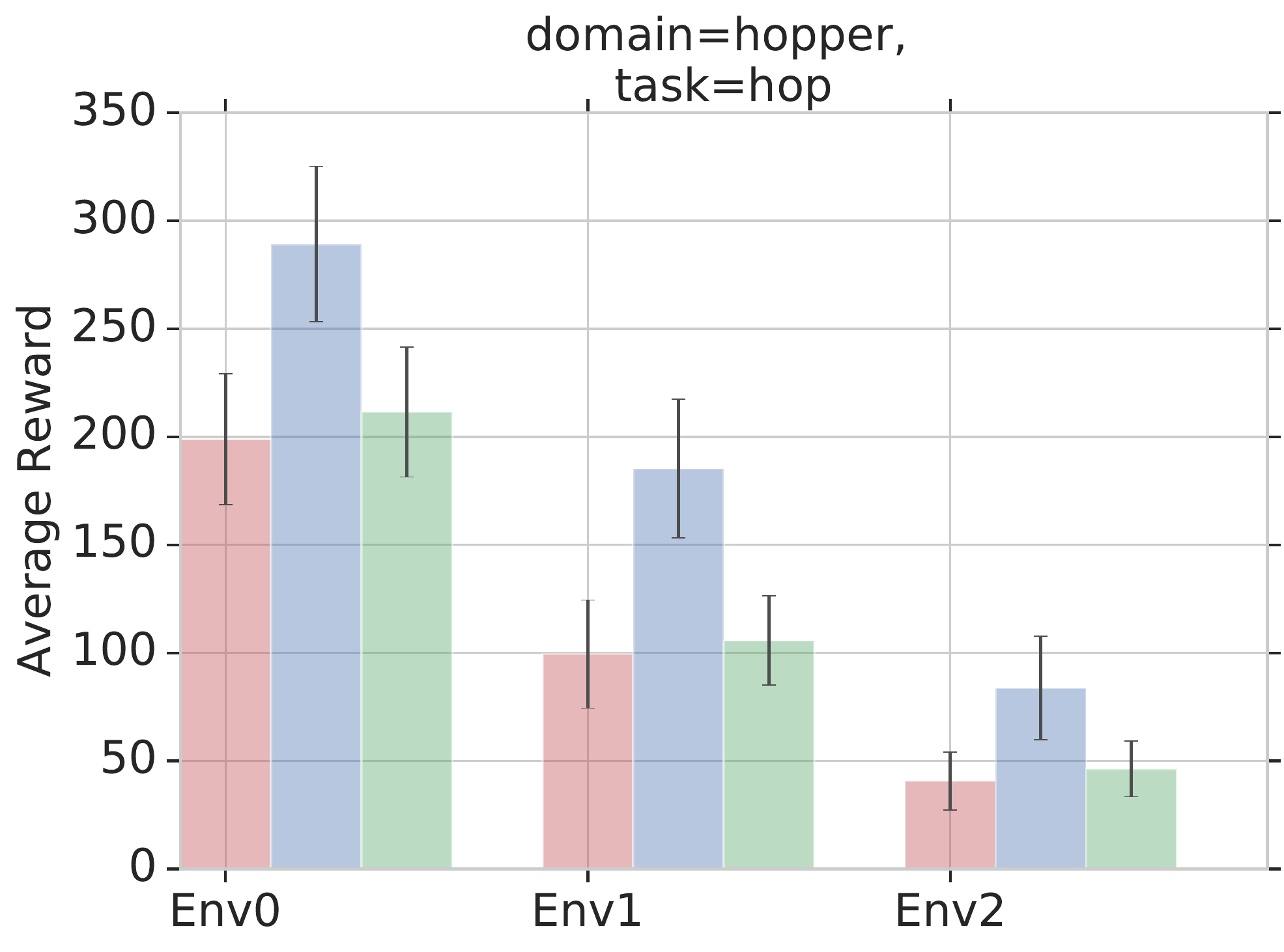}
 }
  \subfigure{
 \includegraphics[scale=\scl]{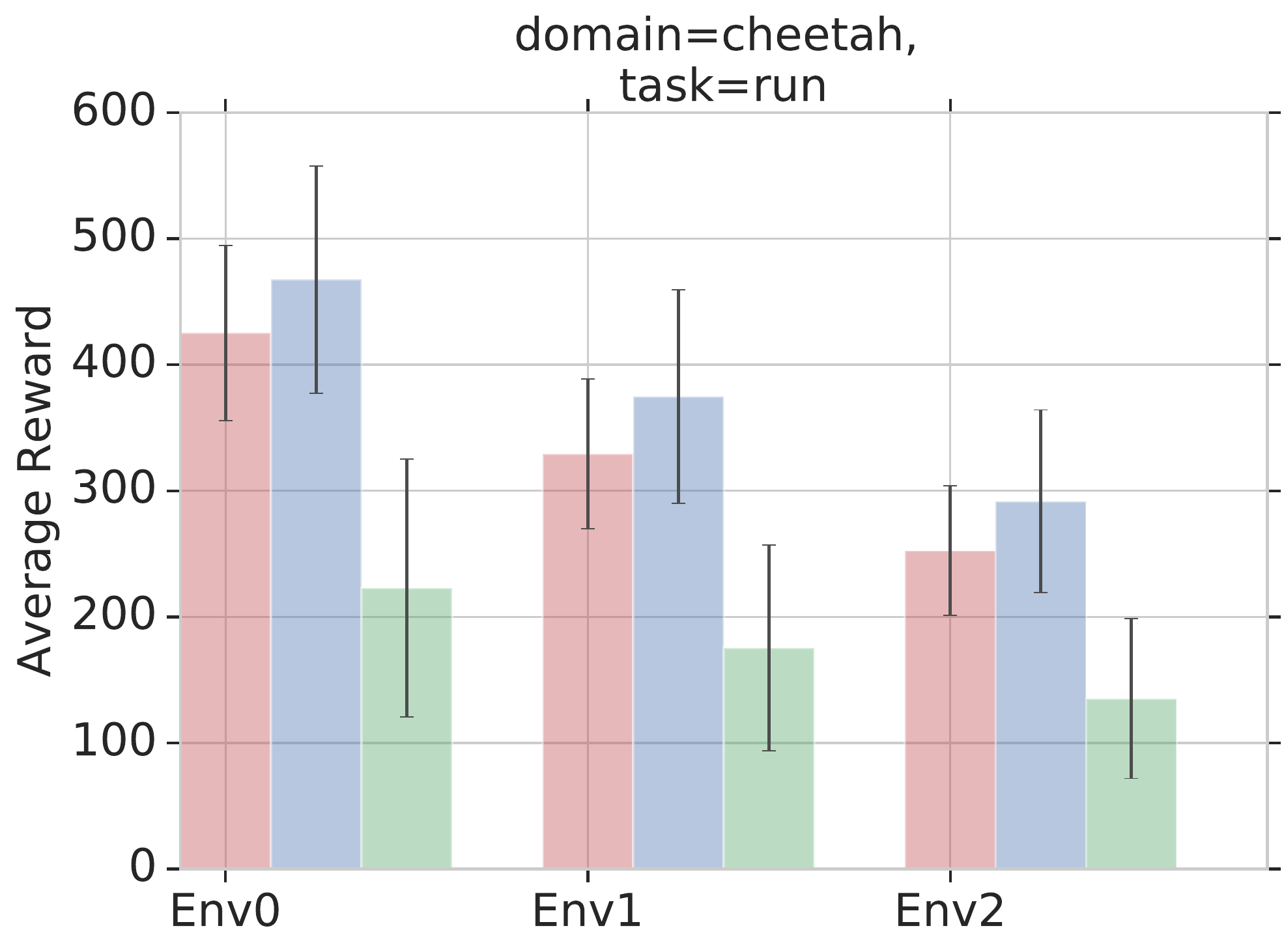}
 }
 \subfigure{
 \includegraphics[scale=\scl]{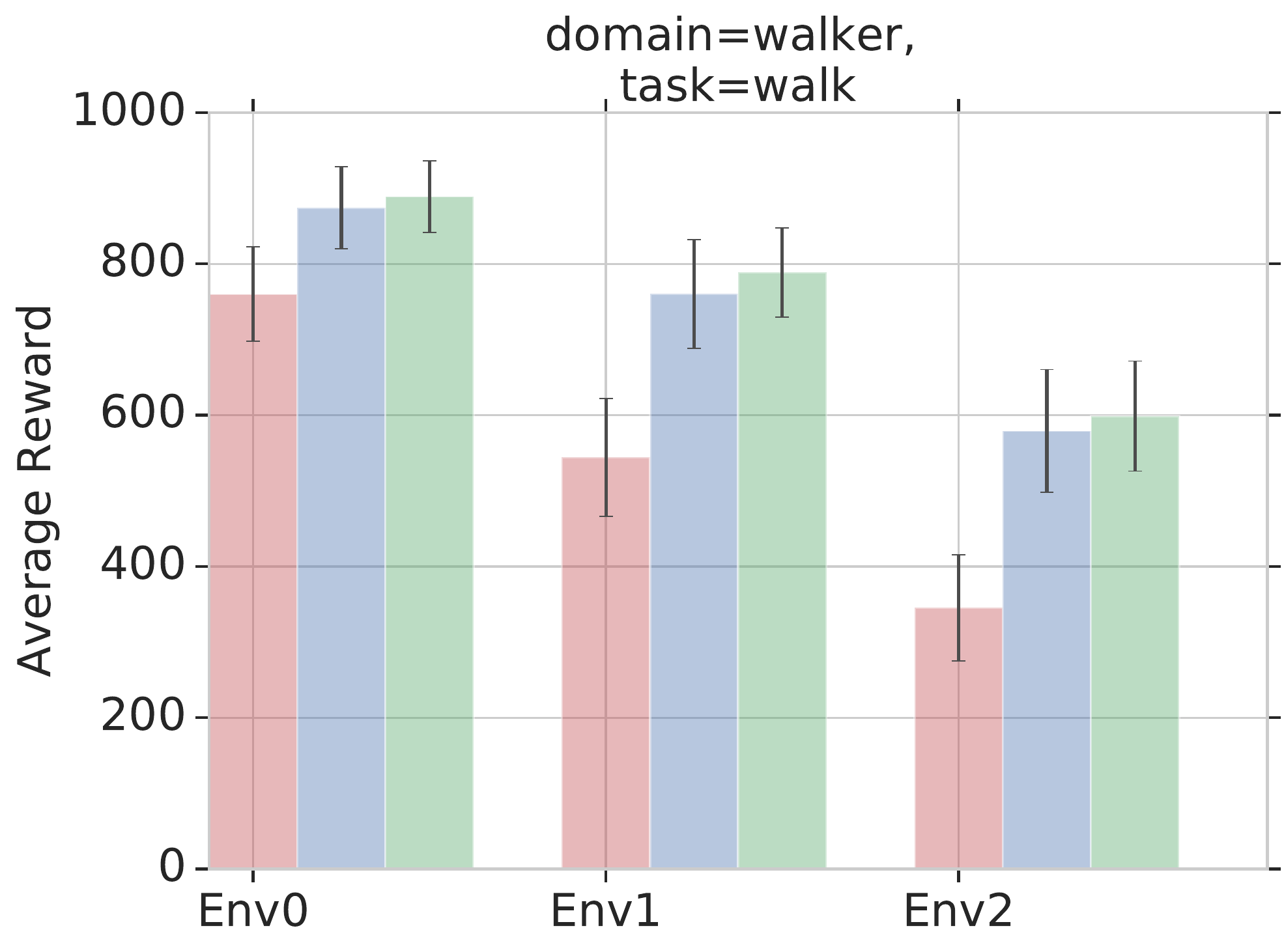}
 }
   \subfigure{
 \includegraphics[scale=\scl]{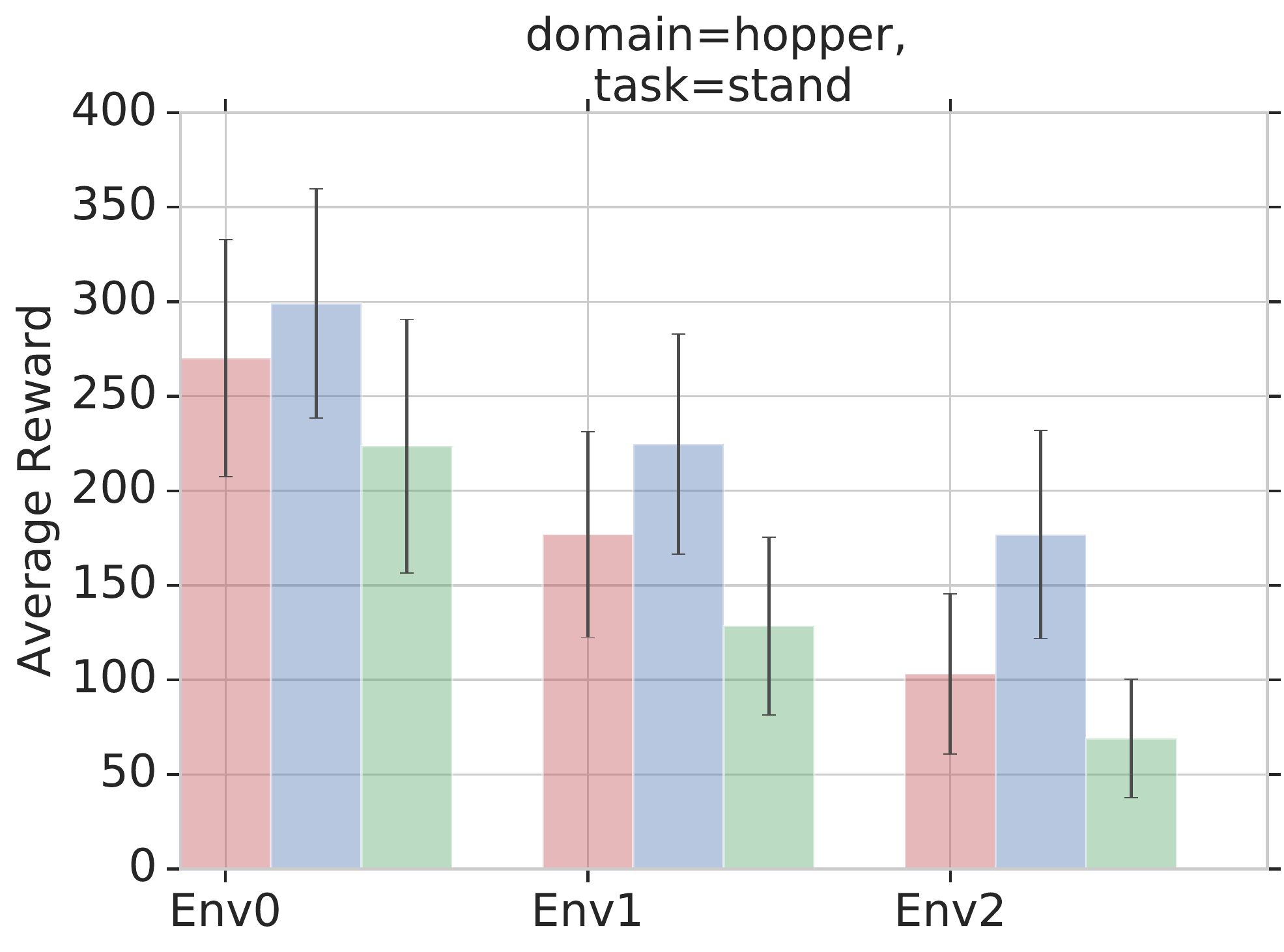}
 }
  \subfigure{
 \includegraphics[scale=\scl]{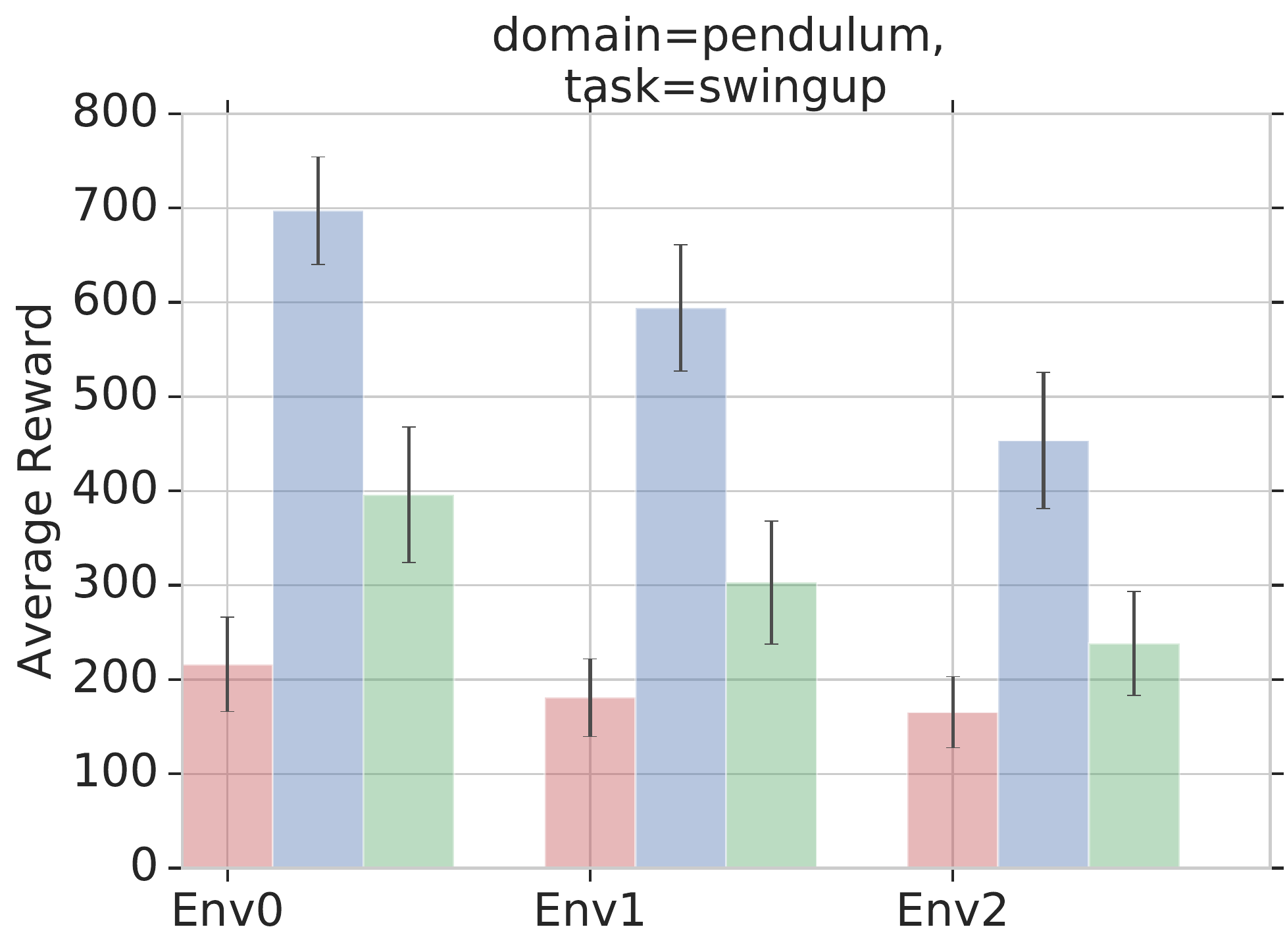}
 }
 \subfigure{
 \includegraphics[scale=\scl]{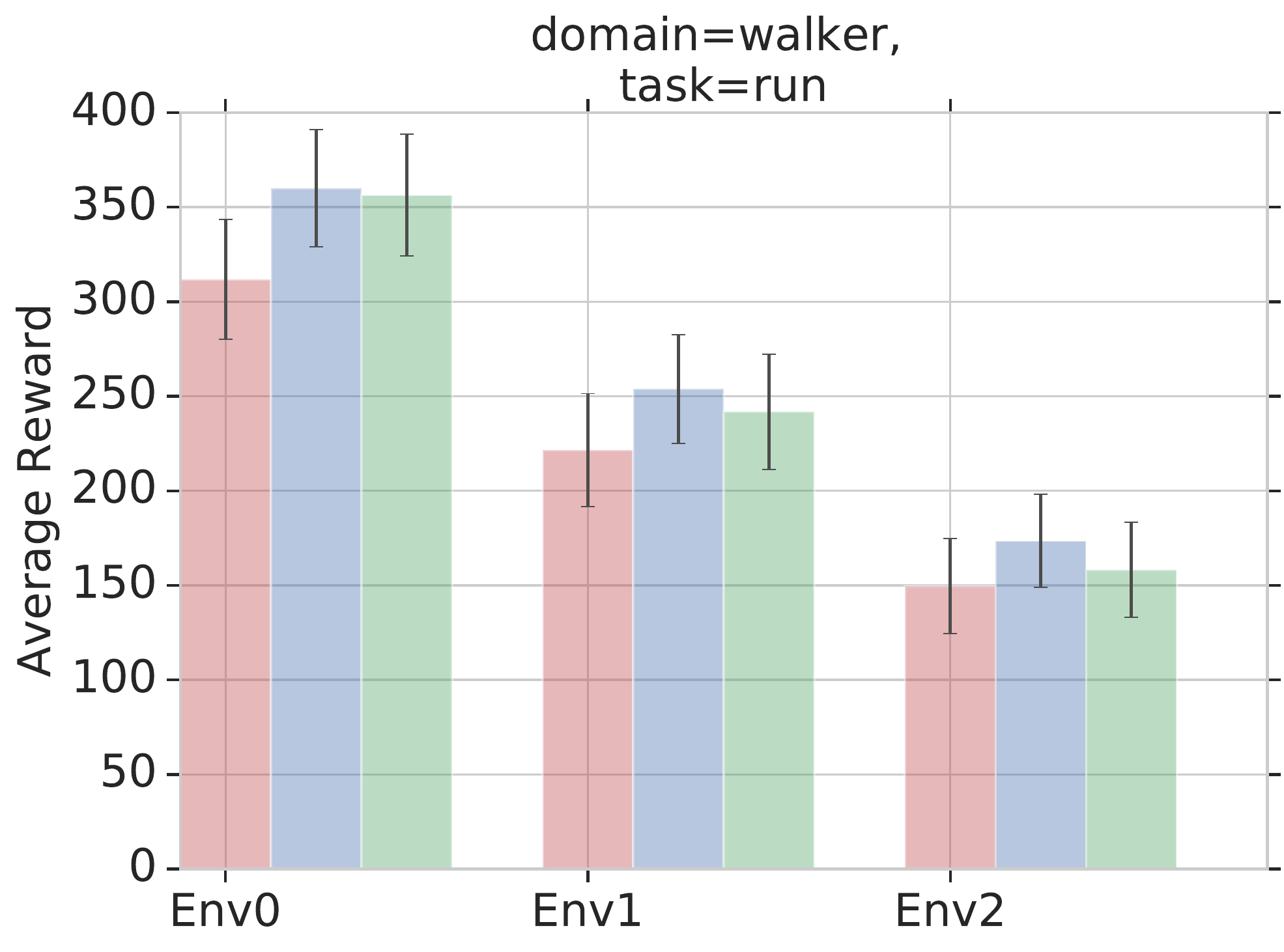}
 }
\caption{
All nine domains showing R-MPO (blue), SR-MPO (green) and MPO (red).
}
\label{fig:mpo_no_kl}
\end{figure*}

\begin{figure*}
\centering
\newcommand{\scl}{0.2}
 \subfigure{
 \includegraphics[scale=\scl]{./figures/mpo_kl/eval_agg/cartpole_balance}
 }
  \subfigure{
 \includegraphics[scale=\scl]{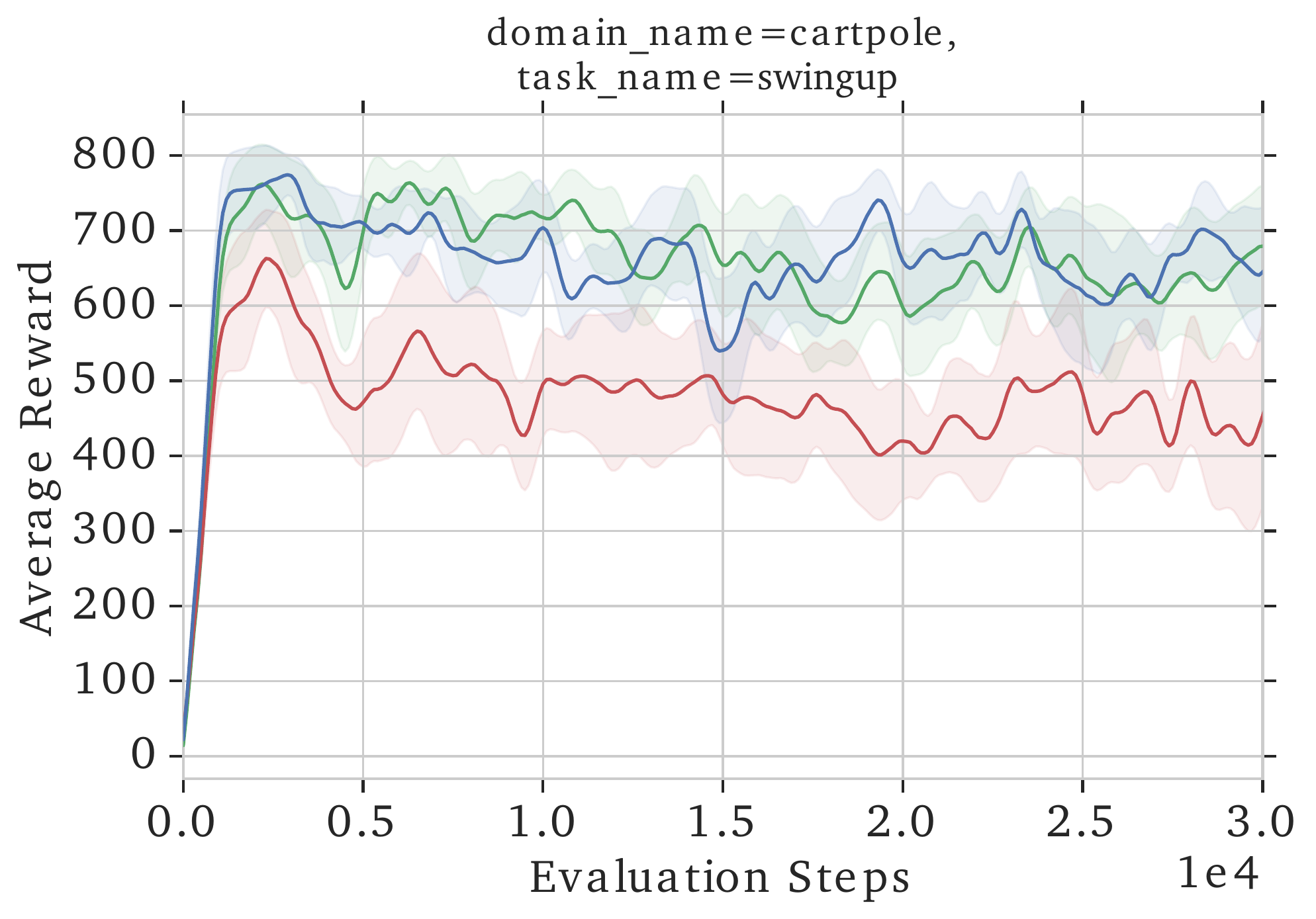}
 }
 \subfigure{
 \includegraphics[scale=\scl]{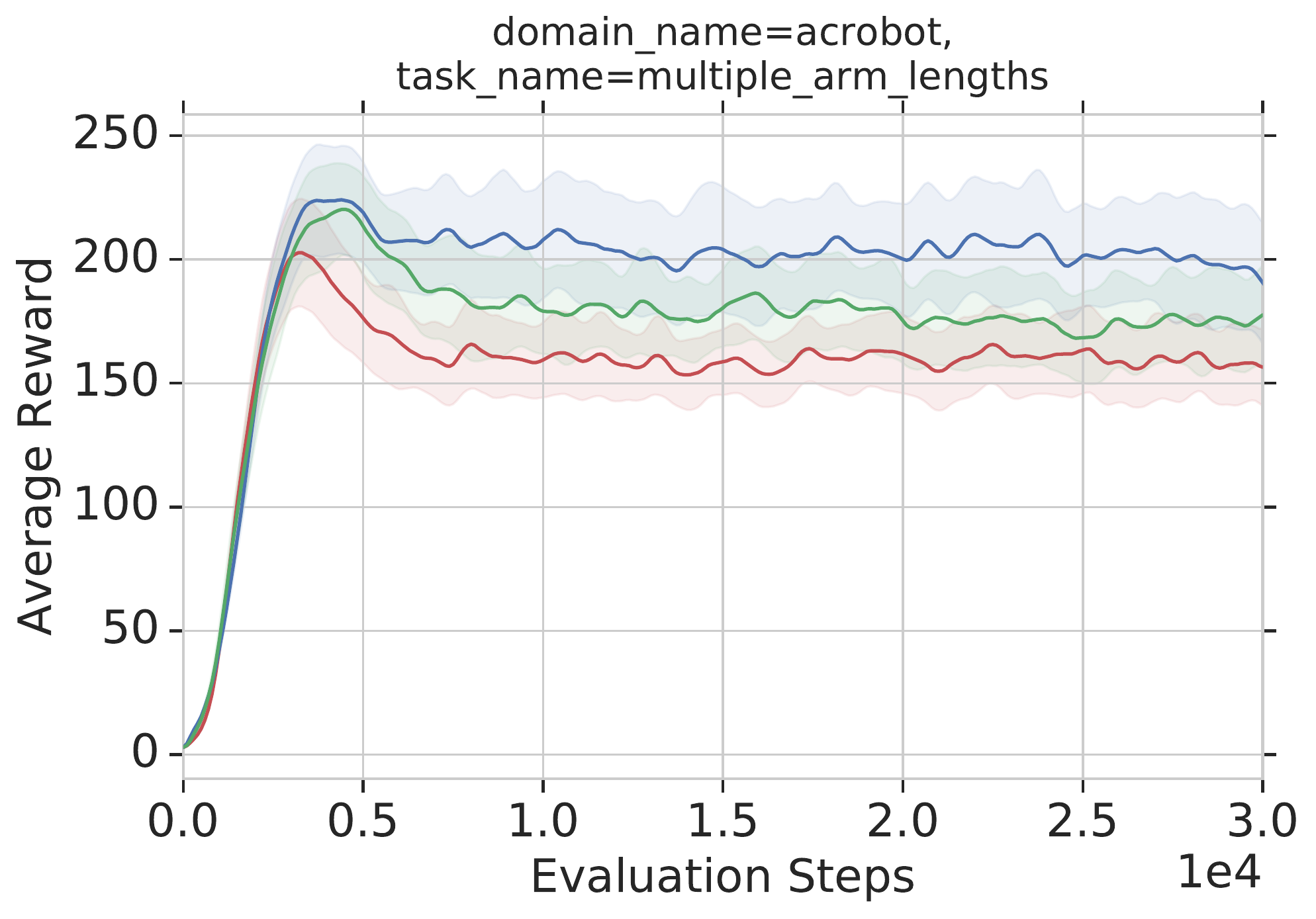}
 }
  \subfigure{
 \includegraphics[scale=\scl]{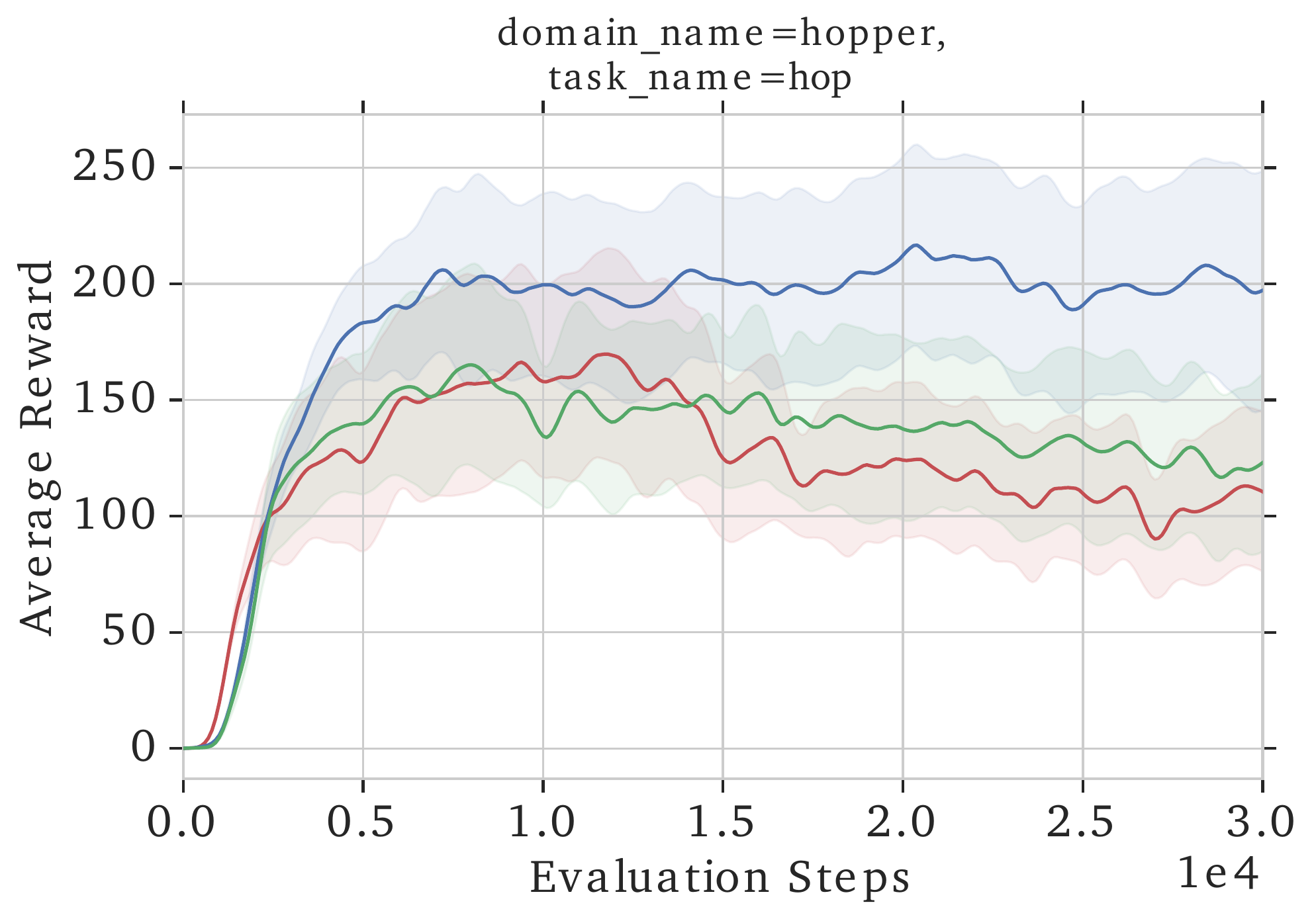}
 }
  \subfigure{
 \includegraphics[scale=\scl]{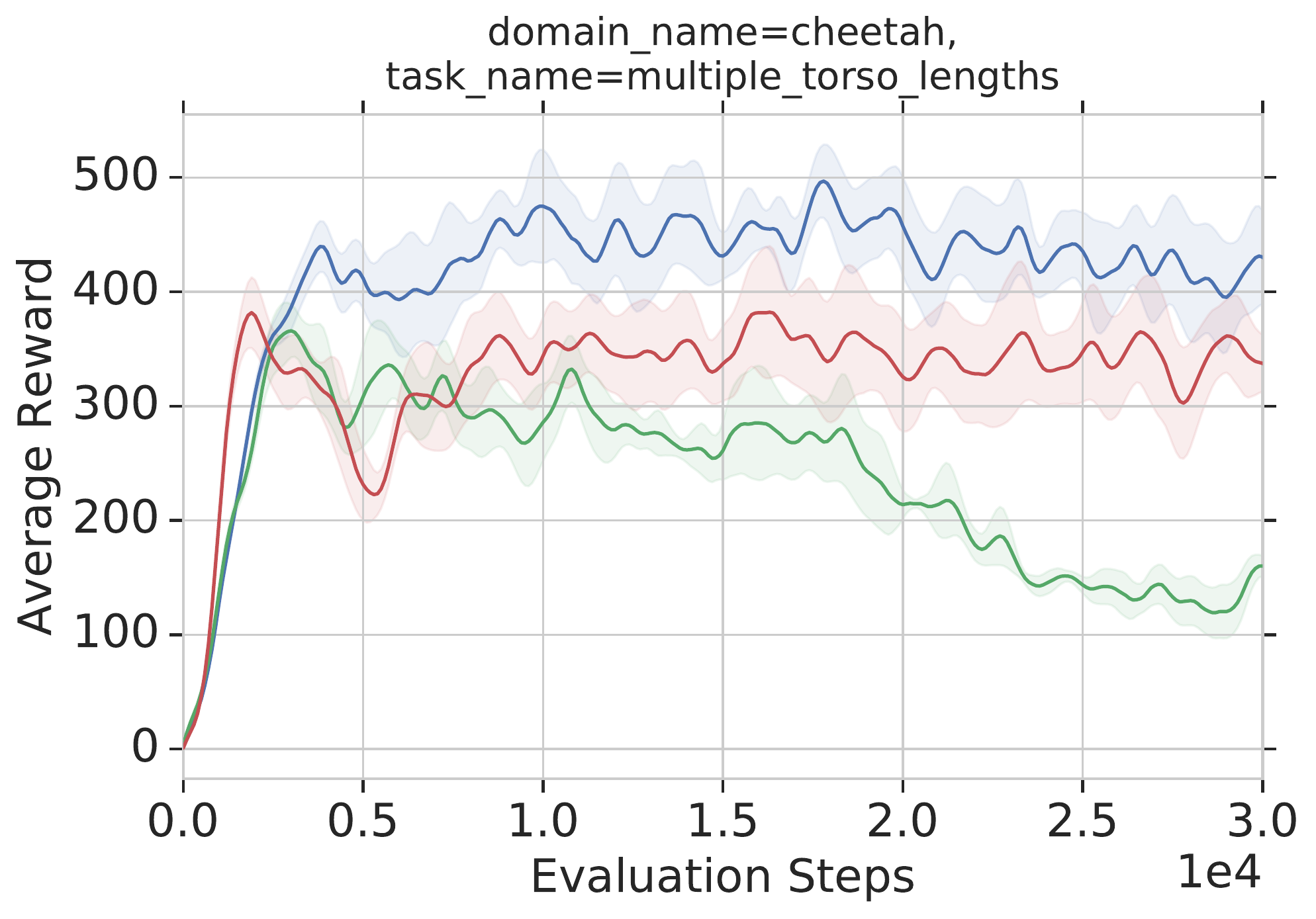}
 }
 \subfigure{
 \includegraphics[scale=\scl]{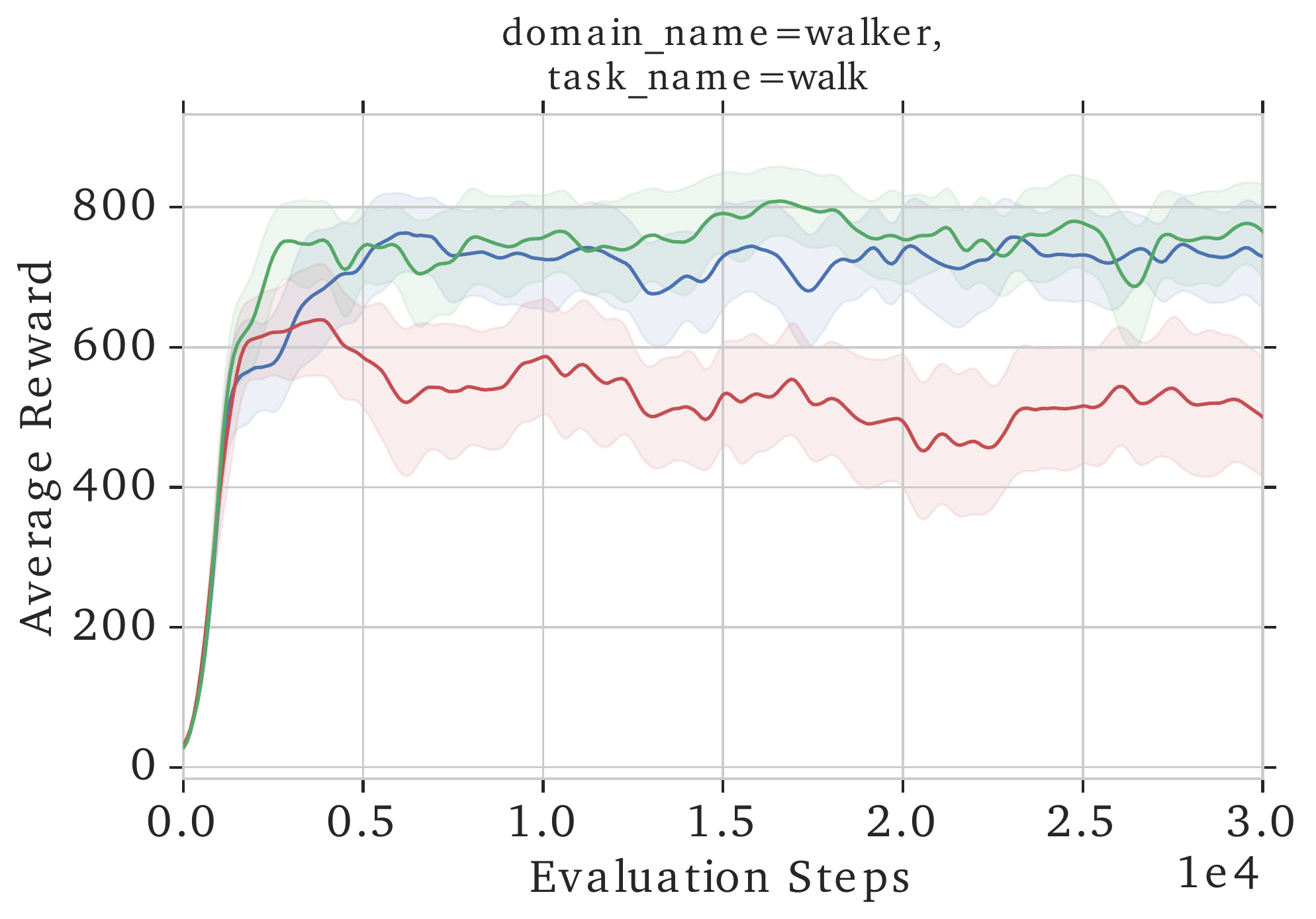}
 }
   \subfigure{
 \includegraphics[scale=\scl]{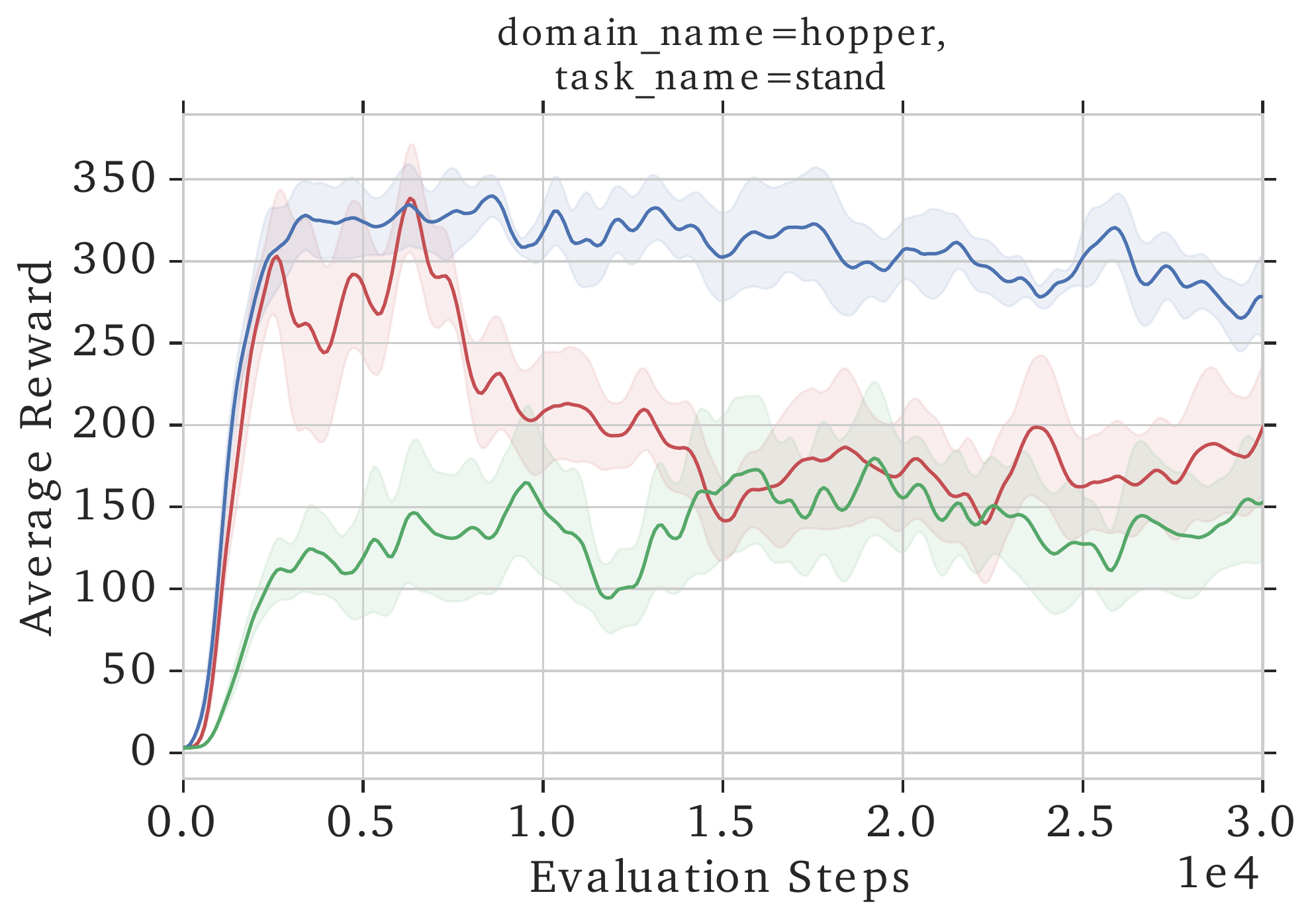}
 }
  \subfigure{
 \includegraphics[scale=\scl]{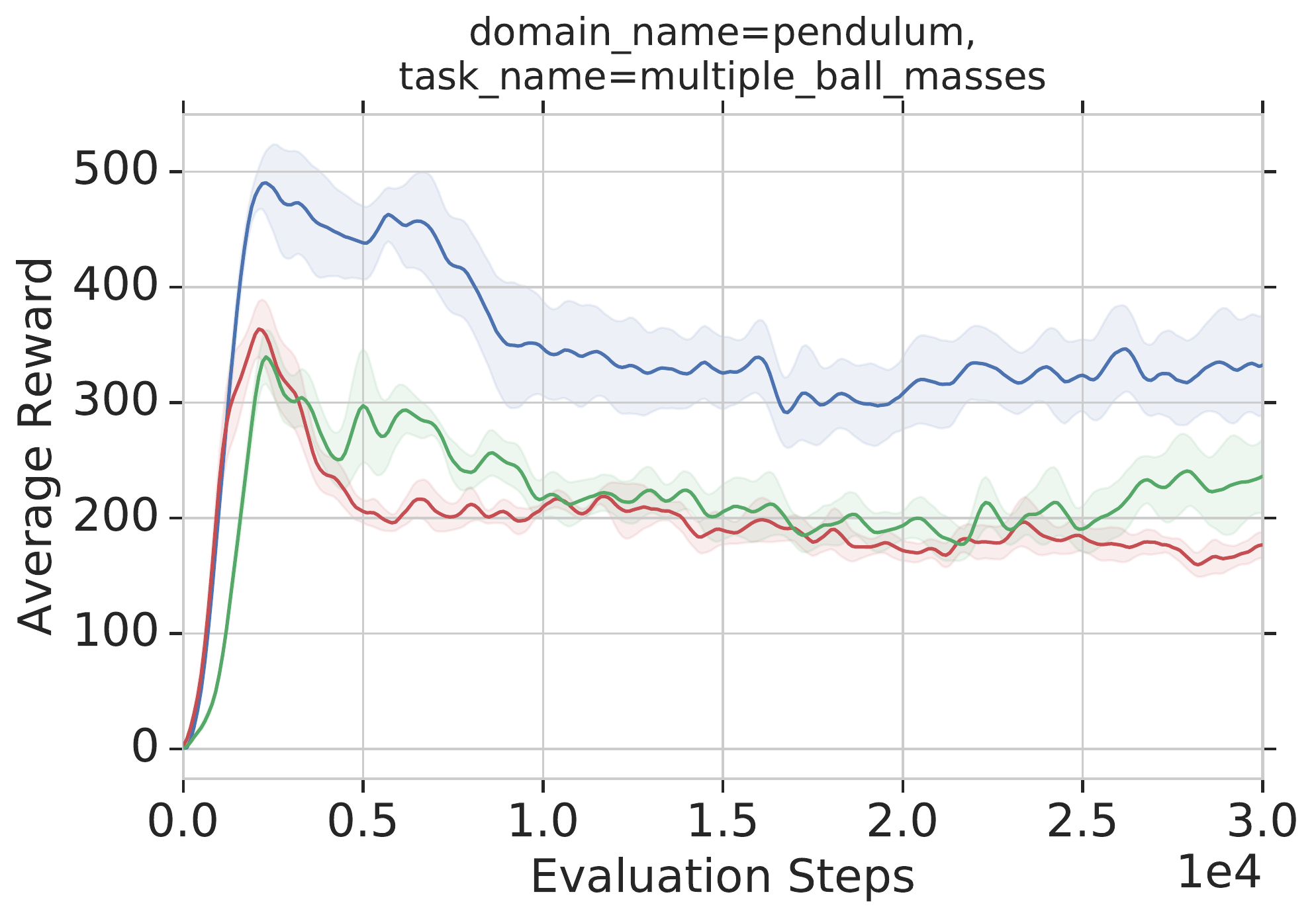}
 }
 \subfigure{
 \includegraphics[scale=\scl]{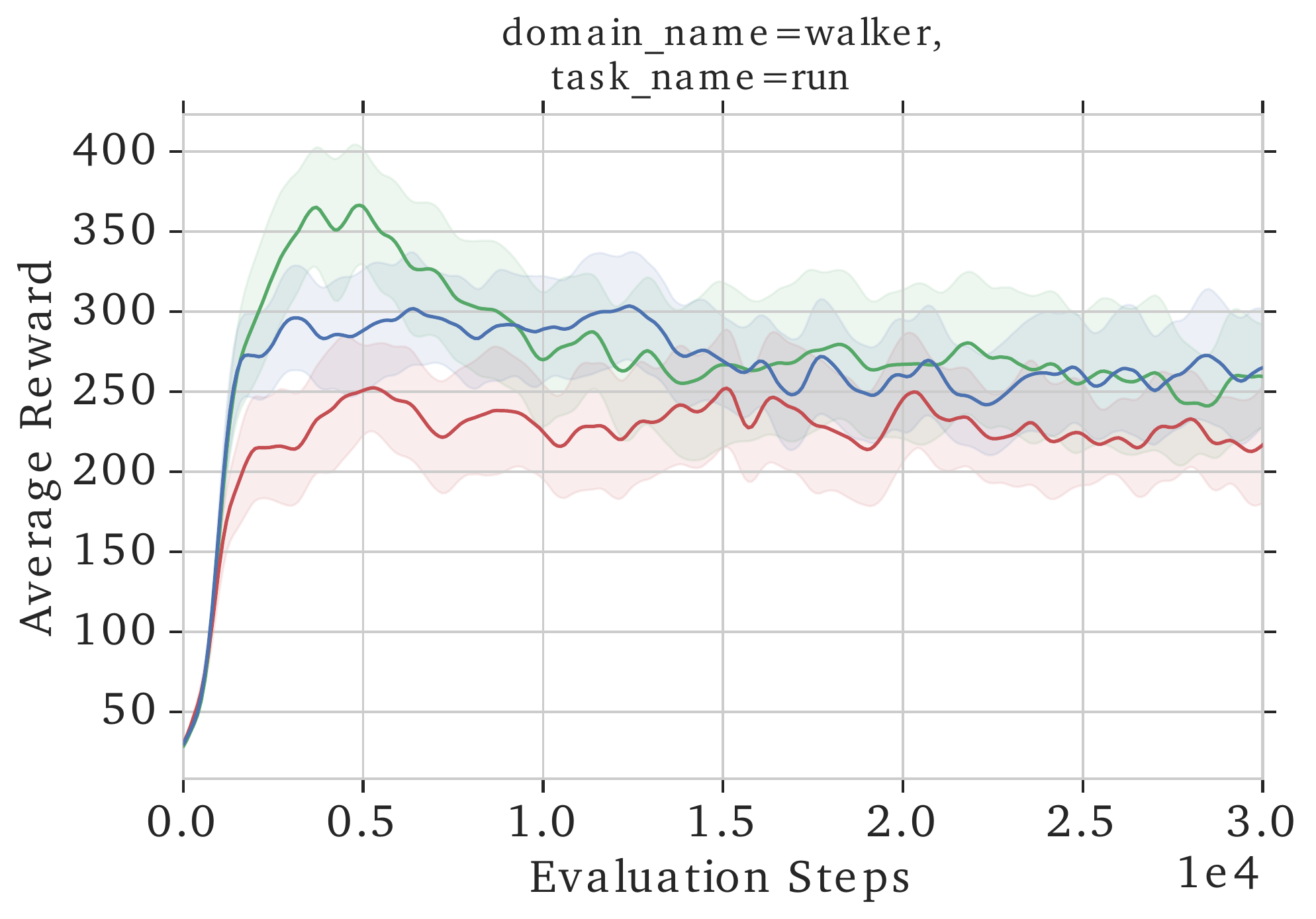}
 }
\caption{
All nine domains showing R-MPO (blue), SR-MPO (green) and MPO (red) as a function of evaluation steps during training.
}
\label{fig:mpo_no_kl_eval_agg}
\end{figure*}

This section contains two sets of plots. Figure \ref{fig:mpo_kl} contains bar plots comparing the performance of RE-MPO (blue bars), SRE-MPO (green bars) and E-MPO (red bars) across nine Mujoco domains. The performance of the agents as a function of evaluation steps is shown in Figure \ref{fig:mpo_kl_eval_agg} for all nine domains respectively. Figrue \ref{fig:mpo_no_kl} shows the bar plots for R-MPO, SR-MPO and MPO and Figure \ref{fig:mpo_no_kl_eval_agg} shows the corresponding performance of the agents as a function of evaluation steps.

%\nir{you give much lengthier explanation to what these plots are in the next section of the investigative experiments. maybe it makes sense to put it here..}

\subsection{Investigative Experiments}
\label{app:investigative}
This section contains additional investigative experiments that were mentioned in the main paper. 

\begin{figure*}
\centering
\newcommand{\scl}{0.3}
 \subfigure{
 \includegraphics[scale=\scl]{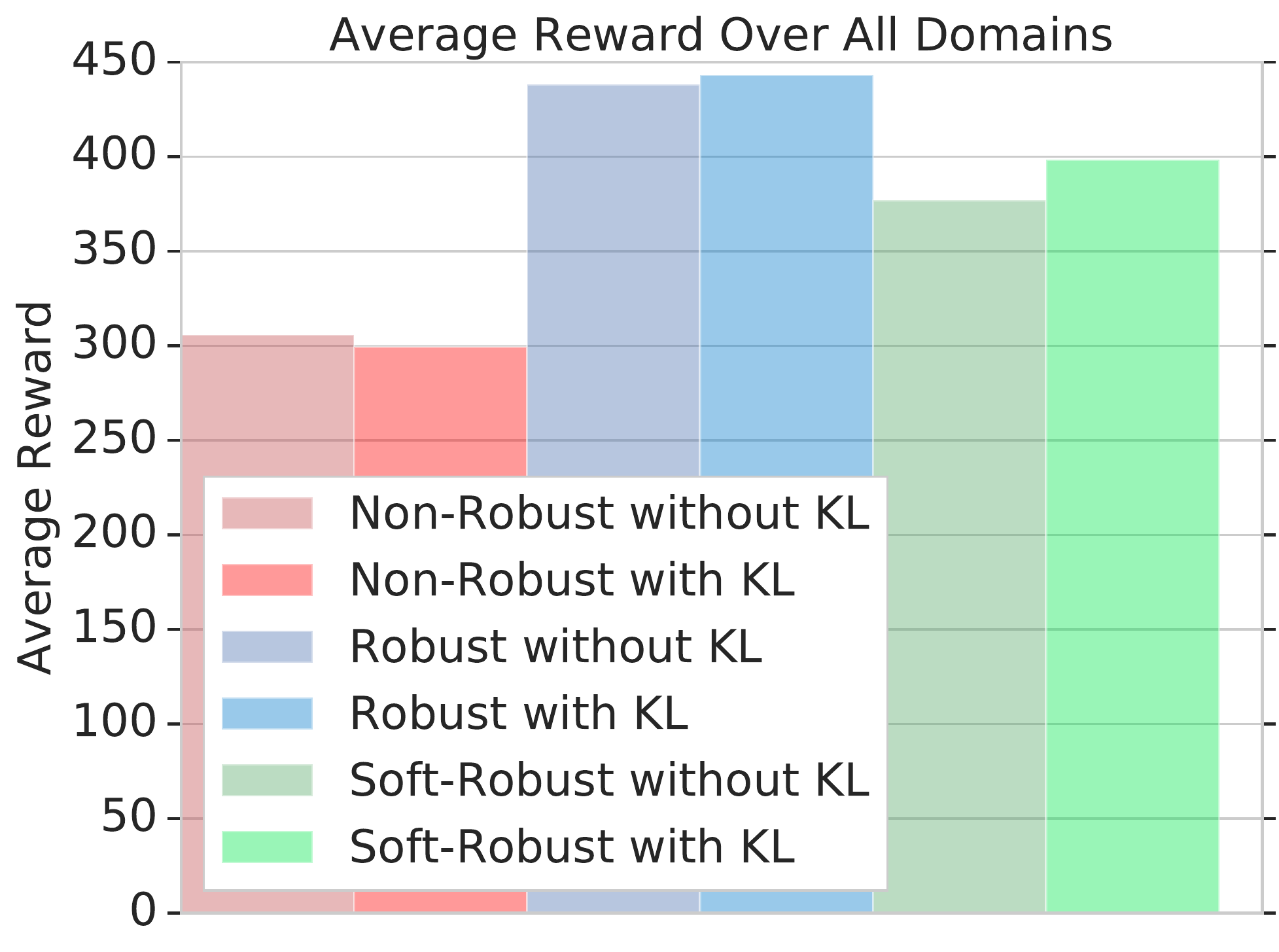}
 }
 \subfigure{
 \includegraphics[scale=\scl]{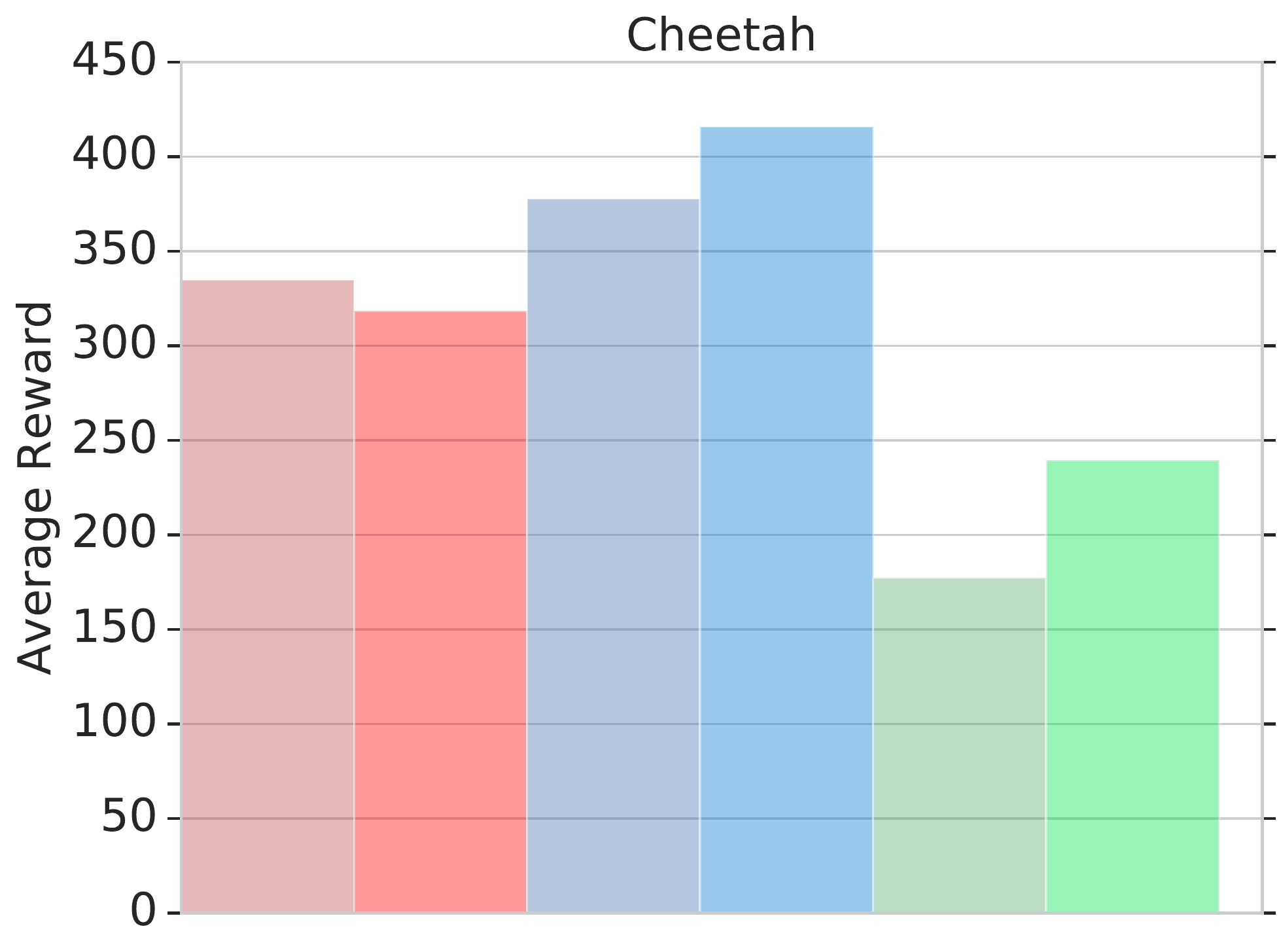}
 }
\caption{
Comparing entropy-regularized objective to the non-entropy regularized objective (left figure). The entropy-regularized version does no worse than the non entropy-regularized setup and in some cases, for example Cheetah, performs considerably better than the expected return objective (right figure).
}
\label{fig:app:entropy}
\end{figure*}

Figure \ref{fig:app:entropy} presents the difference in performance between the entropy-regularized agents and the \textbf{non} entropy-regularized agents agents. Although the performance is comparable (left figure), the entropy-regularized version performs no worse on average than the non-entropy-regularized agent. In addition, there are some tasks where there is a large improvement in performance, such as the Cheetah task for the entropy-regularized agent variants non entropy-regularized agent variants (right figure).

\begin{figure*}
\centering
\newcommand{\scl}{0.2}
 \subfigure{
 \includegraphics[scale=\scl]{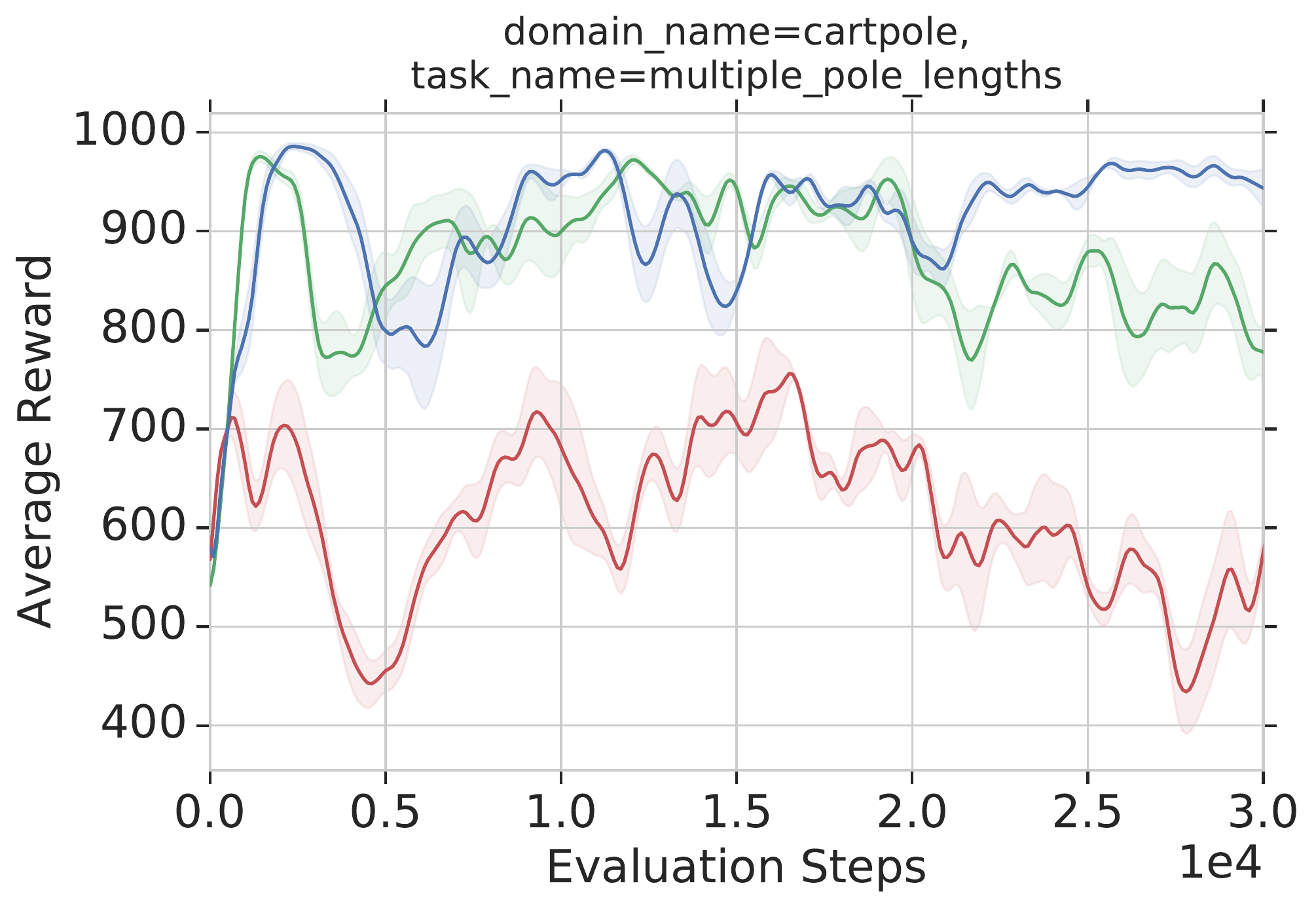}
 }
  \subfigure{
 \includegraphics[scale=\scl]{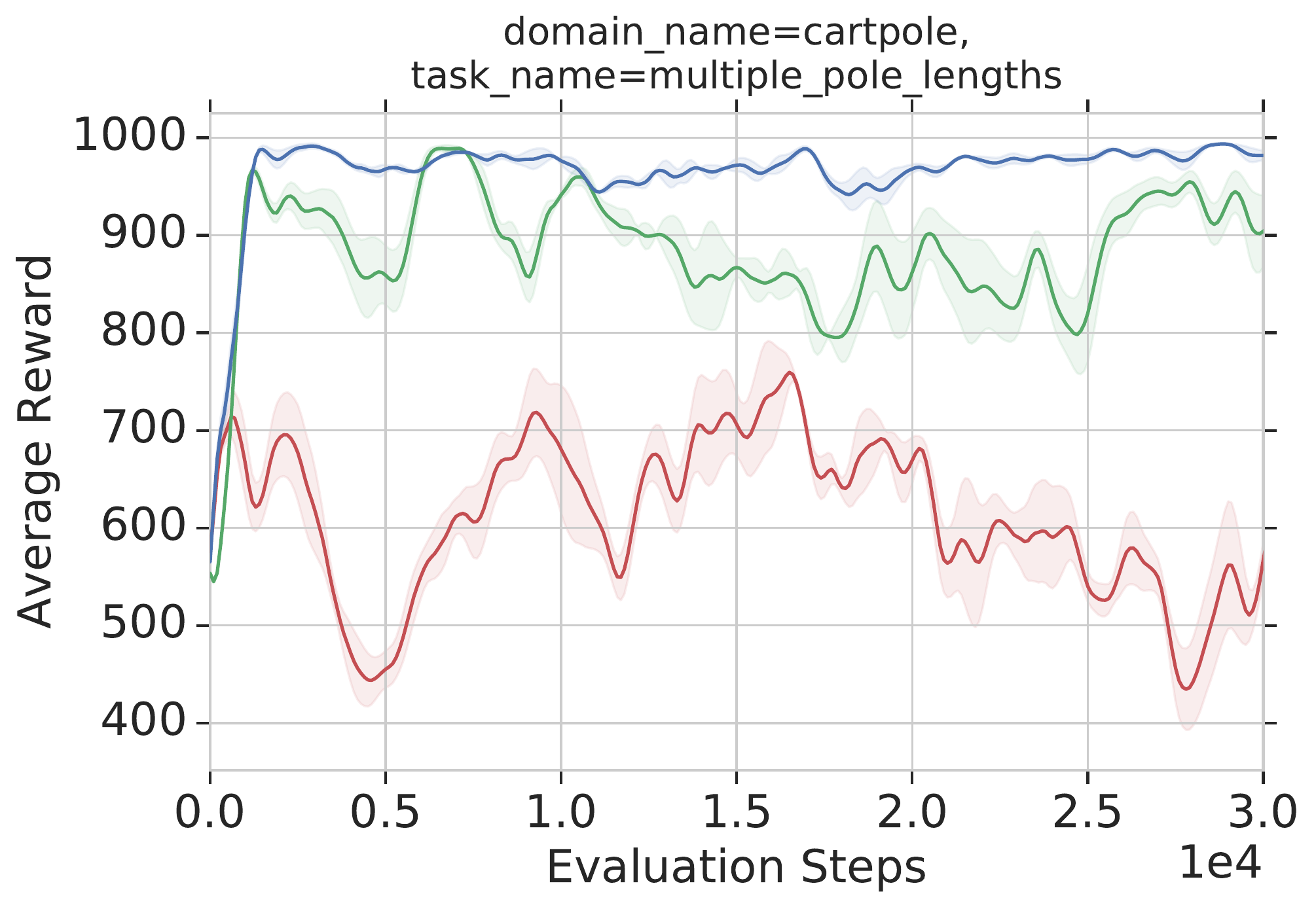}
 }
 \subfigure{
 \includegraphics[scale=\scl]{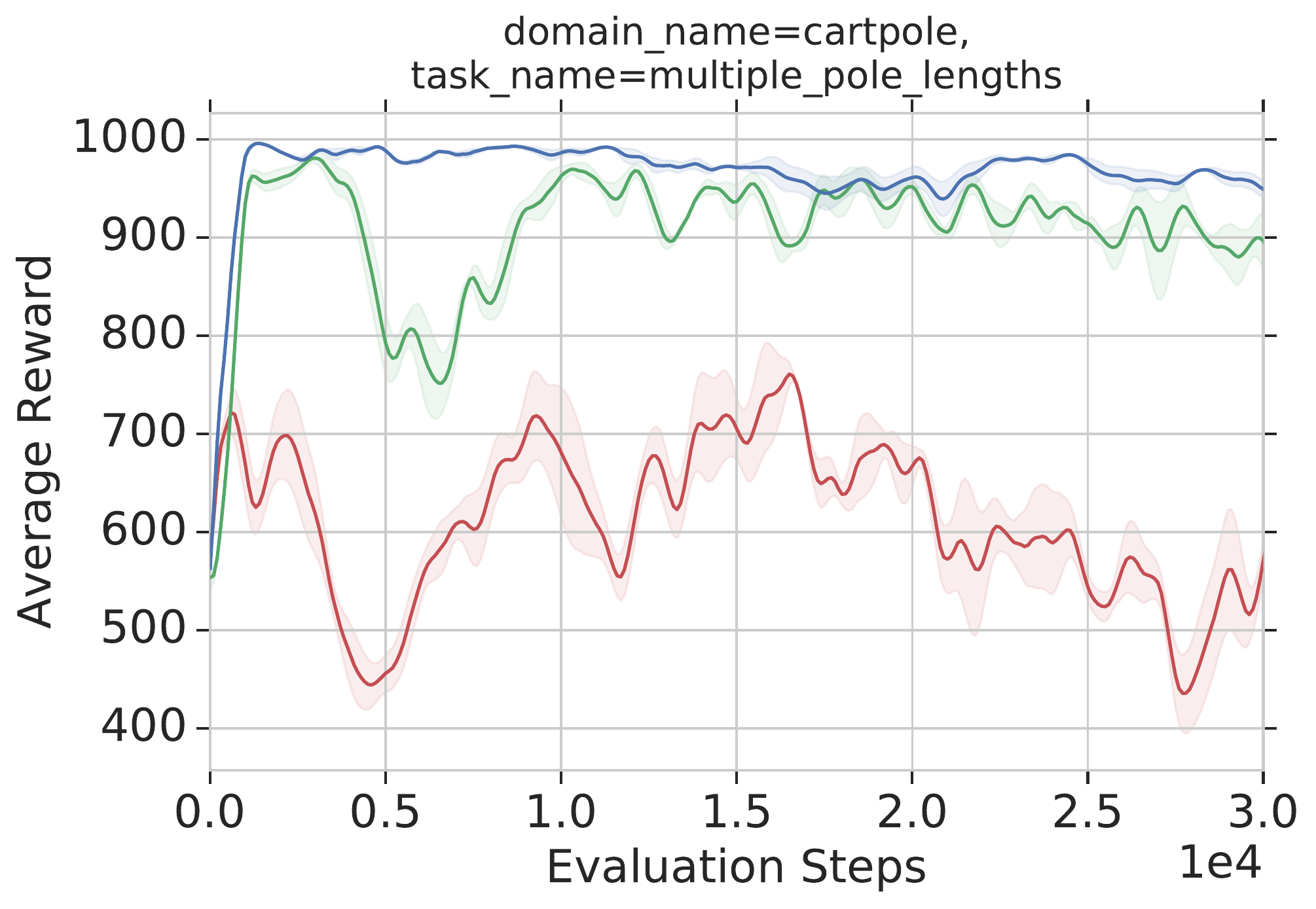}
 }
  \subfigure{
 \includegraphics[scale=\scl]{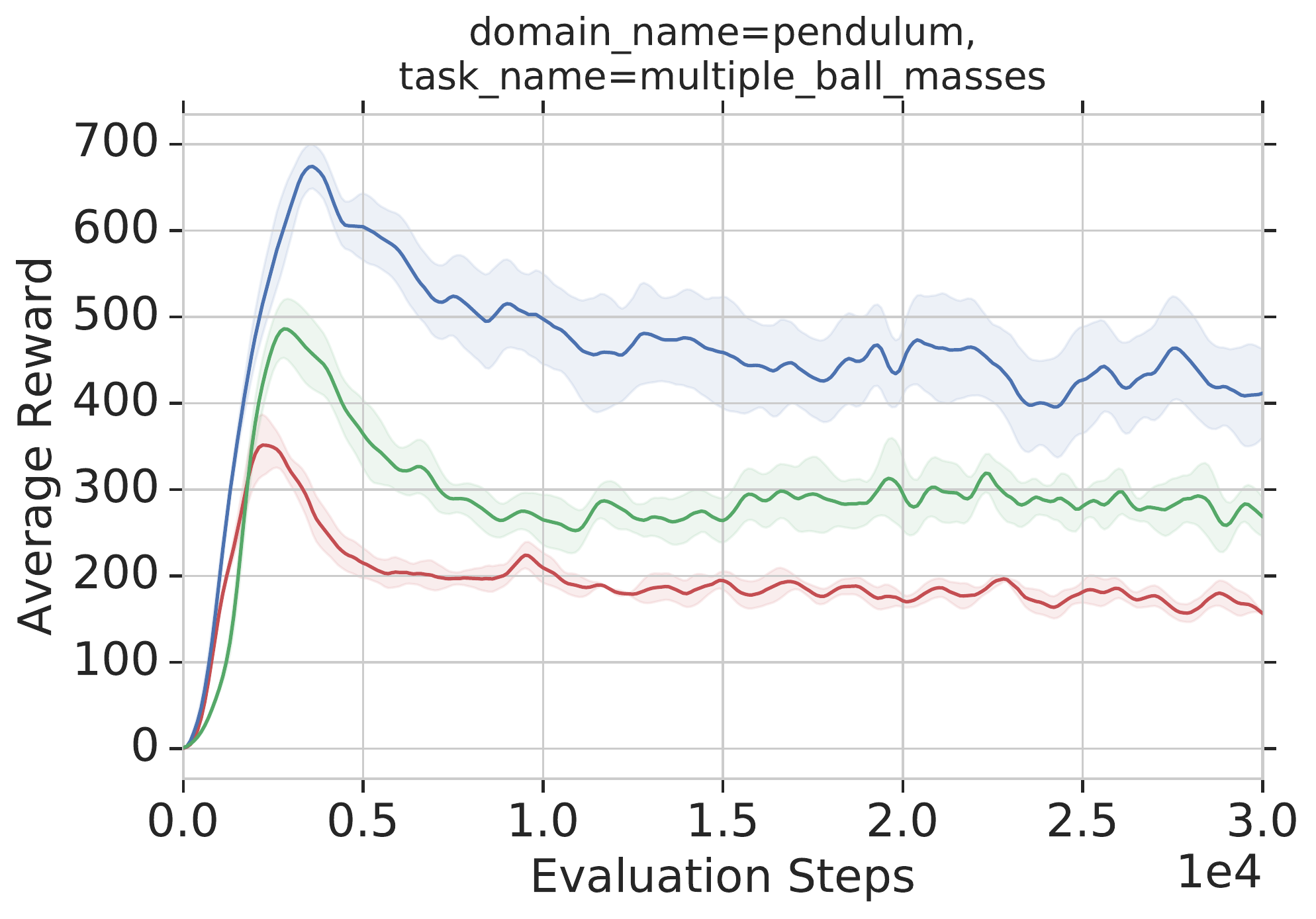}
 }
  \subfigure{
 \includegraphics[scale=\scl]{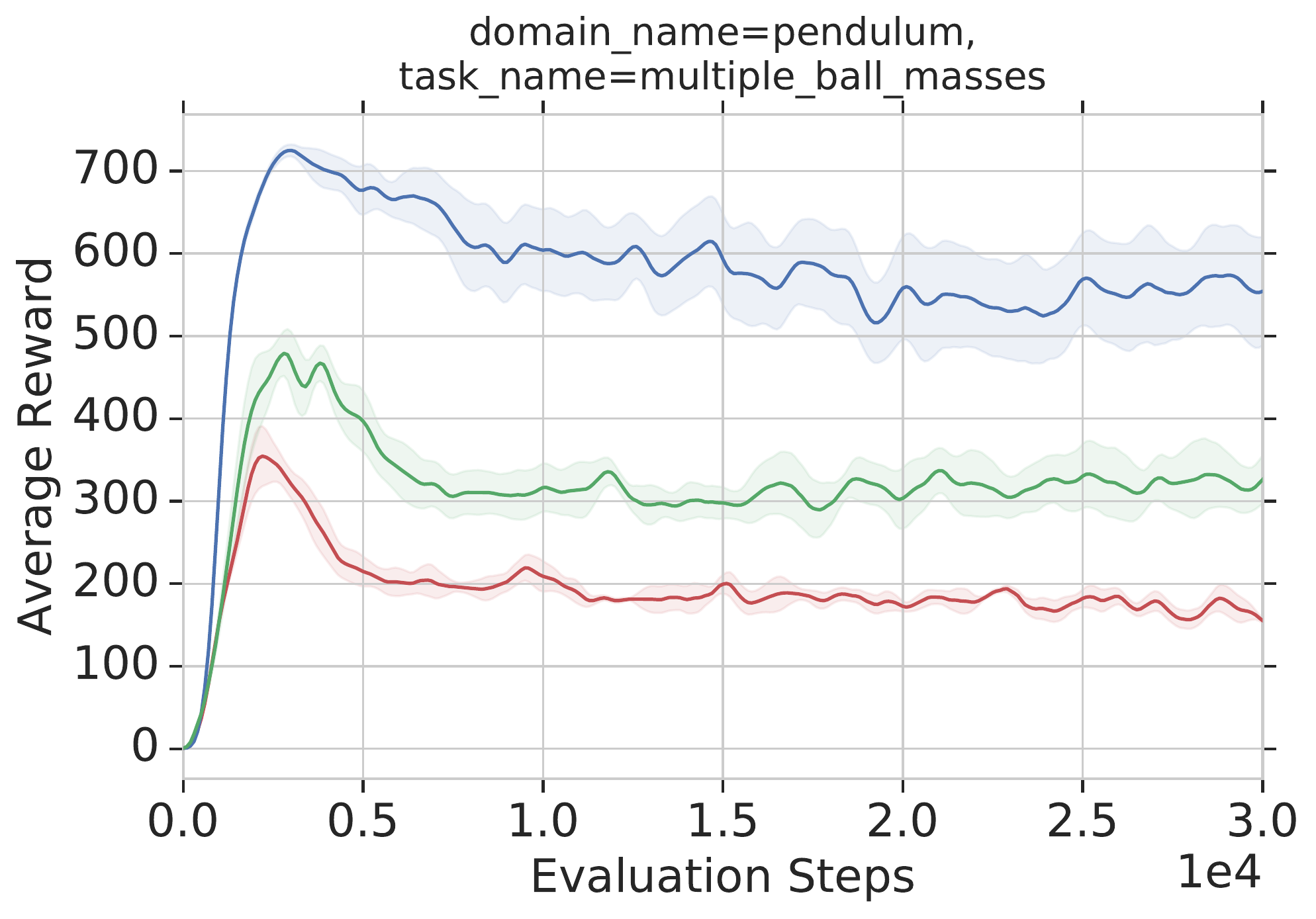}
 }
 \subfigure{
 \includegraphics[scale=\scl]{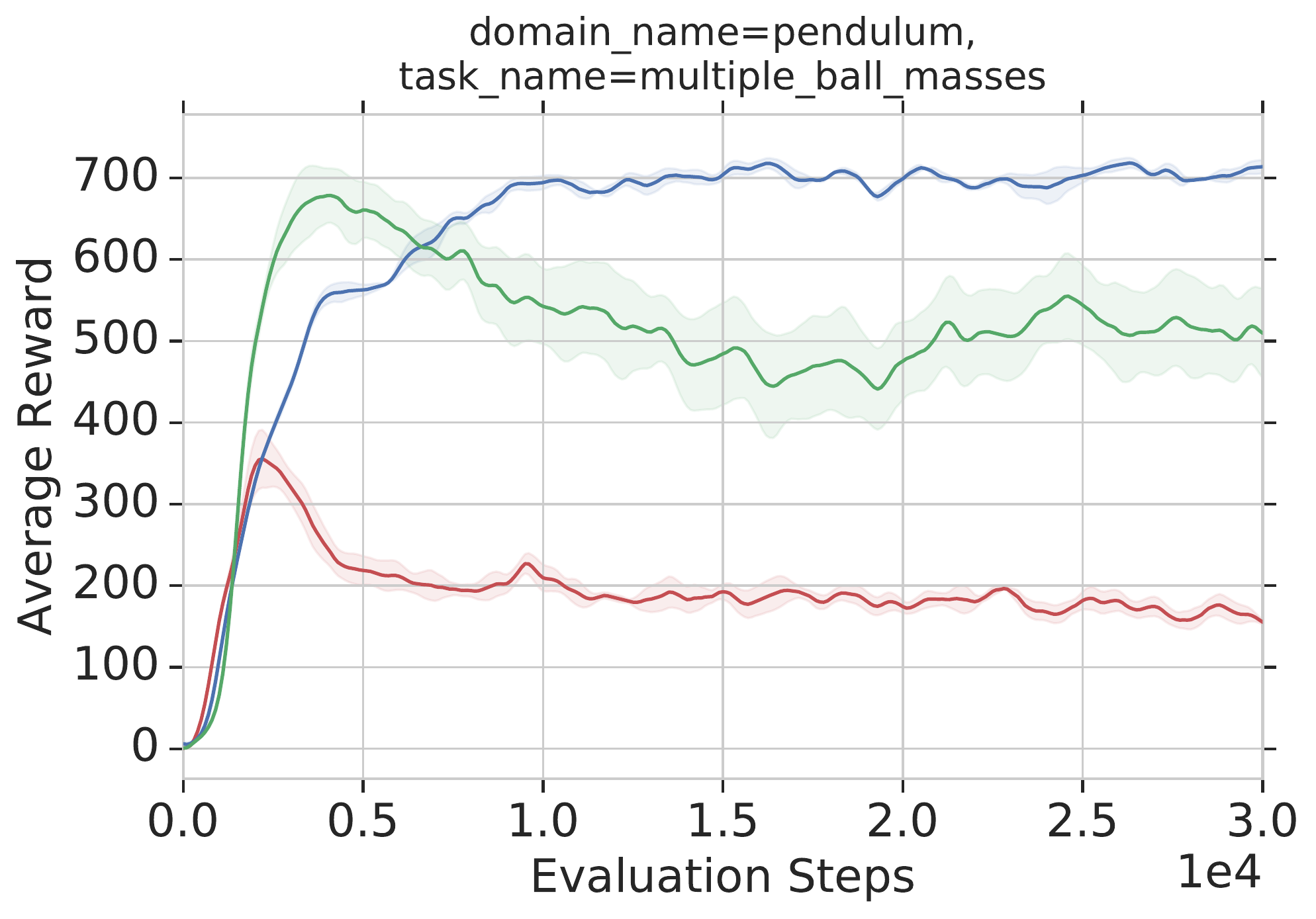}
 }
\caption{
Increasing the range of the training uncertainty set for Cartpole balance (top row) and Pendulum swingup (bottom row).
}
\label{fig:increasing_uncertainty}
\end{figure*}

\begin{figure*}
\centering
\newcommand{\scl}{0.3}
 \subfigure{
 \includegraphics[scale=\scl]{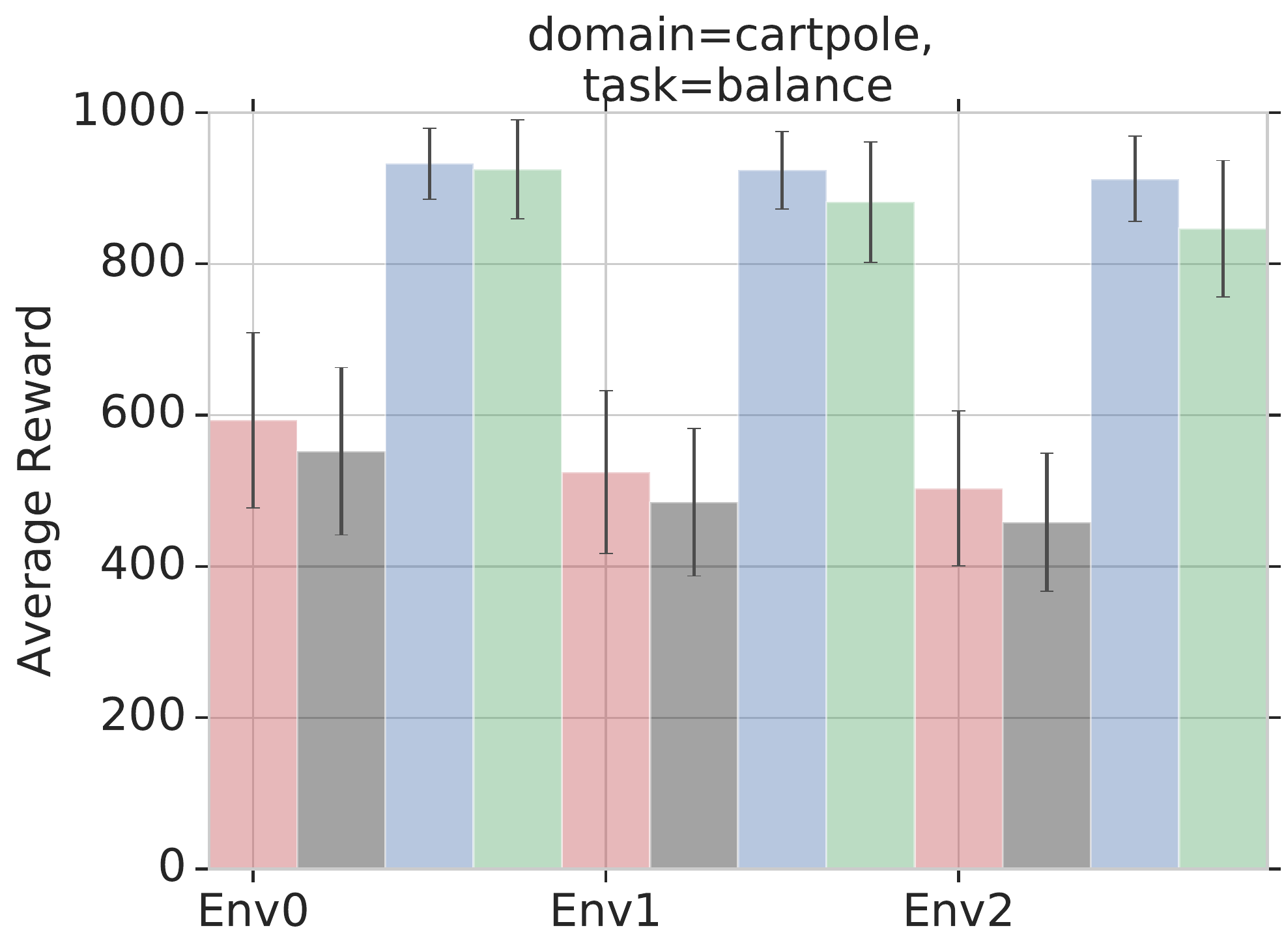}
 }
  \subfigure{
 \includegraphics[scale=\scl]{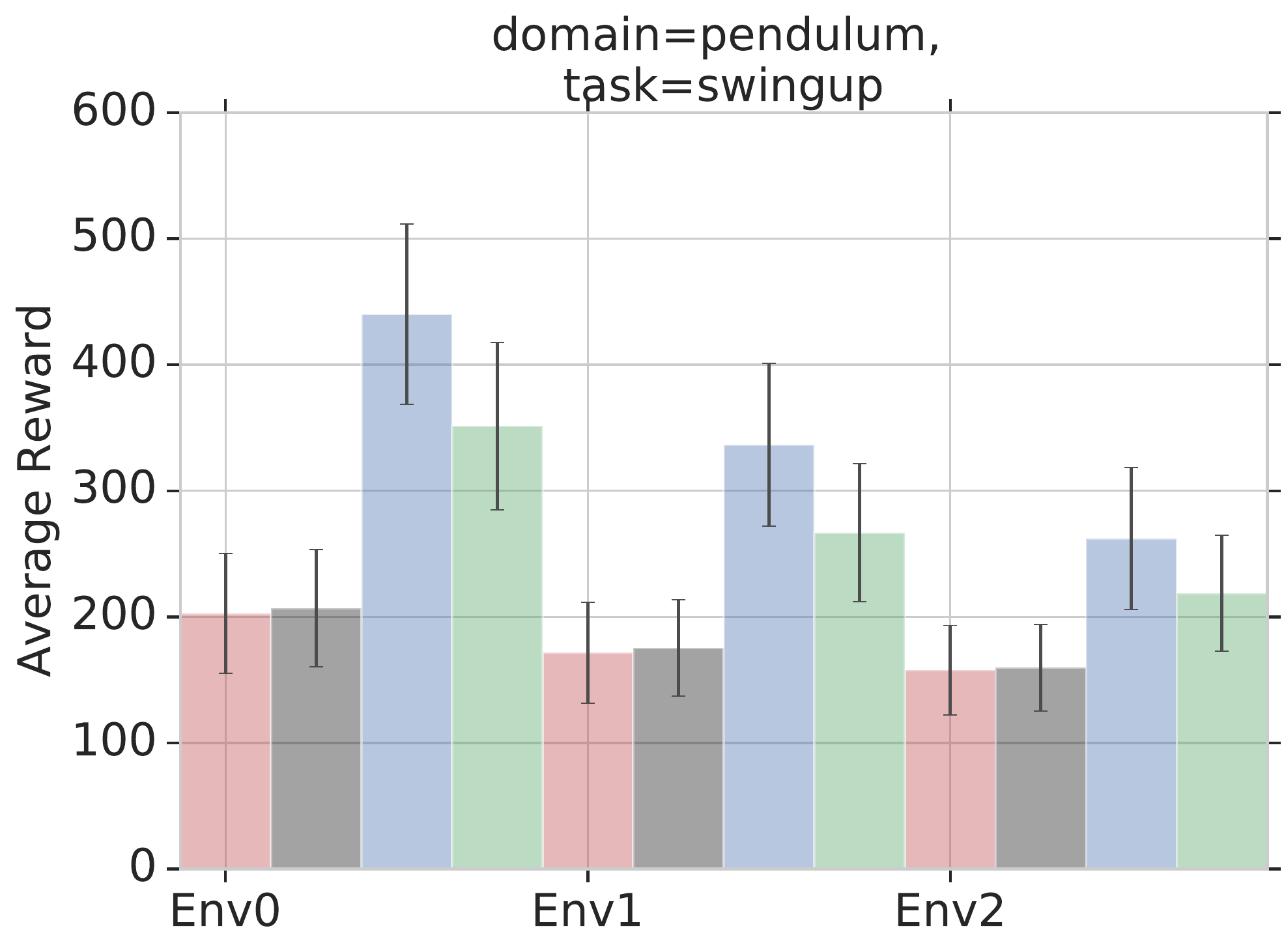}
 }
\caption{
Additional Training Samples: Two plots show $3$ times more additional training samples for non-robust E-MPO (dark grey) in the Cartpole Balance and Pendulum Swingup tasks respectively.} 
\label{fig:app:90k}
\end{figure*}

\textbf{Training with more samples}: Adding three times more samples to the non-robust baseline still yields significantly inferior performance compared to that of the robust and soft-robust versions as seen in Figure \ref{fig:app:90k} for Cartpole balance and Pendulum swingup respectively.

\begin{figure*}
\centering
\newcommand{\scl}{0.22}
 \subfigure{
 \includegraphics[scale=\scl]{./figures/Ablation/domain_randomization/cartpole_balance_dr_3_models.pdf}
 }
  \subfigure{
 \includegraphics[scale=\scl]{./figures/Ablation/domain_randomization/pendulum_swingup_dr_3_models}
 }
  \subfigure{
 \includegraphics[scale=\scl]{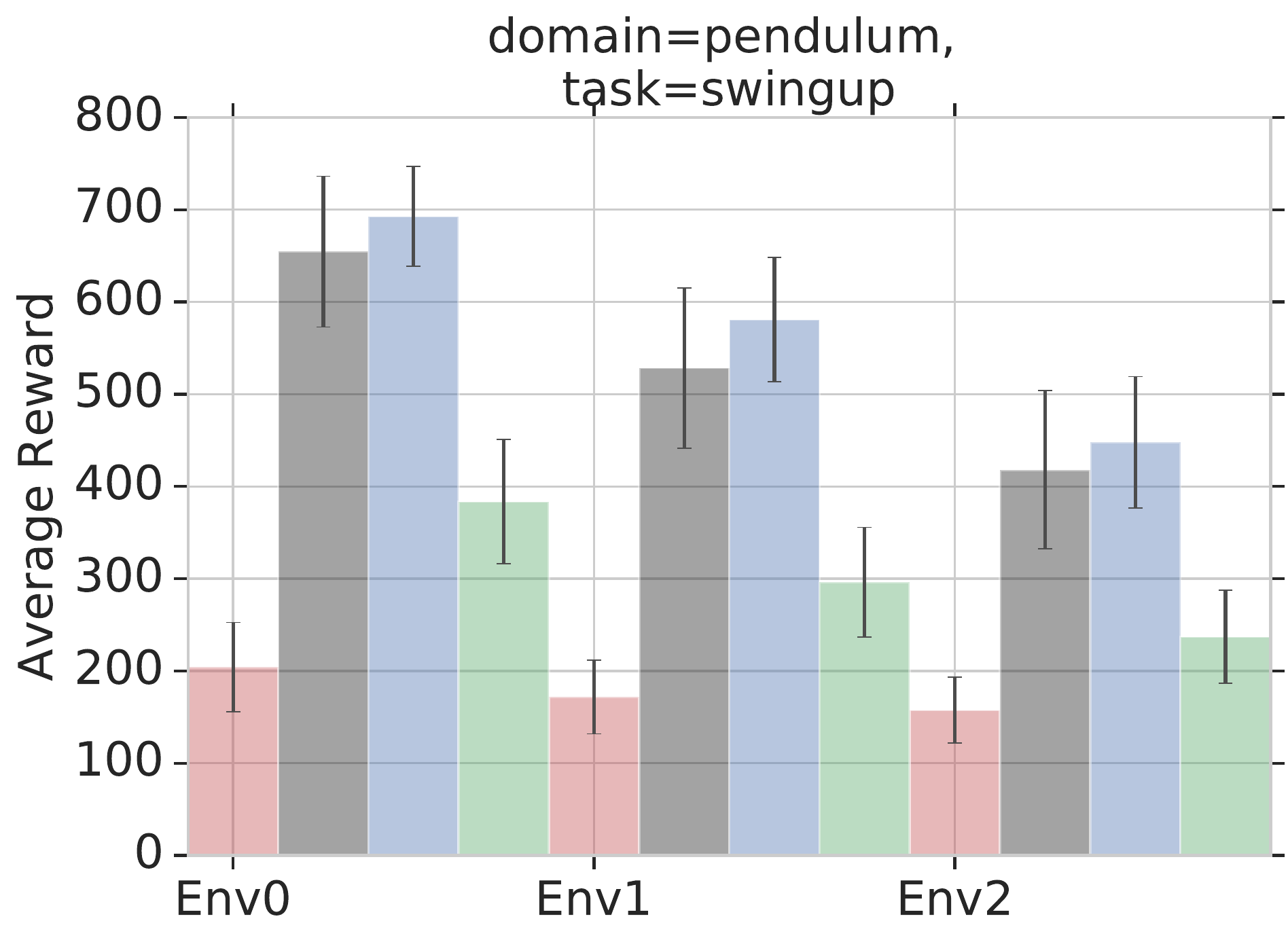}
 }
 
 \vspace{-0.3cm}
\caption{
Domain Randomization (DR): Domain randomization performance for the Cartpole balance (left) and Pendulum swingup (middle) tasks. As we increase the number of perturbations for DR to $100$ (right figure), we see that performance improves but still does not outperform RE-MPO, which still only uses $3$ perturbations.
}
\vspace{-0.3cm}
\label{fig:ablation3}
\end{figure*}

\textbf{What about Domain Randomization?} The DR results are shown in Figure \ref{fig:ablation3}. As can be seen in the figure, RE-MPO makes better use of a limited number of perturbations compared to \textit{Limited}-DR in Cartpole Balance (left) and Pendulum Swingup (middle) respectively. If the number of perturbations are increased to $100$ (right figure) for Pendulum Swingup, DR, which uses approximately $30$ times more perturbations, improves but still does not outperform RE-MPO.

\begin{figure*}
\centering
\newcommand{\scl}{0.2}
 \subfigure{
 \includegraphics[scale=\scl]{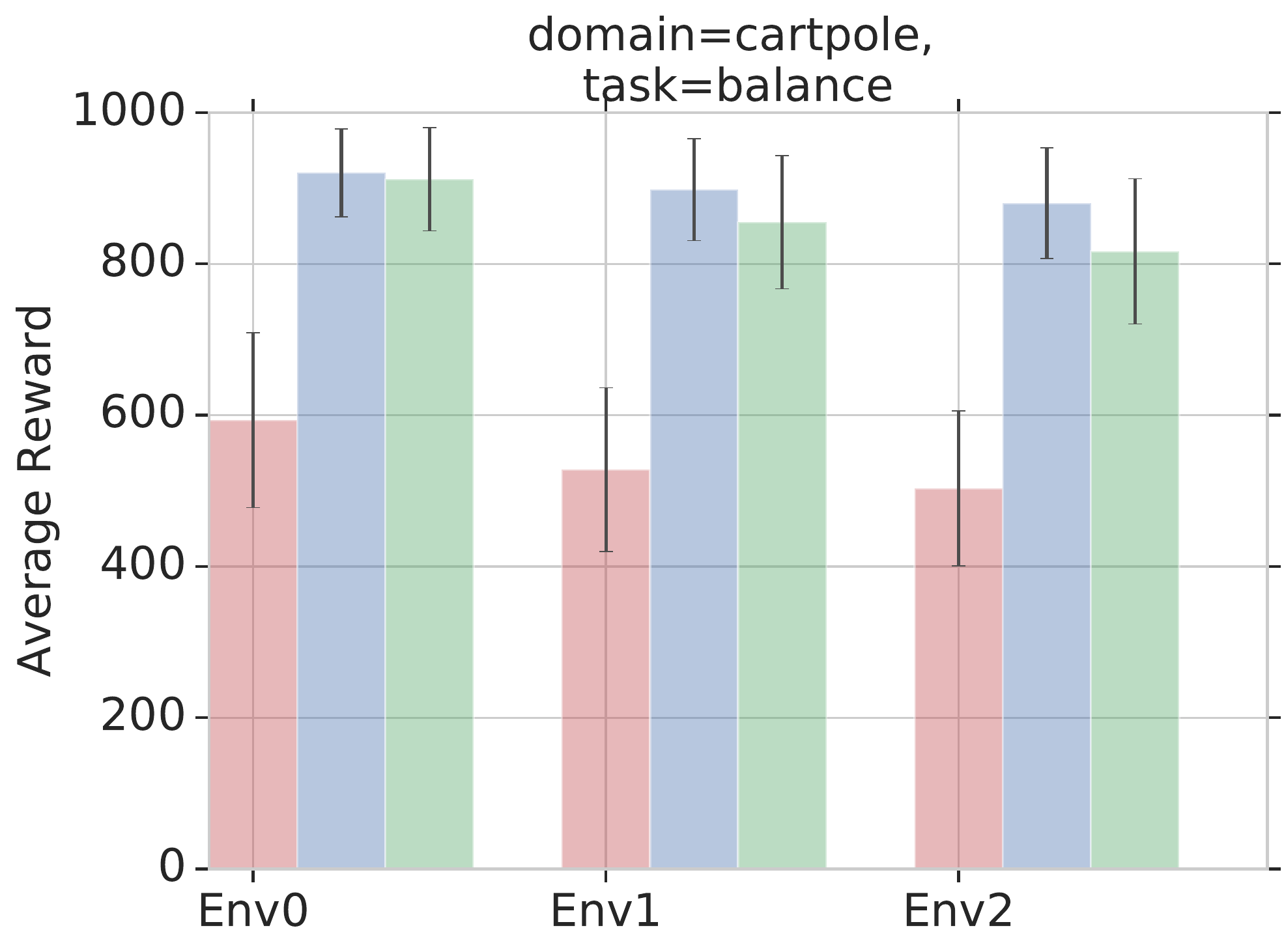}
 }
  \subfigure{
 \includegraphics[scale=\scl]{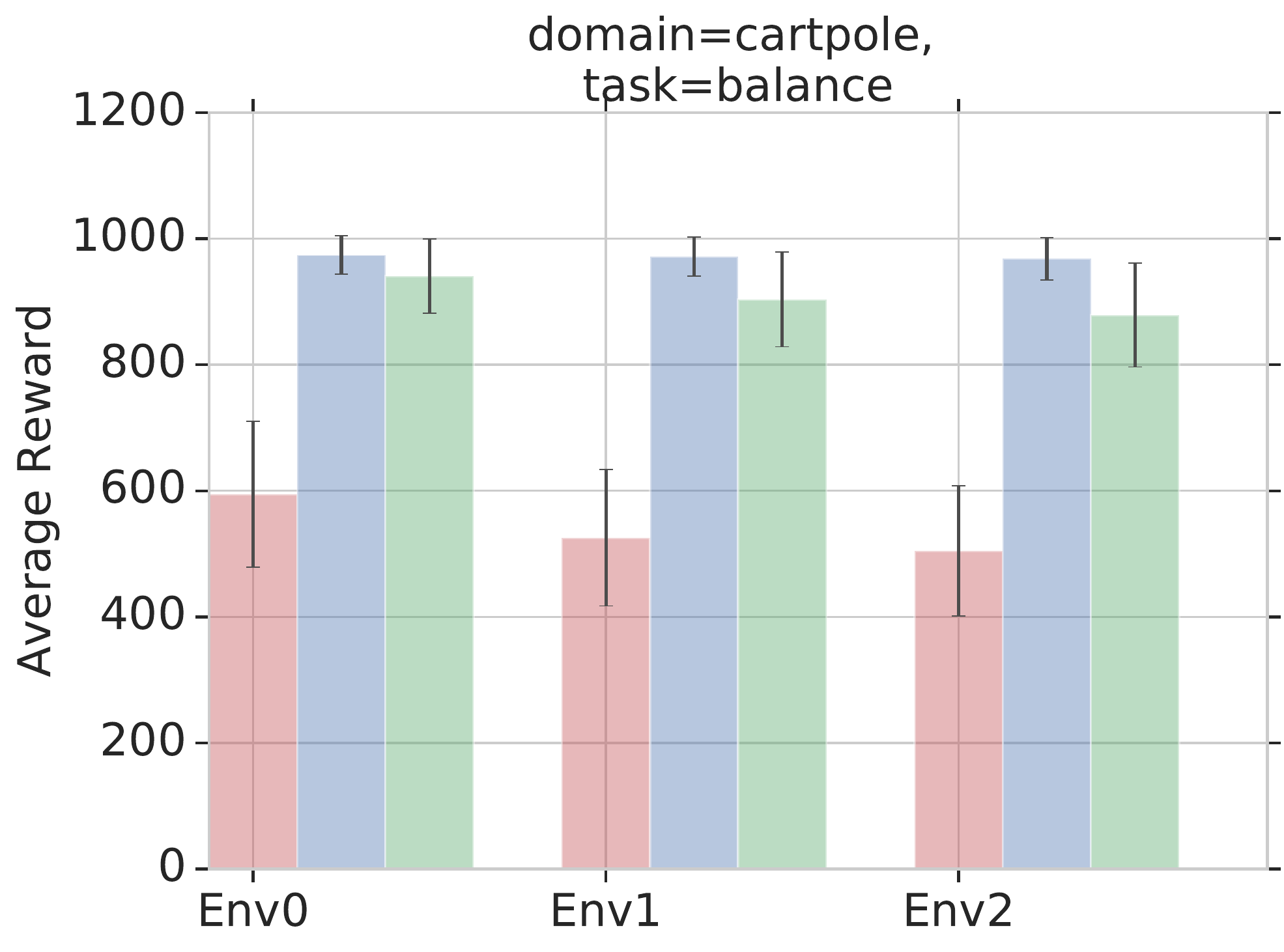}
 }
 \subfigure{
 \includegraphics[scale=\scl]{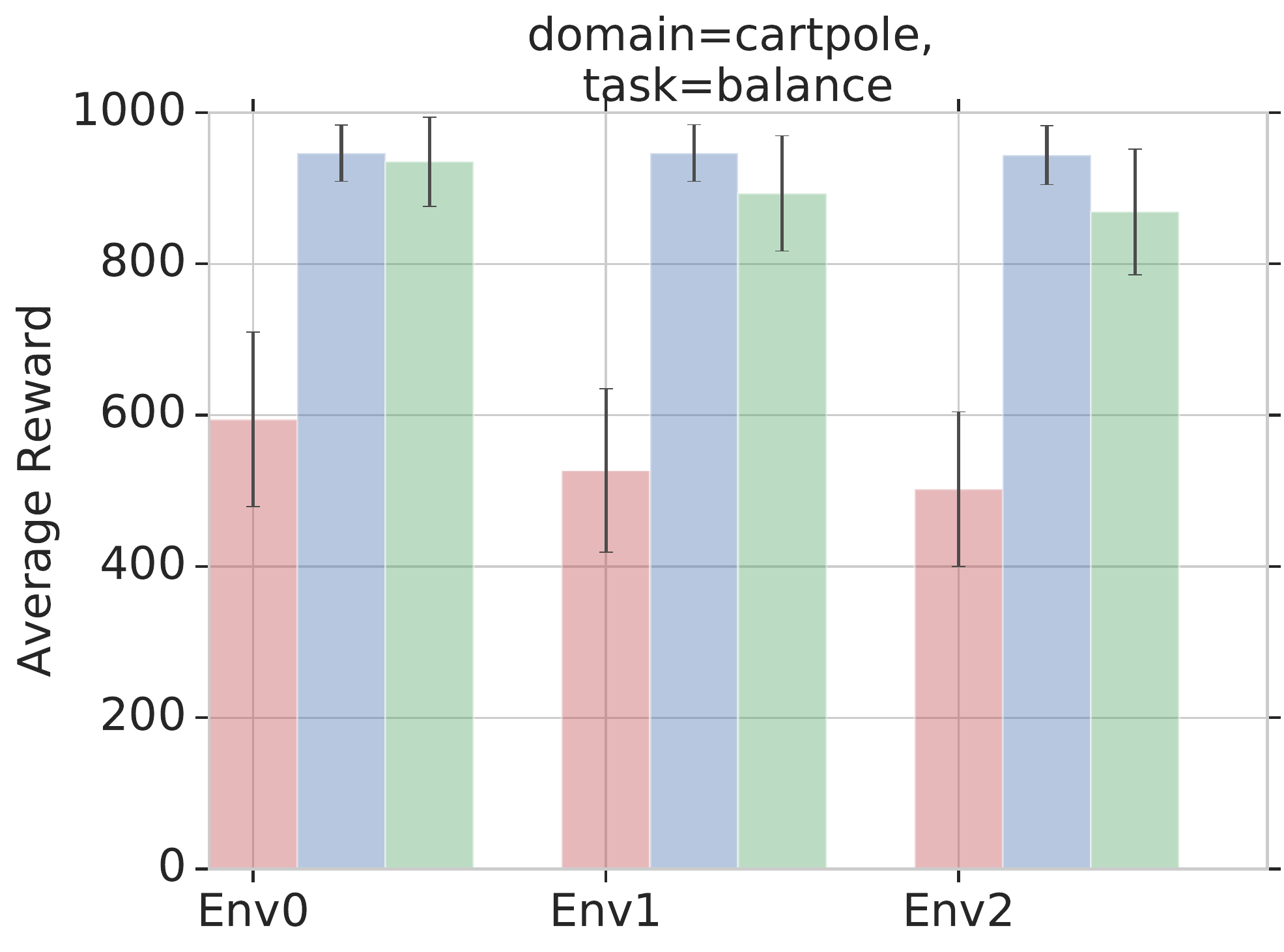}
 }
  \subfigure{
 \includegraphics[scale=\scl]{./figures/Ablation/perturb_uncertainty/pendulum_swingup_1p2}
 }
  \subfigure{
 \includegraphics[scale=\scl]{./figures/Ablation/perturb_uncertainty/pendulum_swingup_1p3}
 }
  \subfigure{
 \includegraphics[scale=\scl]{./figures/Ablation/perturb_uncertainty/pendulum_swingup_2p0}
 }
\caption{
Modifying the uncertainty set: The top row indicates the change in performance for Cartpole balance as the third perturbation of the uncertainty set is modified to $1.5, 2.5$ and $3.5$ meters respectively. The bottom row shows the performance for Pendulum Swingup for final perturbation changes of $1.2, 1.3$ and $2.0$ meters respectively.}
\label{fig:app:uncertainty}
\end{figure*}

\textbf{Modifying the uncertainty set}: Figure \ref{fig:app:uncertainty} contains the performance for cartpole balance (top row) and pendulum swingup (bottom row) when modifying the uncertainty set. 
For the Cartpole Balance task, the original uncertainty set training values are $0.5, 1.4$ and $2.1$ meters for the cartpole arm length. We modified the third perturbation ($2.1$ meters) of the uncertainty set to pole lengths of $1.5, 2.5$ and $3.5$ meters respectively. The agent is evaluated on pole lengths of $2.0, 2.2$ and $2.3$ meters respectively. As seen in the top row of Figure \ref{fig:app:uncertainty}, as the training perturbation is near the evaluation set, the performance of the robust and soft-robust agents are near optimal. However, as the perturbation increases further (i.e., $3.5$ meters), there is a drop in robustness performance. This is probably due to the agent learning a policy that is robust with respect to perturbations that are relatively far from the unseen evaluation set. However, the agent still performs significantly better than the non-robust baseline in each case.  
For Pendulum Swingup, the original uncertainty set values of the pendulum arm are $1.0, 1.1$ and $1.4$ meters. We modified the final perturbation to values of $1.2, 1.3$ and $2.0$ meters respectively. The agent is evaluated on unseen lengths of $1.5, 1.6$ and $1.7$ meters. A significant increase in performance can be seen in the bottom row of Figure \ref{fig:app:uncertainty} as the third perturbation approaches that of the unseen evaluation environments. Thus it appears that if the agent is able to approximately capture the dynamics of the unseen test environments within the training set, then the robust agent is able to adapt to the unseen test environments. Figure \ref{fig:increasing_uncertainty} presents the evaluation curves for the corresponding Cartpole Balance (top row) and Pendulum swingup (bottom row) tasks as the third perturbation of the uncertainty set is modified.

\begin{figure*}
\centering
\newcommand{\scl}{0.24}
 \subfigure{
 \includegraphics[scale=\scl]{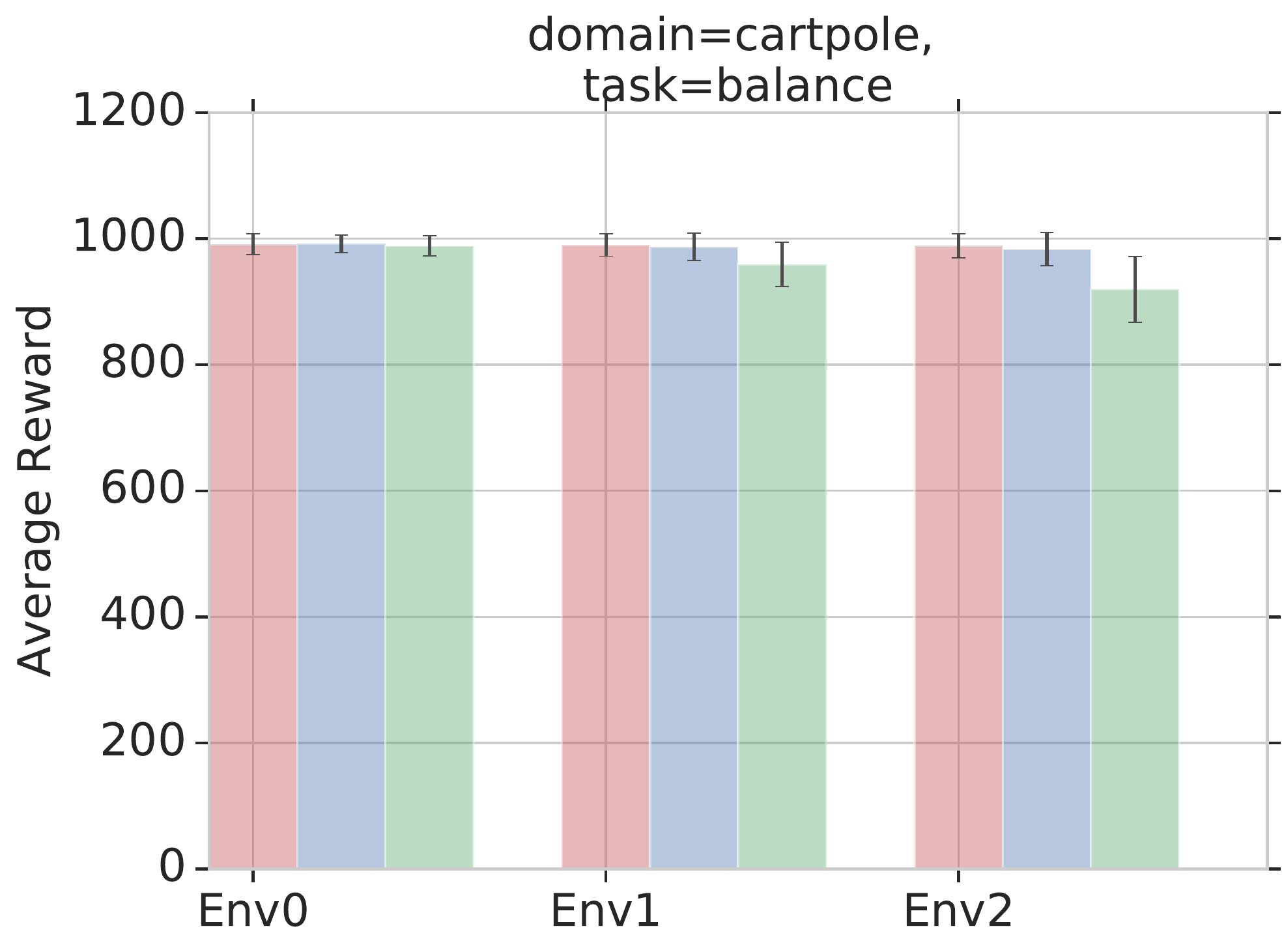}
 }
  \subfigure{
 \includegraphics[scale=\scl]{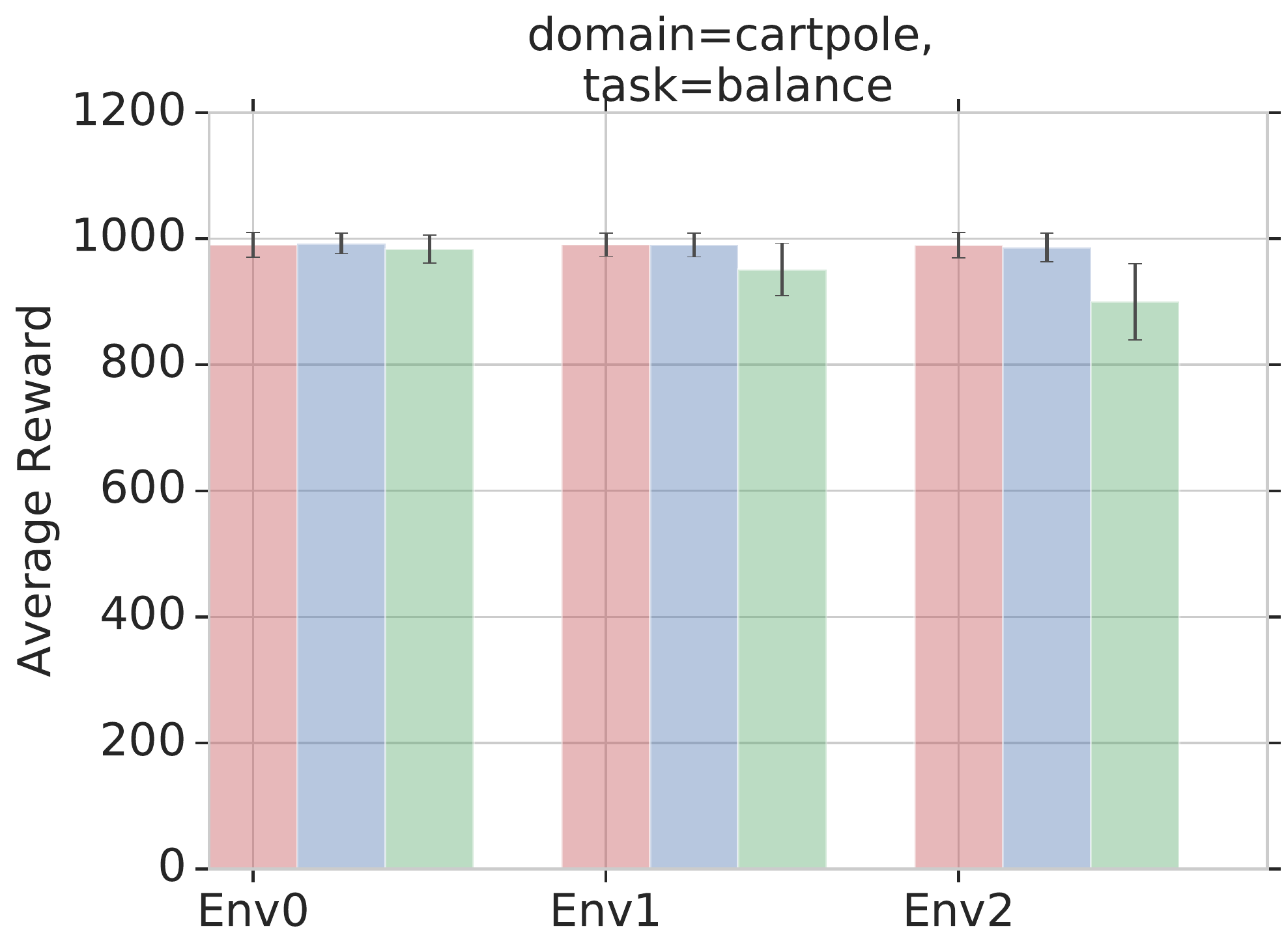}
 }
  \subfigure{
 \includegraphics[scale=\scl]{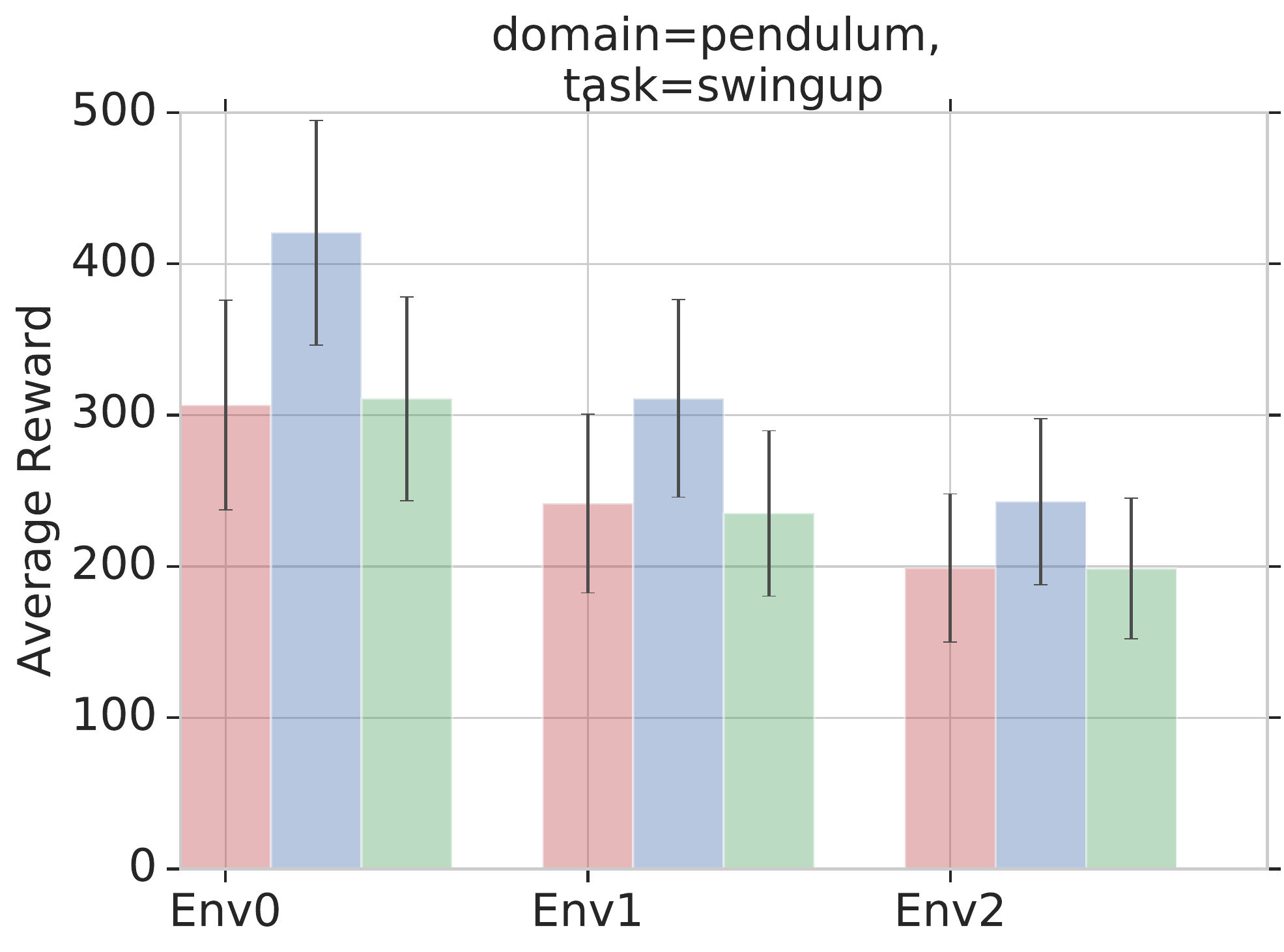}
 }
  \subfigure{
 \includegraphics[scale=\scl]{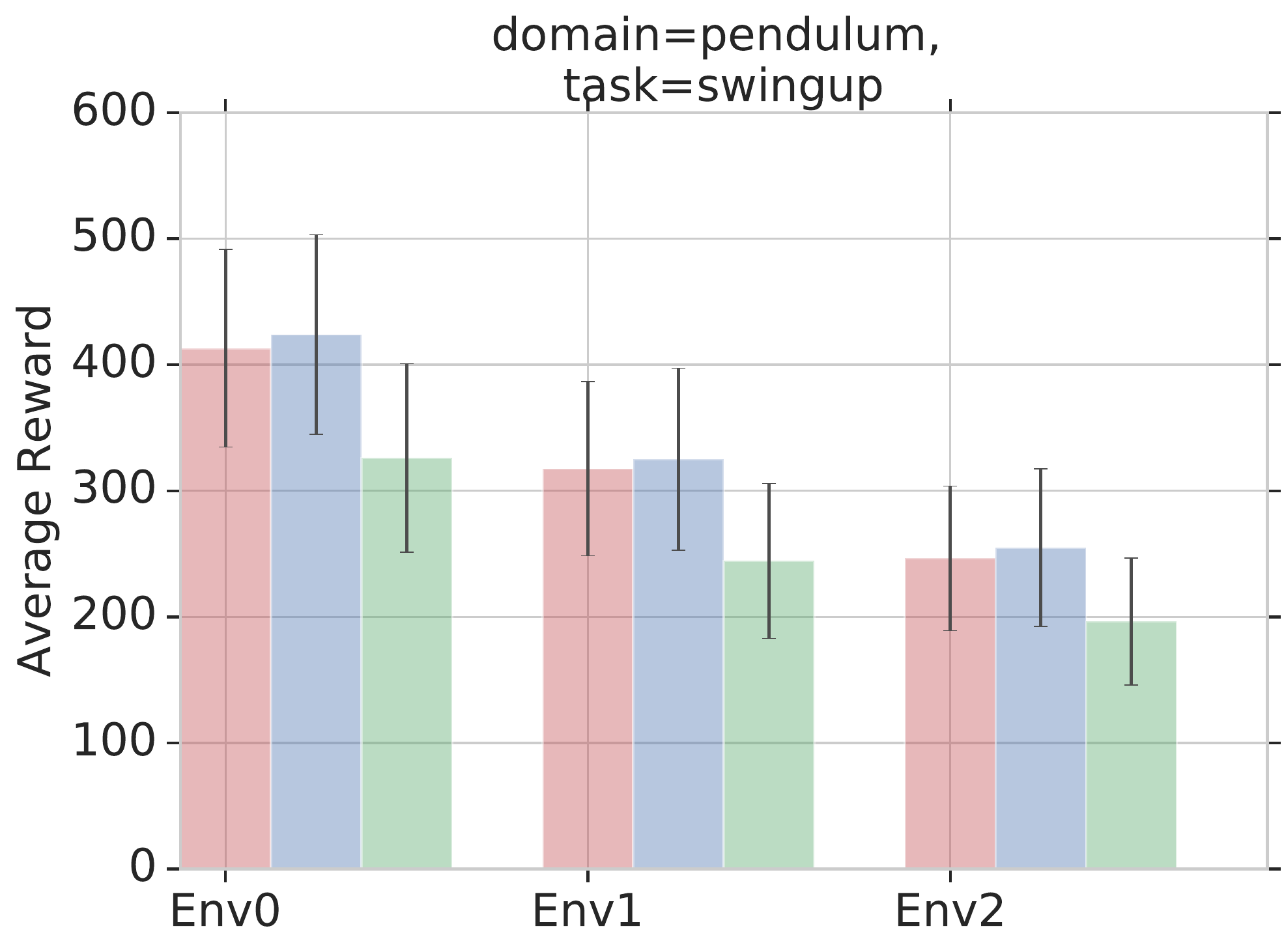}
 }
\caption{
Changing the nominal model: The top two figures indicate setting the nominal model as the median and largest perturbation of the uncertainty set for Cartpole Balance respectively. The right two figures are the same setting but for the Pendulum swingup domain. Legend: E-MPO (red), RE-MPO (blue), SRE-MPO (green).}
\label{fig:app:nominal}
\end{figure*}

\textbf{Different Nominal Models}: Figure \ref{fig:app:nominal} indicates the effect of changing the nominal model to the median and largest perturbation from the uncertainty set for the Cartpole balance (top row) and Pendulum swingup (bottom row) tasks respectively. For Cartpole, since the median and largest perturbations are significantly closer to the evaluation set, performance of the non-robust, robust and soft-robust agents are comparable. However, for Pendulum swingup, the middle actor is still far from the evaluation set and here the robust agent significantly outperforms the non-robust agent.

\clearpage
\newpage
\section{Algorithm}
\label{app:algorithm}

The Robust MPO algorithm is defined as Algorithm \ref{Alg:GradientFree}. The algorithm can be divided into three steps: Step (1) perform policy evaluation on the policy $\pi_k$; Step (2) build a proposal distribution $q(a|s)$ from the action value function $Q_\theta^{\pi_{k}}$; Step (3) update the policy by minimizing the KL divergence between the proposal distribution $q$ and the policy $\pi$. The corresponding robust entropy-regularized version can be seen in Algorithm \ref{Alg:GradientFreeEntropy} and the soft-robust entropy-regularized version in Algorithm \ref{Alg:GradientFreeEntropySoft}.

\begin{algorithm}[t]
\small
\caption{Robust MPO (R-MPO) algorithm for a single iteration}\label{Alg:GradientFree}
\begin{algorithmic}[1]
\STATE {\bf given} batch-size (K), number of actions (N), old-policy $\pi_{k}$ and replay-buffer
\STATE {\bf // Step 1: Perform policy evaluation on $\pi_{k}$ to yield $Q_{\theta}^{\pi_{k}}$}
\STATE 

\begin{equation*}
\min_{\theta} \biggl(r_t + \gamma \inf_{p \in \mathcal{P}(s_t, a_t)} \biggl[Q_{\hat{\theta}}^{\pi_{k}}(s_{t+1} \sim p(\cdot | s_t, a_t), a_{t+1}\sim \pi_{k}( \cdot \vert s_{t+1}))  \biggr] - Q_{\theta}^{\pi_{k}}(s_t, a_t) \biggr)^2 \enspace ,
\end{equation*}

% \begin{equation*}
% \min_{\theta} \biggl(r_t + \gamma \inf_{p \in \mathcal{P}(s_t, a_t)} \mathbb{E}^p \biggl[Q_{\theta}^{\pi_{k}}(s_{t+1}, a_{t+1}\sim \pi_{k}( \cdot \vert s_{t+1})), p \in \mathcal{P}(s_t,a_t)  \biggr] - Q_{\theta}^{\pi_{k}}(s_t, a_t) \biggr)^2 \enspace ,
% \label{eqn:rmpotd}
% \end{equation*}

\REPEAT
\STATE {Sample batch of size N from replay buffer}
\STATE {\bf // Step 2: sample based policy (weights)}
\STATE $q(a_i | s_j) = q_{ij}$, {\bf computed as:}
\FOR{j = 1,...,$K$} 
\FOR{i = 1,...,$N$}
\STATE {$a_{i} \sim \pi_\text{k}(a|s_j)$}
\STATE {$Q_{ij} = Q^{\pi_{k}}(s_{j}, a_i)$} 
\STATE $q_{ij} =$ {\bf Compute Weights}($\{Q_{ij}\}_{i=1\dots N}$) \Comment{See \citep{abdolmaleki2018maximum}}
\ENDFOR
\ENDFOR
\STATE {\bf // Step 3: update parametric policy}
\STATE {Given the data-set $\{s_j,(a_{i},q_{ij})_{i=1...N}\}_{j=1...K}$}
\STATE {\bf Update the Policy by finding }
\STATE $\pi_{k+1} = \textrm{argmax}_\pi \sum_j^K \sum_i^N q_{ij} \log \pi(a_i|s_j)$
\STATE {\bf (subject to additional (KL) regularization)}
\UNTIL{Fixed number of steps}
\STATE return $\pi_{k+1}$
\end{algorithmic}
\label{alg:1}
\end{algorithm}
%\vspace{-0.5cm}

\begin{algorithm}[t]
\small
\caption{Robust Entropy-Regularized MPO (RE-MPO) algorithm  for a single iteration}\label{Alg:GradientFreeEntropy}
\begin{algorithmic}[1]
\STATE {\bf given} batch-size (K), number of actions (N), old-policy $\pi_{k}$ and replay-buffer
\STATE {\bf // Step 1: Perform policy evaluation on $\pi_{k}$ to yield $Q_{\theta}^{\pi_{k}}$}
\STATE 

\begin{equation*}
\begin{aligned}
    \min_{\theta} \biggl(r_t + & \gamma \inf_{p \in \mathcal{P}(s_t,a_t)} \bigg[\widetilde Q_{\text{R-KL},\hat{\theta}}^{\pi_k}(s_{t+1} \sim p(\cdot | s_t, a_t), a_{t+1}\sim \pi_k(\cdot \vert s_{t+1}) ; \bar{\pi})\\
    & - \tau \text{KL}(\pi_k(\cdot | s_{t+1} \sim p(\cdot | s_t, a_t)) \| \bar{\pi}(\cdot | s_{t+1} \sim p(\cdot | s_t, a_t))) \bigg]  - \widetilde Q_{\text{KL},\theta}^{\pi_k}(s_t, a_t ; \bar{\pi}) \biggr)^2,
\end{aligned}
\end{equation*}

\REPEAT
\STATE {Sample batch of size N from replay buffer}
\STATE {\bf // Step 2: sample based policy (weights)}
\STATE $q(a_i | s_j) = q_{ij}$, {\bf computed as:}
\FOR{j = 1,...,$K$} 
\FOR{i = 1,...,$N$}
\STATE {$a_{i} \sim \pi_\text{k}(a|s_j)$}  
\STATE {$Q_{ij} = Q^{\pi_{k}}(s_{j}, a_i)$} 
\STATE $q_{ij} =$ {\bf Compute Weights}($\{Q_{ij}\}_{i=1\dots N}$) \Comment{see \citep{abdolmaleki2018maximum}}
\ENDFOR
\ENDFOR
\STATE {\bf // Step 3: update parametric policy}
\STATE {Given the data-set $\{s_j,(a_{i},q_{ij})_{i=1...N}\}_{j=1...K}$}
\STATE {\bf Update the Policy by finding }
\STATE $\pi_{k+1} = \textrm{argmax}_\pi \sum_j^K \sum_i^N q_{ij} \log \pi(a_i|s_j)$
\STATE {\bf (subject to additional (KL) regularization)}
\UNTIL{Fixed number of steps}
\STATE return $\pi_{k+1}$
\end{algorithmic}
\label{alg:2}
\end{algorithm}
%\vspace{-0.5cm}

\begin{algorithm}[t]
\small
\caption{Soft-Robust Entropy-Regularized MPO (SRE-MPO) algorithm  for a single iteration}\label{Alg:GradientFreeEntropySoft}
\begin{algorithmic}[1]
\STATE {\bf given} batch-size (K), number of actions (N), old-policy $\pi_{k}$ and replay-buffer
\STATE {\bf // Step 1: Perform policy evaluation on $\pi_{k}$ to yield $Q_{\theta}^{\pi_{k}}$}
\STATE 

\begin{equation*}
\begin{aligned}
    \min_{\theta} \biggl(r_t + & \gamma  \bigg[\widetilde Q_{\text{R-KL},\hat{\theta}}^{\pi_k}(s_{t+1} \sim \bar{p}(\cdot | s_t, a_t), a_{t+1}\sim \pi_k(\cdot \vert s_{t+1}) ; \bar{\pi})\\
    & - \tau \text{KL}(\pi_k(\cdot | s_{t+1} \sim \bar{p}(\cdot | s_t, a_t)) \| \bar{\pi}(\cdot | s_{t+1} \sim \bar{p}(\cdot | s_t, a_t))) \bigg]  - \widetilde Q_{\text{KL},\theta}^{\pi_k}(s_t, a_t ; \bar{\pi}) \biggr)^2,
\end{aligned}
\end{equation*}

\REPEAT
\STATE {Sample batch of size N from replay buffer}
\STATE {\bf // Step 2: sample based policy (weights)}
\STATE $q(a_i | s_j) = q_{ij}$, {\bf computed as:}
\FOR{j = 1,...,$K$} 
\FOR{i = 1,...,$N$}
\STATE {$a_{i} \sim \pi_\text{k}(a|s_j)$}  
\STATE {$Q_{ij} = Q^{\pi_{k}}(s_{j}, a_i)$} 
\STATE $q_{ij} =$ {\bf Compute Weights}($\{Q_{ij}\}_{i=1\dots N}$) \Comment{see \citep{abdolmaleki2018maximum}}
\ENDFOR
\ENDFOR
\STATE {\bf // Step 3: update parametric policy}
\STATE {Given the data-set $\{s_j,(a_{i},q_{ij})_{i=1...N}\}_{j=1...K}$}
\STATE {\bf Update the Policy by finding }
\STATE $\pi_{k+1} = \textrm{argmax}_\pi \sum_j^K \sum_i^N q_{ij} \log \pi(a_i|s_j)$
\STATE {\bf (subject to additional (KL) regularization)}
\UNTIL{Fixed number of steps}
\STATE return $\pi_{k+1}$
\end{algorithmic}
\label{alg:3}
\end{algorithm}

\end{document}